\documentclass[10pt,journal,compsoc]{IEEEtran}
\usepackage[utf8]{inputenc}

\title{Enhance Accuracy: Sensitivity and Uncertainty Theory in LiDAR Odometry and Mapping}
\author{Zeyu Wan,
	Yu Zhang,
	Bin He,
	Zhuofan Cui,
	Weichen Dai,
	Lipu Zhou,
	Guoquan Huang
	\IEEEcompsocitemizethanks{
		\IEEEcompsocthanksitem Zeyu Wan, Yu Zhang, Bin He and Zhuofan Cui are with State Key Laboratory of Industrial Control Technology, College of Control Science and Engineering, Zhejiang University, Hangzhou, 310027, China. E-mail: \{zeyuwan,zhangyu80,binhe,zhuofancui\}@zju.edu.cn\protect\\
		\IEEEcompsocthanksitem Weichen Dai is with the College of Computer Science, Hangzhou Dianzi University, Hangzhou, China. E-mail: weichendai@hotmail.com\protect\\
		\IEEEcompsocthanksitem Lipu Zhou is with Meituan, 7 Rongda Road, Chaoyang District, Beijing, 100012, China. E-mail: zhoulipu@meituan.com\protect\\
		\IEEEcompsocthanksitem Guoquan Huang is with the Robot Perception and Navigation Group (RPNG), University of Delaware, Newark, DE 19716, USA. E-mail: ghuang@udel.edu.
	}
    \thanks{(Corresponding author: Yu Zhang. Zeyu Wan and Yu Zhang contributed equally to this work.)}
}

\usepackage{cite}
\usepackage{graphicx}
\usepackage{subfigure}
\usepackage{multirow}
\usepackage{multicol}
\usepackage{float}
\usepackage{indentfirst}
\usepackage{amsmath}
\usepackage{amssymb}
\usepackage{amsfonts}
\usepackage{algorithmic}
\usepackage{gensymb}
\usepackage[section]{placeins}
\usepackage[ruled]{algorithm2e}

\newtheorem{theorem}{Theorem}
\newtheorem{lemma}{Lemma}
\newtheorem{proof}{Proof}
\newtheorem{remark}{Remark}

\newtheorem{conjecture}{Conjecture}

\begin{document}
	\markboth{Journal of \LaTeX\ Class Files,~Vol.~14, No.~8, August~2021}
	{Shell \MakeLowercase{\textit{et al.}}: Bare Demo of IEEEtran.cls for Computer Society Journals}
	
	\IEEEtitleabstractindextext{
		\begin{abstract}
			Currently, the improvement of LiDAR poses estimation accuracy is an urgent need for mobile robots. Research indicates that diverse LiDAR points have different influences on the accuracy of pose estimation. This study aimed to select a good point set to enhance accuracy. Accordingly, the sensitivity and uncertainty of LiDAR point residuals were formulated as a fundamental basis for derivation and analysis. High-sensitivity and low -uncertainty point residual terms are preferred to achieve higher pose estimation accuracy. The proposed selection method has been theoretically proven to be capable of achieving a global statistical optimum. It was tested on artificial data and compared with the KITTI benchmark. It was also implemented in LiDAR odometry (LO) and LiDAR inertial odometry (LIO), both indoors and outdoors. The experiments revealed that utilizing selected LiDAR point residuals simultaneously enhances optimization accuracy, decreases residual terms, and guarantees real-time performance.
		\end{abstract}
		
	}
	
	\maketitle
	
    \IEEEraisesectionheading{\section{Introduction}\label{section:Introduction}}

\begin{figure*}[htbp]
    	\centering
    	\includegraphics[width=0.95\textwidth]{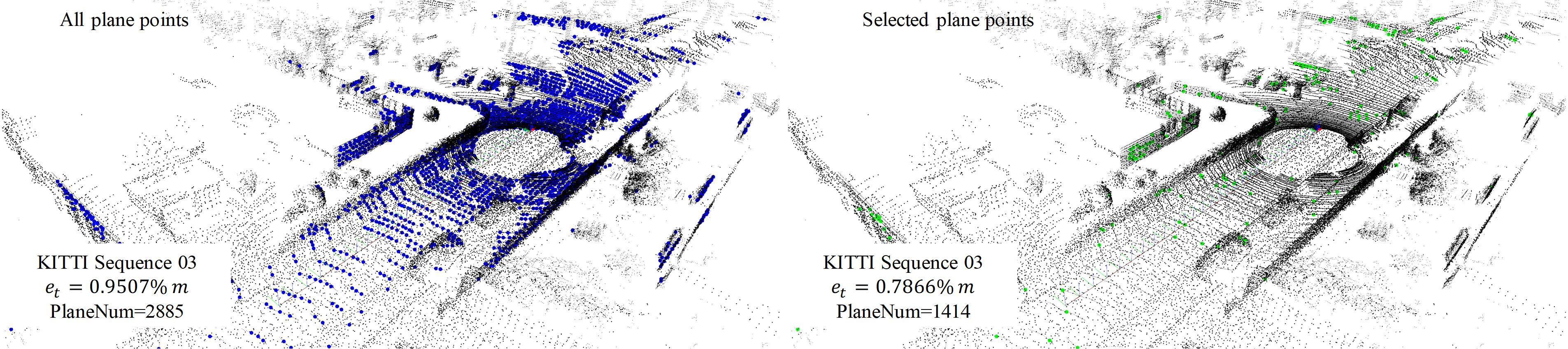}
    	\caption{Original (blue) and our applied selection scheme (green) LO results are shown here. The selected planar points are almost half less than the original, but obtain fewer translation errors simultaneously.}
    	\label{figure:AfterSelected}
    \end{figure*}

    \IEEEPARstart{S}{imultaneous} localization and mapping (SLAM) methods have been applied to solve localization and map-building problems in robotics. LiDAR odometry and local mapping algorithms are widely used in SLAM systems. However, integration processes unavoidably cause an accumulation of pose errors. This drift causes map distortion and estimation failure. Although SLAM is a loop-closing algorithm, it only disperses errors in the history trace, rather than truly eliminating every pose error \cite{2017StateEstimation}. The key to improving the long-term performance relies on the enhancement of front-end accuracy \cite{2016PastPA}. Inspired by our experience with visual odometry (VO) in a dynamic environment \cite{2020PointCorrelations}, diverse feature points are considered owing to their different influences on pose estimation. We presume that diverse LiDAR point residuals have different sensitivities and uncertainties for pose estimation accuracy. Therefore, a novel sensitivity and uncertainty theory that distinguishes residuals from diverse pattern representations was proposed. The theory quantifies the influence of every residual term’s pose estimation accuracy into six dimensions: three for rotation and three for translation. The theory classifies and selects a subset of high sensitivity and low uncertainty, which enters the optimization to achieve a higher accuracy than the utilization of all points. As shown in Fig. \ref{figure:AfterSelected}, the left is the original LO using all valid planar feature points, and the right is adding our selection scheme to this LO. They run on the KITTI benchmark \cite{2012KITTI} sequence 03. The selected planar points were almost half of the original; however, they obtained fewer translation errors simultaneously.
    
\begin{figure}[htbp]
    	\centering
    	\includegraphics[width=0.45\textwidth]{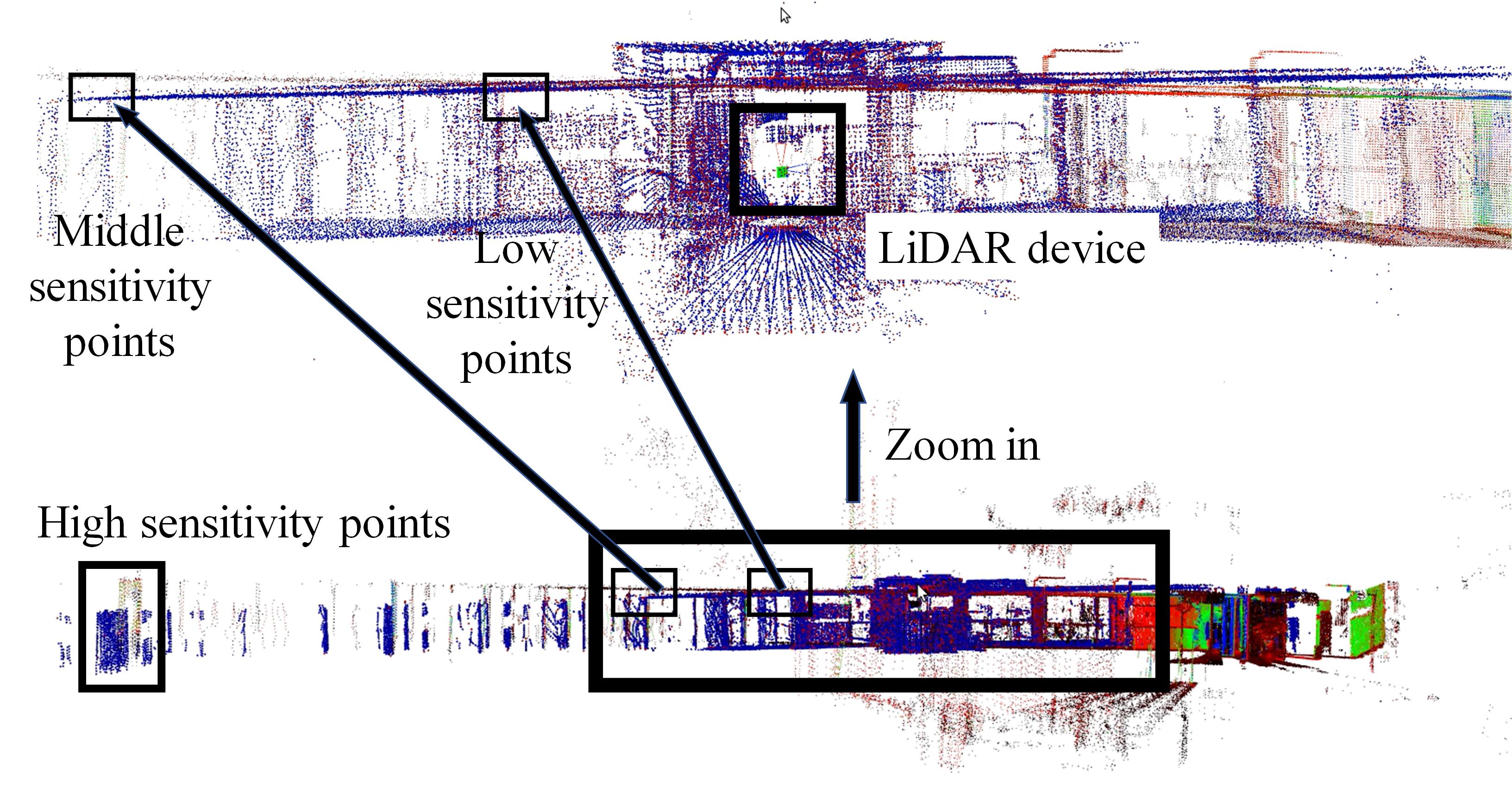}
    	\caption{Sensitivity describes how much a registration residual changes. When there exists a small pose error, higher sensitivity points have more distance errors.}
    	\label{figure:Sensitivity}
    \end{figure}

\begin{figure}[htbp]
    	\centering
    	\includegraphics[width=0.45\textwidth]{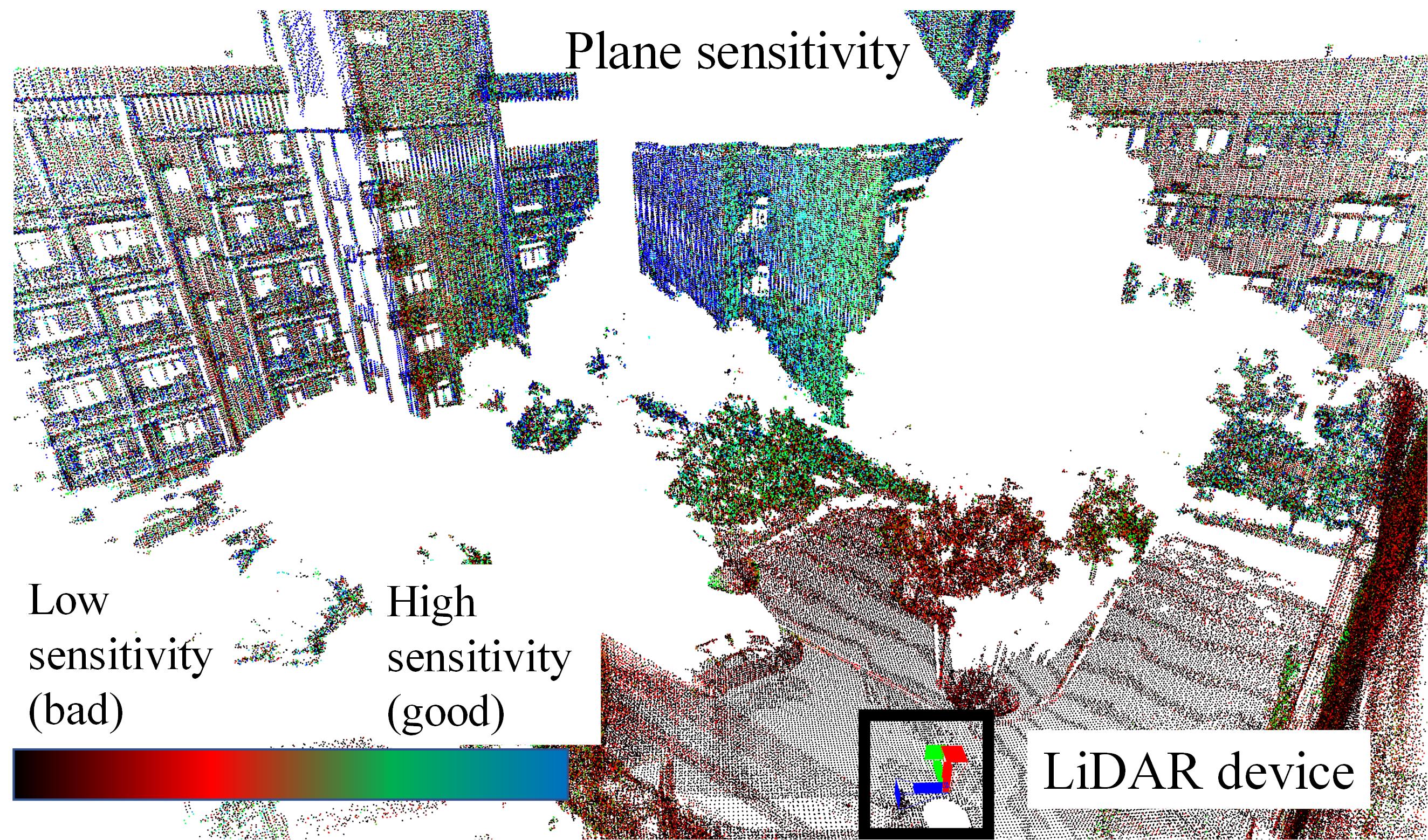}
    	\caption{Every planar point’s sensitivities to the yaw angle are drowned in color. Black and red are low, and green and blue are high. The near-ground points are not sensitive to the yaw angle, and the middle-building walls are more sensitive than the left and right walls.}
    	\label{figure:PlaneSensitivity}
    \end{figure}

    Sensitivity describes the extent to which a registration residual changes when a standard pose disturbance is applied to the sensor. It is defined as a six-dimensional vector: three for rotation angles and others for translation. In Fig. \ref{figure:Sensitivity}, calculating a LiDAR rotation angle, using high sensitivity points is better, which equals the lever principle. In Fig. \ref{figure:PlaneSensitivity}, every planar point’s sensitivities to yaw angle are drowned in color. Black and red were low, while green and blue were high. The near-ground points were not sensitive to the yaw angle, and the middle-building walls were more sensitive than the left and right walls.
    
\begin{figure}[htbp]
    	\centering
    	\includegraphics[width=0.45\textwidth]{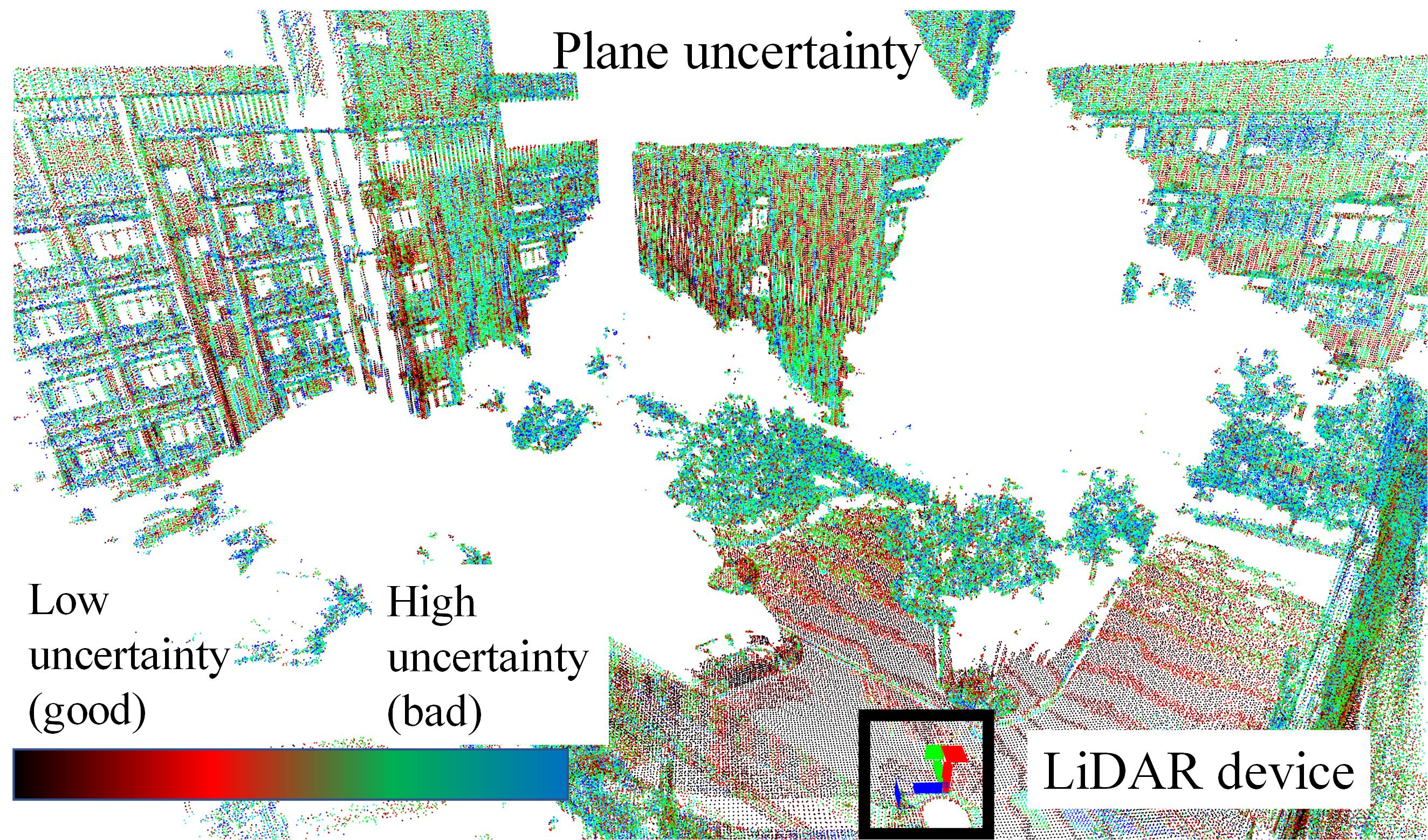}
    	\caption{High uncertainty planar points in blue are trees. The red regions are smooth walls and near-ground points, which are reliable for pose estimation.}
    	\label{figure:PlaneUncertainty}
    \end{figure}

    Uncertainty describes the reliability of a registration residual term that combines a LiDAR point measurement and its corresponding geometric model credibility. It is defined as a three-dimensional Gaussian distribution, such as a line or a plane pattern. In Fig. \ref{figure:PlaneUncertainty}, high uncertainty planar points in blue are trees, which are unsuitable for pose estimation. The red regions represent smooth walls and near-ground points, which are reliable for pose estimation.
    
    This research aims to find calculus approaches for LiDAR point residual sensitivity and uncertainty. We comprehensively considered these two properties in a score vector and decouple them into six dimensions. Thereafter, all LiDAR point residuals were sorted to select a subset that included high sensitivity and low uncertainty points. Finally, these residuals were sent for optimization. In code realization, a threshold rule is defined to stop the selection. Theoretically, we demonstrated that sorting residuals using the proposed method achieves a global statistical optimum. This algorithm is independent of the specific LiDAR SLAM algorithms and LiDAR hardware configurations. It is a general module to enhance accuracy and can be added to any existing optimization-based code realizations. Our experiments on LO and LIO indicate that utilizing selected residuals simultaneously enhances optimization accuracy, decreases residual terms, and guarantees real-time performance.
	
    The main contributions of this study are as follows:
	
	(1) To the best of our knowledge, this is the first study to theoretically prove the global statistical optimal point selection scheme for enhancing pose estimation accuracy in LiDAR odometry and mapping.
	
	(2) This paper proposed a sensitivity model for point-to-plane and point-to-line distance, as well as uncertainty model for LiDAR point measurement and its corresponding geometry pattern. Sensitivity and uncertainty decoupling into six-dimensional methods were also proposed.
	
	(3) Experiments were conducted using the KITTI benchmark. The ALOAM translation error decreased from $1.7318\%$ to $1.5781\%$ in virtually half of the number of planes and lines used. Different types of LiDAR scan modes were evaluated in indoor and outdoor environments using LO and LIO, increasing the average accuracy by approximately $20\%$. Experiments reveal that the time consumption depends on the residual amount rather than the feature detection and residual selection parts. The proposed selection scheme guarantees real-time performance.
	
	\section{Related Work}
	\label{section:Related Work}
	Related studies can be classified into three categories: sensitivity, uncertainty, and entropy-based LO methods.
	
	\subsection{Sensitivity Model}
	This report \cite{2003GeometricallySS} provides considerable motivation for employing the sensitivity model. It proposes a technique for determining whether a pair of meshes is unstable in the iterative closest point (ICP) algorithm. It estimates a covariance matrix from the sparse uniform sampling of the input. Subsequently, it develops a strategy that attempts to minimize this instability and draws a new set of sample points primarily from the stable areas of the input meshes. However, this study concentrates on the registration problem; it does not consider measurement uncertainties and analyzes only the mesh plane errors. However, this technique was fundamental to our theory. LO-Degeneracy \cite{2016OdometryDegeneracy} aims to avoid a degenerate environment, which is regarded as a condition in which one-dimensional sensitivity is zero. It determines and separates the degenerate dimensions in the state space and partially solves the problem in well-conditioned directions. It linearizes the cost function and uses the dot product of the coefficient matrix with its transpose. A matrix containing the geometric structures of the problem constraints is formed. The IMLS \cite{2018IMLSSLAM} technique is a complete LO that uses \cite{2003GeometricallySS} method to select points. Therefore, numerous points can alter the constraints to shrink the final pose. However, it does not solve the problem theoretically and only considers a point-to-plane sensitivity model. The LeGOLOAM algorithm \cite{2018LeGOLOAM} uses normal vector clustering to detect true line points and obtain better matching. Optimization was achieved in two stages using the ground vehicle hypothesis. LeGOLOAM is regarded as improving accuracy from the pattern recognition perspective. However, this is strongly limited by the ground-vehicle hypothesis. Two-stage optimization enables all observations to calculate the rotation and translation separately. LION \cite{2021LION} can self-assess its performance using an observability metric that evaluates whether the pose estimation is geometrically ill-constrained. This is similar to LO-Degeneracy \cite{2016OdometryDegeneracy} and is applied to a real tunnel scene. SGLO \cite{2021SGLO} considers the derivative of the residuals; however, it has not been discussed in depth. It does not consider constraint information in every dimension, which is the core content. MULLS \cite{2021MULLS} clarified the residual linearization process. It uses all observations in the estimations with diverse weights, implying that estimations in different directions can be balanced. In \cite{2020SetCardinalityMax}, inline set cardinality maximization was used to select suitable feature for a 3D-2D pose estimation. Bearing vectors play an important role in the selection and avoidance of degeneration.
	
	From these studies, the proposed theory theoretically clarifies sensitivity. It is inspired by IMLS and extends to a point-to-line residual type. Compared to the MULLS, the point-to-line catches a Hessian matrix in the MULLS, which cannot be sorted directly. The proposed theory uses the main direction projection to regroup into a linear form, which is convenient for sorting.
	
	\subsection{Uncertainty Model}
	The uncertainty model consisted of two parts. The first part independently models the uncertainty of every laser point measurement in 3D and is referred to as the laser scan beam. The second part models the uncertainty of the geometric pattern in the map. The essential difference between these two models is that the first part describes the uncertainty of the current observations and the second part describes the uncertainty of history-measured information.
	
	\subsubsection{Laser Scan Beam}
	For laser points, \cite{2007LidarError} proposed rigorous first-order error analysis. It measures the horizontal and vertical errors of a laser pulse and determines the nonlinear error growth, as recently reported in \cite{2020LidarComparing}. Comparing various LiDAR sensors available in the market \cite{2020LidarComparing}, measurement errors were found to be relevant to the target range. In \cite{2007AirLaser}, a laser point was modeled as a projected footprint and used to represent an uncertainty matrix. A 3D Gaussian distribution was proposed \cite{2018VLP16Model} to model LiDAR uncertainty points and to clarify their propagation.
	
	\subsubsection{Geometry Pattern}
	For geometry patterns, point cloud data (PCD) are direct and easy to use for localization. Accordingly, this investigation focuses on the map geometry pattern implemented using PCD. The LOAM algorithm \cite{2014LOAM} generates five points to simulate a plane and  line by decoupling the eigenvalues. It fundamentally calculates them as a 3D Gaussian distribution but discards irrelevant directions. The Gaussian mixture model \cite{2016GMM} (GMM) is a continuous distribution function method; however, it adopts multiple Gaussians and regroups them with different weights. A multilayer tree structure \cite{2018GMMtree} can fuse flat areas into one Gaussian or decompose a complicated area into several Gaussians. In addition to these explicit function representations, implicit methods are relevant to PCD applications. A moving least square (MLS) surface is defined in \cite{2004MLS}; it is a $C^\infty$ smooth surface generated from a raw PCD. An implicit version was defined in \cite{2005ProvablyMLS}, which represents the distance of a location on a surface composed of neighboring points.
	
	Based on the aforementioned studies, real LiDAR emitting and receiving structures were considered to build a laser point uncertainty model. Thereafter, the measured points were added to the map and fused together. Our study aimed to improve the real-time estimation accuracy of mobile robots. Compared with a 3D reconstruction, sacrificing some of the complicated area details and focusing on the main direction constraints are advantageous for SLAM. Because nearby LiDAR points are dense, the remote points are sparse. The leaves, trees, and other irregular objects are not suitable for pose estimation, indicating that their fused uncertainties are higher. Therefore, the Gaussian method is preferred, and the uncertainties of the map points are considered. Consequently, these points are completely used in a plane or line.
	
	\subsection{Entropy Based LO Method}
	Since 2020, some research has applied the entropy concept from information theory \cite{Shannon1948Entropy} to SLAM systems to improve robustness and accuracy. \cite{2020GoodFeatureMatching} proposed sub-matrix selection by choosing a scoring metric for VO. It models estimation as a linear matrix to obtain the best subset that yields the metric Max-$log$Det. With this metric, satisfactory feature selection becomes an NP-hard problem. They designed a lazy greedy algorithm to determine the maximum submatrix. In a continuation of \cite{2020GoodGraphOptimize}, it focused on selecting satisfactory poses for graph optimization. In \cite{2020InformationDrivenDirectRGBD}, the most numerous mutual information points were selected, and the metrics were similar. MLOAM \cite{2021ICRAmloam} imitates and applies entropy to a multi-LiDAR field. The two LiDAR sensors were set at diverse angles to cover a wide area. In addition, a greedy method was designed. The two LiDARs were run in real time with satisfactory accuracy.
	
	Compared to the aforementioned studies, the most novel contribution of our proposed theory is that we provide an analytical demonstration of why the selected points obtain higher accuracy. Another advantage is that the proposed theory is endogenously explainable, which is derived from a singular value decomposition (SVD)-based registration problem \cite{1992ICP}. Finally, the sensitivity and uncertainty processes are modeled in a linear form, avoiding the calculation of a sub-matrix metric using a greedy method.
	
	\section{Notations and Preliminaries}
	\label{section:Notations and Preliminaries}
	Before introducing our theory, the interpretation of the ICP registration problem aids in understanding the theory. The SVD method was used as a standard solution. Zero-mean normalization is applied to decouple it into calculating the rotation (first step) and translation (second step). That is, it solves $\mathbf{R}$ in $\rm{SO}(3)$ space and then returns to $\rm{SE}(3)$ space to calculate $\mathbf{t}$. This is known as the Wahba problem \cite{Wahba1965} since 1965, or rotation search \cite{2004RotationSearch} in the recent robotic community.
	
	Assume that a no-disturbance point set (source) is $P=\{\mathbf{p}^*_i\},i=1,\ldots,N$ and its corresponding no-disturbance point set (target) is $Q=\{\mathbf{q}^*_i\},i=1,\ldots,N$. The standard $L_2$ norm point-to-point registration problem is expressed as follows:
	\begin{equation}
		\label{equation:ICP}
		\begin{aligned}
			\mathbf{R}^* & =\mathop{\arg\min}\limits_{\mathbf{R}^*\in\rm{SO}(3)}\sum_{i=1}^N ||\mathbf{R}^*\mathbf{p}^*_i-\mathbf{q}^*_i||^2           \\
			& =\mathop{\arg\max}\limits_{\mathbf{R}^*\in\rm{SO}(3)} tr(\mathbf{R}^*\sum_{i=1}^N \mathbf{p}^*_i {\mathbf{q}^*_i}^T)
		\end{aligned}
	\end{equation}
	\begin{equation}
		\label{equation:ICPSVD}
		\mathbf{H}=\sum_{i=1}^N \mathbf{p}^*_i {\mathbf{q}^*_i}^T=\mathbf{U\Sigma V}^T
	\end{equation}
	\begin{equation}
		\label{equation:ICPSVD2}
		\mathbf{R}^*=\mathbf{VU}^T
	\end{equation}
	where $\mathbf{R}$ is a $3\times3$ rotation matrix and the proof of Eq. (\ref{equation:ICP}) is provided in Appendix A. By linearizing the rotation parameters from the Lie group manifold to its corresponding location in the tangent vector space (i.e., $\phi_\mathbf{R}^\wedge\in\mathfrak{so}(3),\phi_\mathbf{R}\in\mathbb R^3$ Lie algebra), the problem becomes a linear least-squares problem. Accordingly, it is solved using the Gauss-Newton\cite{1974GN} or Levenberg-Marquardt\cite{1977LM} techniques. The optimal rotation is found as the singular vectors $\mathbf{V}$ and $\mathbf{U}$ regroups of $\mathbf{H}$. 
	
\begin{figure}[htbp]
		\centering
		\includegraphics[width=0.45\textwidth]{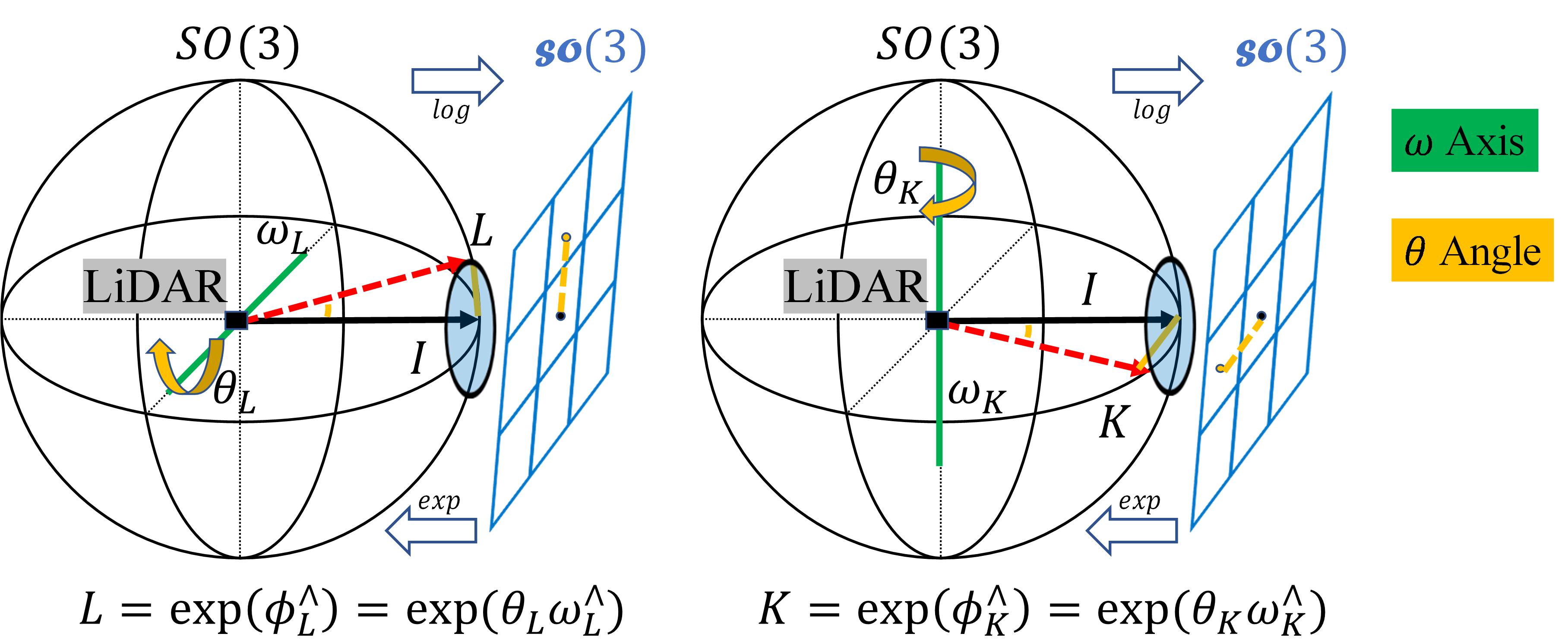}
		\caption{Axis angle (rotation vector) representation is $\phi_\mathbf{L}=\theta_\mathbf{L}\omega_\mathbf{L}$ and $\mathbf{L}=\exp(\phi_\mathbf{L}^\wedge)=\exp(\theta_\mathbf{L}\omega_\mathbf{L}^\wedge)$. The black sphere is the $\rm{SO}(3)$ space Lie group, and the blue grid is the $\mathfrak{so}(3)$ space Lie algebra, which is tangent to the expansion location. $\exp$ and $\log$ mappings connect each other. This enable gradient descent-based optimization algorithms to work.}
		\label{figure:AngleAxis}
	\end{figure}

	Any two rotation matrices, $\mathbf{L}$ and $\mathbf{K}$, and their corresponding axis angles (rotation vector), $\phi_\mathbf{L}=\theta_\mathbf{L}\omega_\mathbf{L}$ and $\phi_\mathbf{K}=\theta_\mathbf{K}\omega_\mathbf{K}$, as shown in Fig. \ref{figure:AngleAxis}, are connected by the exponential map from the Lie group to the Lie algebra.
	\begin{equation}
		\label{equation:Angle Axis}
		\begin{aligned}
			\mathbf{L} & =\exp(\phi_\mathbf{L}^\wedge)=\exp(\theta_\mathbf{L}\omega_\mathbf{L}^\wedge) \\
			\mathbf{K} & =\exp(\phi_\mathbf{K}^\wedge)=\exp(\theta_\mathbf{K}\omega_\mathbf{K}^\wedge)
		\end{aligned}
	\end{equation}
	$\theta_\mathbf{L}$ and $\theta_\mathbf{K}$ are the angles (scalar), $\omega_\mathbf{L}$ and $\omega_\mathbf{K}$ are axes ($3\times1$ vector), $\phi^\wedge$ denotes the symmetric skew  matrix of vector $\phi$.
	
	skew-symmetric matrix and corresponding vector satisfying
	\begin{equation}
		\mathbf{a}^\wedge=
		\begin{bmatrix}
			a_1\\
			a_2\\
			a_3
		\end{bmatrix}^\wedge=
		\begin{bmatrix}
			0    & -a_3 & a_2  \\
			a_3  & 0    & -a_1 \\
			-a_2 & a_1  & 0
		\end{bmatrix}
	\end{equation}
	
\begin{figure}[htbp]
		\centering
		\includegraphics[width=0.15\textwidth]{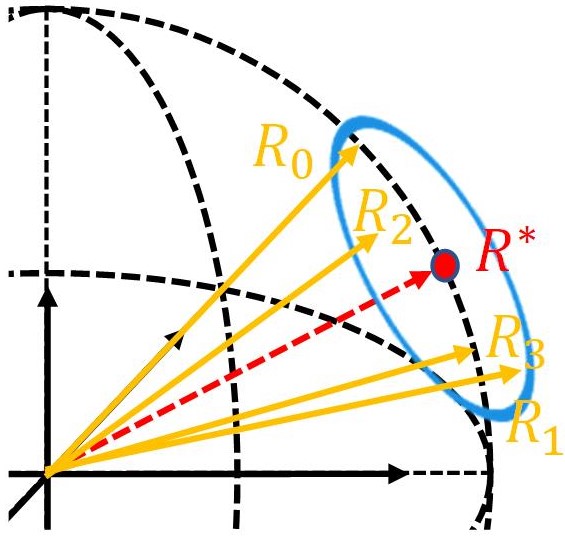}
		\caption{When sensors provide redundant observations, there exist many estimated rotation matrices $\mathbf{R}_i$. The method of calculating the smallest fitting error rotation matrix $\mathbf{R}^*$ is a fundamental optimization problem. This phenomenon is common in a LO or LIO, which requires the sum of Riemannian distances to be the smallest. Specifically, if $\mathbf{R}^*$ is the optimal, it must be close to all valid $\mathbf{R}_i$.}
		\label{figure:RiemannianDistance}
	\end{figure}

	To measure the difference between the two rotation matrices, $\mathbf{R}^*$ and $\mathbf{R}_i$ in Fig. \ref{figure:RiemannianDistance}, the Riemannian metric distance \cite{2004RotationSearch} under the Frobenius Norm is used. It is defined as
	\begin{equation}
		\label{equation:Riemanniandistance}
		\begin{aligned}
			Riem(\mathbf{R}^*,\mathbf{R}_i) & =||\log({\mathbf{R}^*}^T \mathbf{R}_i)||^2_F        \\
			& =||\log(\Delta\mathbf{R})||^2_F                     \\
			& =Riem(\Delta\mathbf{R})                             \\
			& =||\log(\exp(\phi^\wedge_{\Delta\mathbf{R}}))||^2_F \\
			& =tr({\phi^\wedge_{\Delta\mathbf{R}}}^T\phi^\wedge_{\Delta\mathbf{R}})
		\end{aligned}
	\end{equation}
	that satisfies $\Delta\mathbf{R}={\mathbf{R}^*}^T \mathbf{R}_i$. This distance is the length of the shortest geodesic curve connecting the two rotation matrices. The defined symbols are summarized in TABLE \ref{table:Basic Notations}.
	\begin{table}[htbp]
		\begin{center}
			\caption{Basic notations}
			\renewcommand\arraystretch{1.4}
			\resizebox{0.5\textwidth}{!}
			{
				\begin{tabular}{|c|c|c|c|}
					\hline
					$\mathbf{R}$,$\mathbf{L}$,$\mathbf{K}$ & rotation matrix &      $\phi$      & angle axis (rotation vector) \\ \hline
					$\theta$                &      angle      &     $\omega$     &             axis             \\ \hline
					$\mathbf{p}^*_i$            &  source point   & $\mathbf{q}^*_i$ &         target point         \\ \hline
				\end{tabular}
			}
			\label{table:Basic Notations}
		\end{center}
	\end{table}
	
	\section{Sensitivity and Uncertainty}
	\label{section:Sensitivity and Uncertainty}
	This section proves that the proposed sensitivity and uncertainty theory-based selection scheme achieves the global statistical optimal pose estimation accuracy. The optimum is statistical because the analyses are based on probability distributions. In one-shot sampling, a point with a higher uncertainty may be more accurate than other points because the uncertainty is only a probability distribution description. Therefore, multisampling and calculated expectations are superior.
	
	For brevity, we simplified the enhancing accuracy issue as a point registration problem in $\rm{SO}(3)$. Assuming the ideal condition, there exists a ground truth rotation $\mathbf{R}^*$ that satisfies the no-disturbance condition for $N\geq 4$ points:
	\begin{equation}
		\label{equation:RotationStar}
		\mathbf{R}^*\mathbf{p}^*_i=\mathbf{q}^*_i, i=1,\cdots,N
	\end{equation}
	
\begin{figure}[htbp]
		\centering
		\includegraphics[width=0.45\textwidth]{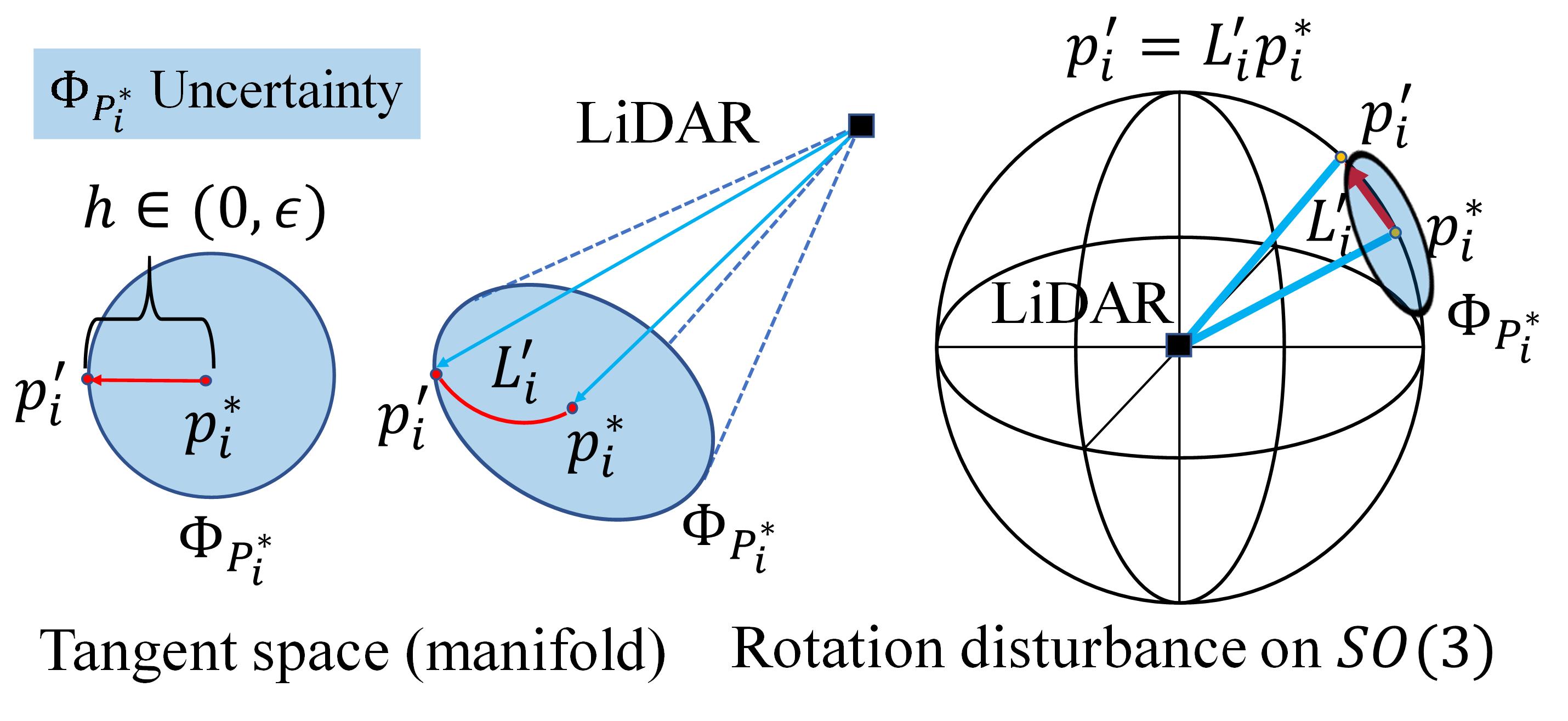}
		\caption {Small disturbance on $\rm{SO}(3)$ can be described as a small rotation matrix $\mathbf{L}^\prime_i$, which equals to a circle uniform distribution whose radius is $h\in(0,\epsilon)$. $h$ is a scalar variable, and $\epsilon$ is a given distance to the maximal far location.}
		\label{figure:DemonstrationUncertaintyModel}
	\end{figure}

	Consider small disturbances on source points $\mathbf{p}^*_i$ as real LiDAR noise, which may occur during capturing or algorithm drift. On $\rm{SO}(3)$, a dynamic rotation matrix $\mathbf{L}^\prime_i$ describes this error, $\mathbf{p}^\prime_i=\mathbf{L}^\prime_i \mathbf{p}^*_i$, which lies on the 3D manifold surface in Fig. \ref{figure:DemonstrationUncertaintyModel}. The same disturbance $\mathbf{K}^\prime_i$ was added to the target point $\mathbf{q}^*_i$. We defined this simple uncertainty model as the only one for the inference. The uncertainty model details are described in Lemmas \ref{lemma:Uncertainty} and Fig. \ref{figure:DemonstrationUncertaintyModel}, which do not hinder understanding at present.
	
	Assuming the selection $M$ point pairs in a certain rule, $M$ depends on the number of points participating in the rotation calculation, that is, at least $4$ points. Therefore, $M$ controls the algorithm to cover all the possible conditions, satisfying $4\leq M\leq N$. The optimal rotation, $\mathbf{R}^\prime$, was calculated using Eq. (\ref{equation:ICP}):
	\begin{equation}
		\label{equation:RotationPrime1}
		\mathbf{R}^\prime=\mathop{\arg\min}\limits_{\mathbf{R}^\prime\in\rm{SO}(3)}\sum_{j=1}^M||\mathbf{R}^\prime\mathbf{p}^\prime_j-\mathbf{q}^\prime_j||^2
	\end{equation}
	If this selection scheme is optimal, the Riemannian distance between $\mathbf{R}^*$ and $\mathbf{R}^\prime$ is very close. Specifically, by exchanging an arbitrary one-point pair inside $M$ with that outside of $N-M$, the optimal rotation is $\mathbf{R}^{\prime\prime}$.
	\begin{equation}
		\label{equation:RotationPrime2}
		\mathbf{R}^{\prime\prime}=\mathop{\arg\min}\limits_{\mathbf{R}^{\prime\prime}\in\rm{SO}(3)}\sum_{j=1}^M||\mathbf{R}^{\prime\prime}\mathbf{p}^{\prime\prime}_j-\mathbf{q}^{\prime\prime}_j||^2
	\end{equation}
	where $\{\mathbf{p}^\prime\}$, $\{\mathbf{q}^\prime\}$ and $\{\mathbf{p}^{\prime\prime}\}$, $\{\mathbf{q}^{\prime\prime}\}$ are at least one point pair that are different. A conjecture must then be established to illustrate why the selection scheme is optimal.
	\begin{conjecture}[\bf{Closest Riemannian Distance}]
		\label{conjecture:Closest Riemannian Distance}
		\begin{equation}
			\label{equation:DistanceRotationPrime12}
			Riem(\mathbf{R}^*,\mathbf{R}^\prime)<Riem(\mathbf{R}^*,\mathbf{R}^{\prime\prime})
		\end{equation}
	\end{conjecture}
	
\begin{figure}[htbp]
		\centering
		\includegraphics[width=0.5\textwidth]{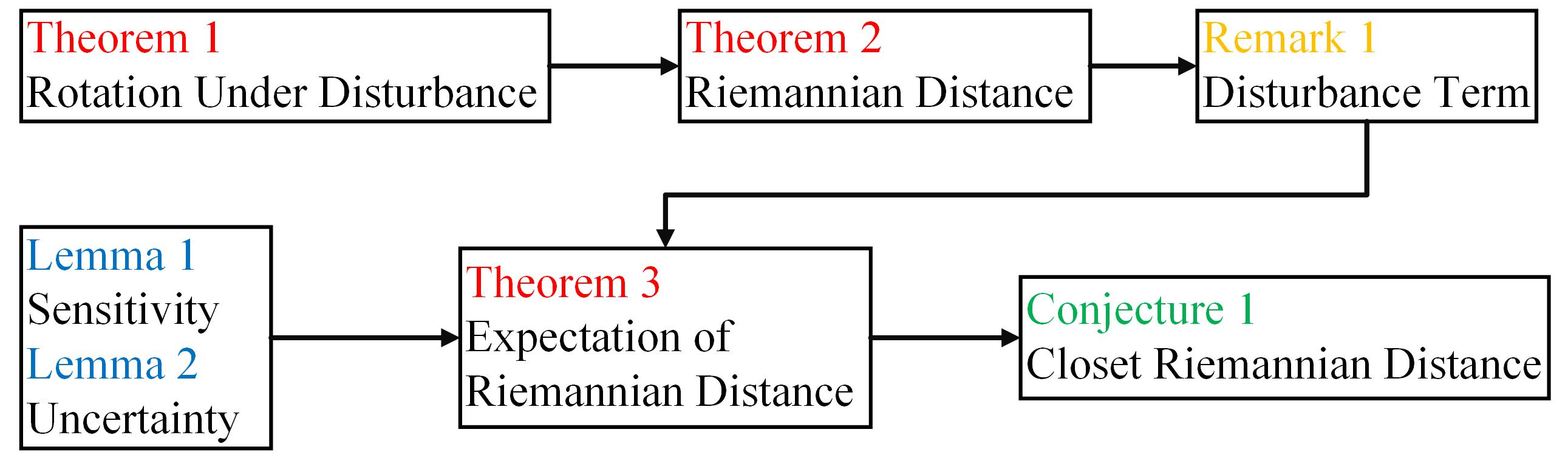}
		\caption{Outline of the demonstration process.}
		\label{figure:Demonstration Pipeline}
	\end{figure}

	The remainder of this section demonstrates Conjecture \ref{conjecture:Closest Riemannian Distance}, and the entire demonstration process is shown in Fig. \ref{figure:Demonstration Pipeline}. The defined symbols are summarized in TABLE \ref{table:Sensitivity Uncertainty Notations}
	\begin{table}[htbp]
		\begin{center}
			\caption{Theory notations}
			\renewcommand\arraystretch{1.8}
			\resizebox{0.5\textwidth}{!}
			{
				\begin{tabular}{|c|c|c|c|}
					\hline
					$\mathbf{R}^*$                 &            ideal rotation            &      $\mathbf{R}^\prime$      &           optimal rotation           \\ \hline
					$\mathbf{R}^{\prime\prime}$          &                      \multicolumn{3}{c|}{after point pair exchanged optimal rotation}                       \\ \hline
					$\mathbf{p}^*_i$                &          ideal source point          &       $\mathbf{q}^*_i$        &          ideal target point          \\ \hline
					$\mathbf{p}^\prime_i$             &    source point after disturbance    &     $\mathbf{q}^\prime_i$     &    target point after disturbance    \\ \hline
					$\mathbf{p}^{\prime\prime}_i$         &                        \multicolumn{3}{c|}{after point pair exchanging source point}                         \\ \hline
					$\mathbf{q}^{\prime\prime}_i$         &                        \multicolumn{3}{c|}{after point pair exchanging target point}                         \\ \hline
					$\mathbf{L}^\prime_i$             & rotation disturbance on source point &     $\mathbf{K}^\prime_i$     & rotation disturbance on target point \\ \hline
					$\Phi_{\mathbf{p}^*_i}$            &             uncertainty              & $\mathbf{J}_{\mathbf{p}^*_i}$ &             sensitivity              \\ \hline
					$\mathbf{A},\mathbf{B},\mathbf{C},\mathbf{D}$ &         \multicolumn{3}{c|}{the first, second, third, fourth term of Riemannian distance expansion}         \\ \hline
				\end{tabular}
			}
			\label{table:Sensitivity Uncertainty Notations}
		\end{center}
	\end{table}
	
	Initially, the relationship between $\mathbf{R}^*$ and $\mathbf{R}^\prime$ should be established in Theorem \ref{theorem:Rotation Under Disturbance}, and then their distances are calculated. This is practicable for analyzing the point disturbance influence on the pose estimation result.
	
	\begin{theorem}[\bf{Rotation Under Disturbance}]
		\label{theorem:Rotation Under Disturbance}
		Assuming the addition of disturbances to the source  and target points, we can establish
		\begin{equation}
			\label{equation:Rotation Under Disturbance}
			\mathbf{R}^\prime=\mathbf{K}^\prime \mathbf{R}^* \mathbf{L}^{\prime T}
		\end{equation}
	\end{theorem}
	
	\begin{proof}[\bf{Rotation Under Disturbance}]
		By substituting the disturbances into Eq. (\ref{equation:ICPSVD}) and (\ref{equation:ICPSVD2}), respectively,
		\begin{equation}
			\begin{aligned}
				\mathbf{H}^\prime & =\sum_{i=0}^N \mathbf{p}_i^\prime \mathbf{q}_i^{\prime T}                               \\
				& =\sum_{i=0}^N \mathbf{L}^\prime \mathbf{p}^*_i \mathbf{q}_i^{* T} \mathbf{K}^{\prime T} \\
				& =\mathbf{L}^\prime \mathbf{H}^* \mathbf{K}^{\prime T}                                   \\
				& =(\mathbf{L}^\prime \mathbf{U})\mathbf{\Sigma}(\mathbf{K}^\prime \mathbf{V})^T
			\end{aligned}
		\end{equation}
		Thus, the optimal rotation is
		\begin{equation}
			\begin{aligned}
				\mathbf{R}^\prime=\mathbf{K}^\prime \mathbf{V}(\mathbf{L}^\prime \mathbf{U})^T=\mathbf{K}^\prime \mathbf{VU}^T \mathbf{L}^{\prime T}=\mathbf{K}^\prime \mathbf{R}^* \mathbf{L}^{\prime T}
			\end{aligned}
		\end{equation}
		\begin{center}
			\bf{Q.E.D.}
		\end{center}
	\end{proof}
	
	Next, the Riemannian distance between $\mathbf{R}^*$ and $\mathbf{R}^\prime$ is expanded. Theorem \ref{theorem:Riemannian Distance} aims to identify the elements that affect the result exactly. Thus, the definition of sensitivity and uncertainty properties is motivated here.
	
	\begin{theorem}[\bf{Riemannian Distance}]
		\label{theorem:Riemannian Distance}
		\begin{equation}
			\label{equation:Riemannian Distance}
			\begin{aligned}
				Riem(\mathbf{R}^*,\mathbf{R}^\prime) & =||\log({\mathbf{R}^*}^T\mathbf{R}^\prime)||_F^2\\
				&=\mathbf{A}+\mathbf{B}+\mathbf{C}+\mathbf{D}
			\end{aligned}
		\end{equation}
		the four terms that satisfy
		\begin{equation}
			\label{equation:Riemannian Distance ABCD}
			\begin{aligned}
				\mathbf{A} & =2{(\mathbf{R}^*\phi_{\mathbf{K}^\prime})}^T(\mathbf{R}^*\phi_{\mathbf{K}^\prime})                                                                             \\
				\mathbf{B} & =2{\phi_{\mathbf{L}^\prime}}^T\phi_{\mathbf{L}^\prime}                                                                                                         \\
				\mathbf{C} & =\frac{1}{2}{[(\mathbf{R}^*\phi_{\mathbf{K}^\prime})^\wedge\phi_{\mathbf{L}^\prime}]}^T (\mathbf{R}^*\phi_{\mathbf{K}^\prime})^\wedge\phi_{\mathbf{L}^\prime} \\
				\mathbf{D} & =4{(\mathbf{R}^*\phi_{\mathbf{K}^\prime})}^T\phi_{\mathbf{L}^\prime}
			\end{aligned}
		\end{equation}
	\end{theorem}
	
	\begin{proof}[\bf{Riemannian Distance}]
		The proof of Theorem \ref{theorem:Riemannian Distance} is provided in Appendix B. The core is the utilization of the Baker (Campbell) Hausdorff formula \cite{1974BCH} and exponential mapping expansion. Thereafter, we acquired six terms, and the last two terms are always zero.
	\end{proof}
	
	Note that, in Theorem \ref{theorem:Riemannian Distance}, $\mathbf{A}$ and $\mathbf{B}$ have additional concise representations. Therefore, we have written Remark \ref{remark:Disturbance Terms} to illustrate this for better comprehension.
	
	\begin{remark}[\bf{Disturbance Terms}]
		\label{remark:Disturbance Terms}
		Based on Theorem \ref{theorem:Riemannian Distance} and the angle-axis representation in Fig. \ref{figure:AngleAxis}
		\begin{equation}
			\label{equation:Disturbance Terms A B}
			\begin{aligned}
				\frac{\mathbf{A}}{2} & =(\mathbf{R}^* \phi_{\mathbf{K}^\prime})^T(\mathbf{R}^* \phi_{\mathbf{K}^\prime})=\theta_{\mathbf{K}^\prime}^2 \\
				\frac{\mathbf{B}}{2} & =\phi_{\mathbf{L}^\prime}^T \phi_{\mathbf{L}^\prime}=\theta_{\mathbf{L}^\prime}^2
			\end{aligned}
		\end{equation}
	\end{remark}
	$\mathbf{C}$ and $\mathbf{D}$ are dynamic, depending on the included angle of these two disturbance rotation vectors.
	
	Subsequently, the dynamic properties of the last two terms, $\mathbf{C}$ and $\mathbf{D}$ can be clarified from another perspective. In one-shot sampling, a point is captured at a specific position where it must be located. Although it belongs to a specific distribution, the one-shot sampling is random. This phenomenon results in Eq. (\ref{equation:DistanceRotationPrime12}) in Conjecture \ref{conjecture:Closest Riemannian Distance} is impossible. Therefore, we considered comparing the expectations of the Riemannian distance to solve this problem. In the next theorem, after double integration throughout the disturbance space, $\mathbf{C}$ is only related to $\theta_{\mathbf{K}^\prime}\theta_{\mathbf{L}^\prime}$ and $\mathbf{D}$ is zero. Eventually, the comparison continued.
	
	\begin{theorem}[\bf{Expectation of Riemannian Distance}]
		\label{theorem:Expectation of Riemannian Distance}
		The expectation of the Riemannian distance is
		\begin{equation}
			\label{equation:Expectation of Riemannian Distance}
			E(Riem(\mathbf{R}^*,\mathbf{R}^\prime))=2\theta_{\mathbf{K}^\prime}^2+2\theta_{\mathbf{L}^\prime}^2+\frac{\theta^2_{\mathbf{K}^\prime}\theta^2_{\mathbf{L}^\prime}}{4}
		\end{equation}
	\end{theorem}
	
	\begin{proof}[\bf{Expectation of Riemannian Distance}]
		The proof of Theorem \ref{theorem:Expectation of Riemannian Distance} is provided in Appendix C.
	\end{proof}
	
	Sensitivity and uncertainty have been clear perspicuities. They are defined from Theorem \ref{theorem:Expectation of Riemannian Distance} with $\theta_{\mathbf{K}^\prime}$ and $\theta_{\mathbf{L}^\prime}$. We write these two properties in Lemmas \ref{lemma:Sensitivity} and \ref{lemma:Uncertainty}.
	
\begin{figure}[htbp]
		\centering
		\includegraphics[width=0.25\textwidth]{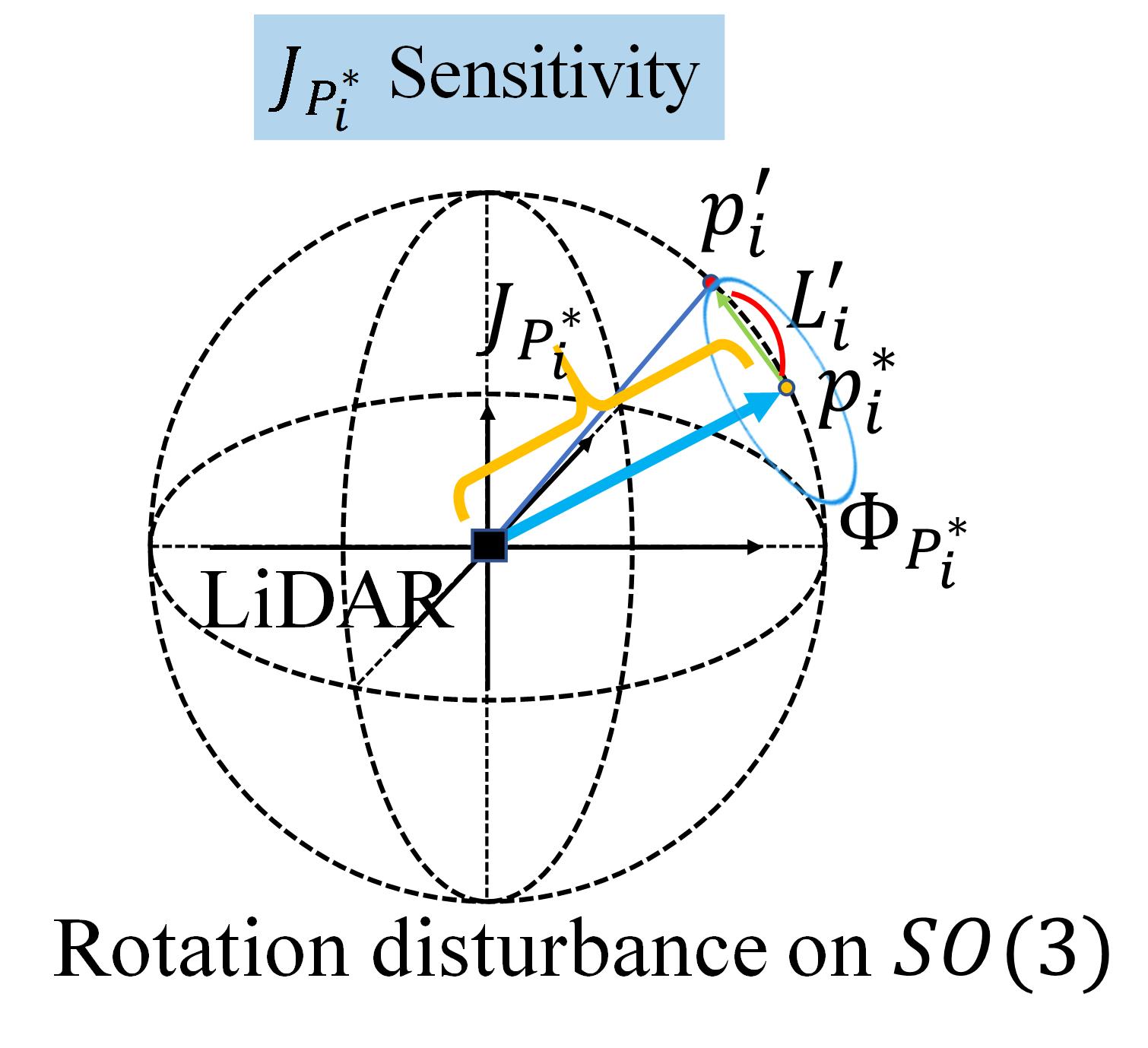}
		\caption{Points that are distant from the center of a LiDAR sensor, undergo more changes when the same rotation is applied to them. Thus, in point-to-point registration, sensitivity is defined by a point's norm.}
		\label{figure:DemonstrationSensitivityModel}
	\end{figure}

	\begin{lemma}[\bf{Sensitivity}]
		\label{lemma:Sensitivity}
		Points that are distant from the center of a LiDAR sensor undergo more changes when the same rotation is applied. Thus, in point-to-point registration, the sensitivity is defined by a point's norm. The point (i.e., $\mathbf{p}^*_i$) sensitivity in Fig. \ref{figure:DemonstrationSensitivityModel} is defined by
		\begin{equation}
			\label{equation:Sensitivity}
			\mathbf{J}_{\mathbf{p}^*_i}=||\mathbf{p}^*_i||^2
		\end{equation}
	\end{lemma}
	
	\begin{lemma}[\bf{Uncertainty}]
		\label{lemma:Uncertainty}
		A small disturbance on $\rm{SO}(3)$ can be described as a small rotation matrix $\mathbf{L}^\prime_i$, which is equal to a circular uniform distribution with radius $h\in(0,\epsilon)$. $h$ is a scalar variable and $\epsilon$ is the distance to the maximal far location. The point (i.e., $\mathbf{p}^*_i$) uncertainty in Fig. \ref{figure:DemonstrationUncertaintyModel} is defined by
		\begin{equation}
			\label{equation:Uncertainty}
			\Phi_{\mathbf{p}^*_i}=\frac{\epsilon}{2}
		\end{equation}
		because $h^2$ indicates the disturbance amplitude, and its expectation integration $E$ is
		\begin{equation}
			\Phi_{\mathbf{p}^*_i}=E(h^2)=\frac{1}{\epsilon}\int_{0}^{\epsilon}\frac{1}{2\pi h}\int_{0}^{2\pi}h^2 d\alpha dh=\frac{\epsilon}{2}
		\end{equation}
		where $\alpha$ denotes the round integration of $\mathbf{p}^\prime$.
	\end{lemma}
	
	Finally, by combining Lemma \ref{lemma:Sensitivity}, Lemma \ref{lemma:Uncertainty} and Theorem \ref{theorem:Expectation of Riemannian Distance}, Conjecture \ref{conjecture:Closest Riemannian Distance} can be proven.
	
	\begin{proof}[\bf{Conjecture \ref{conjecture:Closest Riemannian Distance} Closest Riemannian Distance}]
		As shown in Fig. \ref{figure:TangentAngleAxis}, for a specific point pair (i.e., index $i$), according to the law of sines, the relationship between sensitivity and uncertainty is established.
		\begin{figure}[htbp]
			\centering
			\includegraphics[width=0.2\textwidth]{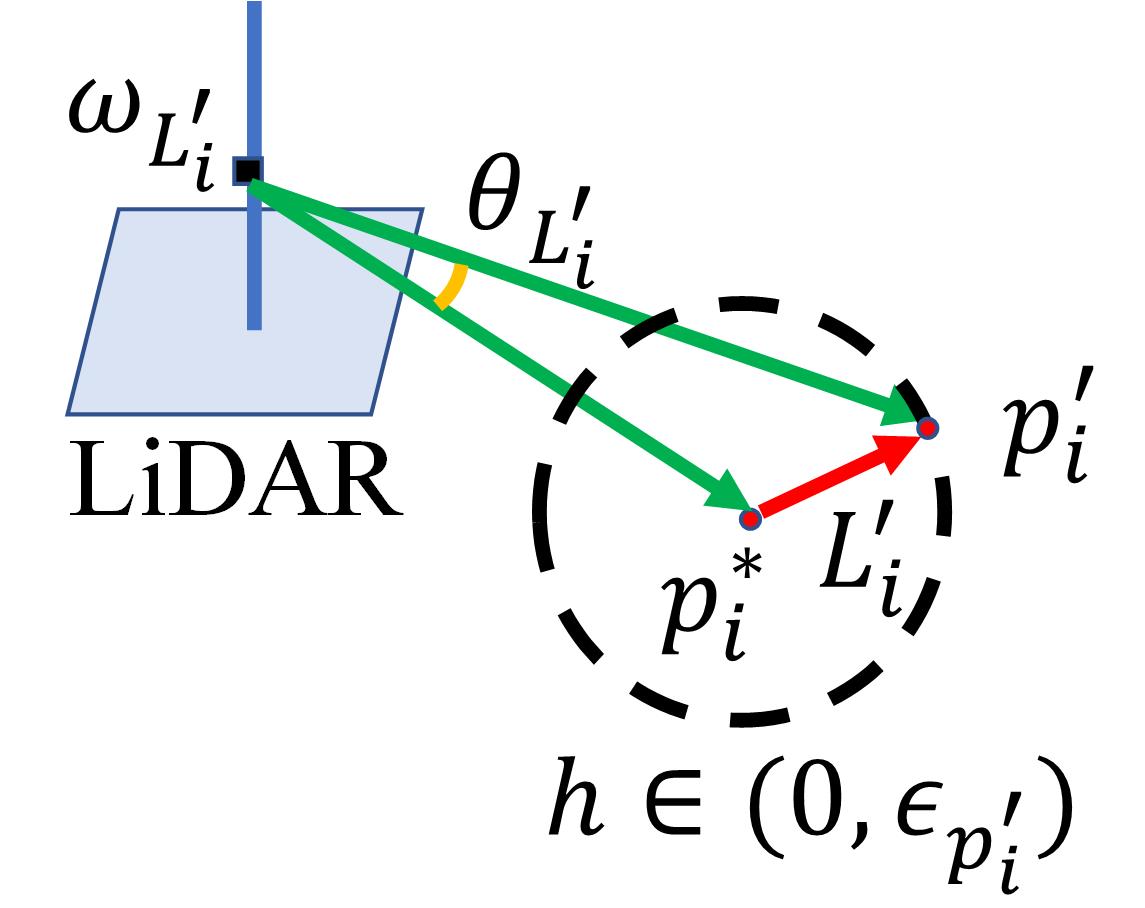}
			\caption{According to the law of sines, the relationship between sensitivity and uncertainty is established.}
			\label{figure:TangentAngleAxis}
		\end{figure}
		There exists
		\begin{equation}
			\label{equation:Point Contribution Model 5}
			\begin{aligned}
				\theta_{\mathbf{L}^\prime_i}&=arcsin(\frac{\epsilon_{\mathbf{p}^\prime_i}}{\sqrt{||\mathbf{p}^\prime_i||^2}})\propto\frac{\Phi_{\mathbf{p}^\prime_i}^2}{\mathbf J_{\mathbf{p}^\prime_i}}\\
				\theta_{\mathbf{K}^\prime_i}&=arcsin(\frac{\epsilon_{\mathbf{q}^\prime_i}}{\sqrt{||\mathbf{q}^\prime_i||^2}})\propto\frac{\Phi_{\mathbf{q}^\prime_i}^2}{\mathbf J_{\mathbf{q}^\prime_i}}
			\end{aligned}
		\end{equation}
		When $M$ point pairs have diverse disturbances, $\mathbf{R}^\prime$ and $\mathbf{R}^{\prime\prime}$ are dynamic owing to the specific disturbances. Fortunately, by solving Eq. (\ref{equation:RotationPrime1}) ($\mathbf{R}^\prime$) and (\ref{equation:RotationPrime2}) ($\mathbf{R}^{\prime\prime}$) are based on Lie algebra, a linear space. Similar to the rotation search in Fig. \ref{figure:RiemannianDistance}, this linear property means argument every $M$ point pairs data around their locations satisfying
		\begin{equation}
			\label{equation:RotationSearch}
			\mathbf{R}^\prime=\mathop{\arg\min}\limits_{\mathbf{R}^*\in\rm{SO}(3)}\sum_{i=1}^M Riem(\mathbf{R}^*,\mathbf{R}^\prime_i)
		\end{equation}
		where $\mathbf{R}^\prime_i$ is the optimal rotation estimation for every point pair (i.e., $\mathbf{p}^\prime_i$ and $\mathbf{q}^\prime_i$). Although $\mathbf{R}^\prime_i$ cannot be solved using only one point pair, the influence of this point pair on the final result can be quantified using Eq. (\ref{equation:RotationSearch}). Because the same point pair can be aligned, their sensitivities are equal.
		\begin{equation}
			\begin{aligned}
				||\mathbf J_{\mathbf{p}^\prime_i}||         & =||\mathbf J_{\mathbf{q}^\prime_i}||         \\
				||\mathbf J_{\mathbf{p}^{\prime\prime}_i}|| & =||\mathbf J_{\mathbf{q}^{\prime\prime}_i}||
			\end{aligned}
		\end{equation}
		Considering the exchange point pairs, $E(Riem(\mathbf{R}^*,\mathbf{R}^\prime))$ and $E(Riem(\mathbf{R}^*,\mathbf{R}^{\prime\prime}))$, the different parts are comparable.
		\begin{equation}
			\label{equation:observation contribution}
			\begin{aligned}
				\sum_{i=1}^M\frac{\Phi_{\mathbf{p}^\prime_i}^2}{\mathbf J_{\mathbf{p}^\prime_i}} & <\sum_{i=1}^M\frac{\Phi_{\mathbf{p}^{\prime\prime}_i}^2}{\mathbf J_{\mathbf{p}^{\prime\prime}_i}} \\
				\sum_{i=1}^M\frac{\Phi_{\mathbf{q}^\prime_i}^2}{\mathbf J_{\mathbf{q}^\prime_i}} & <\sum_{i=1}^M\frac{\Phi_{\mathbf{q}^{\prime\prime}_i}^2}{\mathbf J_{\mathbf{q}^{\prime\prime}_i}}
			\end{aligned}
		\end{equation}
		Subsequently, the two equations in Eq. (\ref{equation:observation contribution}) by substituting Eq. (\ref{equation:Sensitivity}) and (\ref{equation:Uncertainty}). Considering Eq. (\ref{equation:Expectation of Riemannian Distance}), $\theta_{\mathbf{L}^\prime}$, $\theta_{\mathbf{K}^\prime}$, $\theta_{\mathbf{L}^{\prime\prime}}$, and $\theta_{\mathbf{K}^{\prime\prime}}$ are the small disturbances. The term $\theta^2_{\mathbf{K}^\prime}\theta^2_{\mathbf{L}^\prime}$ is of fourth order. The main related terms are of second order.
		\begin{equation}
			\begin{aligned}
				\sum_{i=1}^M\frac{\epsilon_{\mathbf{p}^\prime_i}^2+\epsilon_{\mathbf{q}^\prime_i}^2}{||\mathbf{p}^\prime_i||^2} & <\sum_{i=1}^M\frac{\epsilon_{\mathbf{p}^{\prime\prime}_i}^2+\epsilon_{\mathbf{q}^{\prime\prime}_i}^2}{||\mathbf{p}^{\prime\prime}||^2} \\
				\Leftrightarrow \sum_{i=1}^M \theta_{\mathbf{L}^\prime}^2+\theta_{\mathbf{K}^\prime}^2                          & <\sum_{i=1}^M \theta_{\mathbf{L}^{\prime\prime}}^2+\theta_{\mathbf{K}^{\prime\prime}}^2                                                \\
				\Leftrightarrow E(Riem(\mathbf{R}^*,\mathbf{R}^\prime))                                                         & <E(Riem(\mathbf{R}^*,\mathbf{R}^{\prime\prime}))
			\end{aligned}
		\end{equation}
		\begin{center}
			\bf{Q.E.D.}
		\end{center}
	\end{proof}
	
	Therefore, the demonstration is terminated at the expectation comparison because the dynamic disturbance parts make direct comparison impossible. Therefore, our selection scheme was statistically optimal. In the next section, we define more complex, close-to-reality sensitivity, and uncertainty models to describe the real LiDAR measurement points.
	
	\section{Enhance LiDAR Odometry Accuracy}
	\label{section:Enhance Lidar Odometry Accuracy}
\begin{figure}[htbp]
		\centering
		\includegraphics[width=0.45\textwidth]{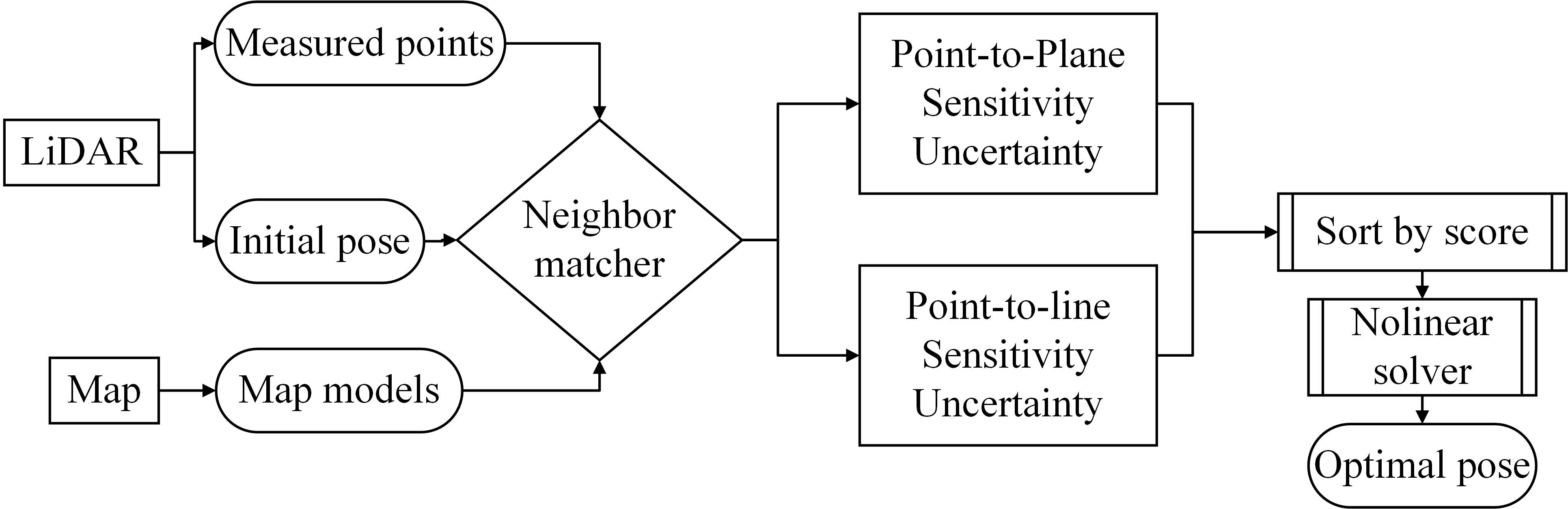}
		\caption{Outline details the selection scheme. The inputs include map points, LiDAR measured points, and an initial pose available from a uniform motion model or IMU. Subsequently, we employed an octree to find neighbors that establish the closest matches. Because measured points are classified into surf and corner (plane and line), the algorithm computes sensitivities and uncertainties separately. Finally, we sorted all residual terms by sensitivity and uncertainty scores, stopped at a threshold, and sent them into the nonlinear solver to derive the optimal pose.}
		\label{figure:Pipline Contribution}
	\end{figure}

	This section describes the practical application of our theory. The complete procedure is shown in Fig. \ref{figure:Pipline Contribution}. An outline details the selection scheme. The inputs included map points, LiDAR measured points, and an initial pose available from a uniform motion model or IMU. Subsequently, we used an octree to find neighbors that established the closest matches. Because the measured points are classified as surf and corner (plane and line), the algorithm computes the sensitivities and uncertainties separately. Finally, we sorted all residual terms by sensitivity and uncertainty scores, stopped at a threshold, and sent them to the nonlinear solver to derive the optimal pose.
	
	This section first presents the method for calculating the sensitivity model. It uses a Taylor expansion and eigenvalue projection tool to decouple residuals into six dimensions depending on the type of point-to-plane and point-to-line residuals. The second section presents the calculation of the uncertainty model. The laser scan beam and geometry patterns are analyzed to describe the uncertainties in this process. The third section explains the final selection standard, which comprehensively considers the influences of sensitivity and uncertainty.
	
	\subsection{Sensitivity model}
	\begin{figure}[htbp]
		\centering
		\subfigure[]{
			\includegraphics[width=0.09\textwidth]{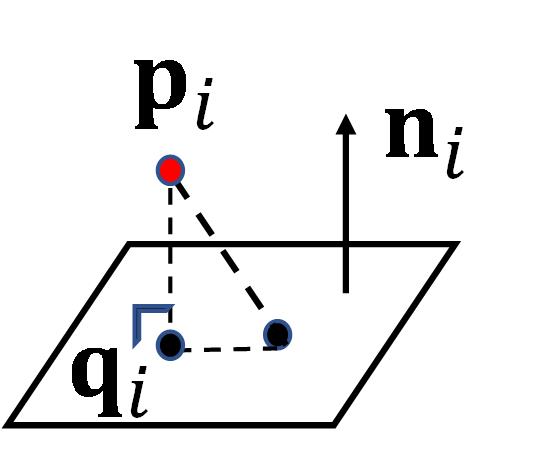}
			\label{figure:PlaneResidual}}
		\hspace{0.1\textwidth}
		\subfigure[]{
			\includegraphics[width=0.1\textwidth]{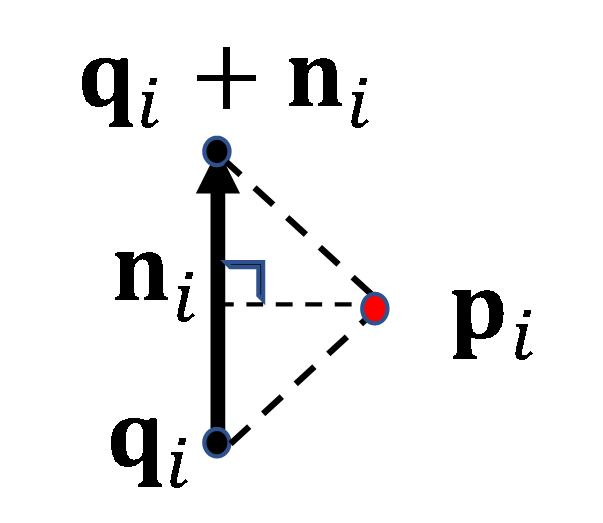}
			\label{figure:LineResidual}}
		\caption{Two types of common residuals used in LO: (a) point-to-plane distance and (b) point-to-line distance.}
	\end{figure}
	
	To satisfy the assumption of an infinitesimal rotation and translation, the linearization error approaches zero.
	\begin{equation}
		\mathbf{R}\approx
		\begin{bmatrix}
			1    & -r_z & r_y  \\
			r_z  & 1    & -r_x \\
			-r_y & r_x  & 1
		\end{bmatrix}
		\approx\rm \mathbf{I}_3 + 
		\begin{bmatrix}
			r_x \\
			r_y \\
			r_z
		\end{bmatrix}^\wedge
	\end{equation}
	\subsubsection{Point-to-plane distance}
	A LiDAR measured point $\mathbf{p}_i$ and the corresponding map point $\mathbf{q}_i$, which is defined as a point on the plane. The normal vector is $\mathbf{n}_i$, as shown in Fig. \ref{figure:PlaneResidual}. The error of the $i$-index residual-term point-to-plane distance is
	\begin{equation}
		e^{pl}_i=(\mathbf{Rp}_i+\mathbf{t}-\mathbf{q}_i)^{\rm T}\mathbf{n}_i
	\end{equation}
	and $e^{pl}_i$ is scalar.
	\begin{equation}
		\begin{aligned}
			e^{pl}_i & \approx\left[(\mathbf{I}+\mathbf{r}^\wedge)\mathbf{p}_i+\mathbf{t}-\mathbf{q}_i\right]^{\rm T}\mathbf{n}_i \\
			& \approx\left[\mathbf{p}_i+\mathbf{r}\times \mathbf{p}_i+\mathbf{t}-\mathbf{q}_i\right]^{\rm T}\mathbf{n}_i
		\end{aligned}
	\end{equation}
	Residual sensitivity describes the $\mathbf{r}$ and $\mathbf{t}$ on $e^{pl}_i$. Thereafter, we used the Jacobian tool and linearized rotation to calculate this property.
	\begin{equation}
		\label{equation:sensitivity point to plane}
		\mathbf{J}_{e^{pl}_i}=\left[\frac{\partial {e^{pl}_i}^{\rm T}}{\partial (\mathbf{r},\mathbf{t})}\right]^{\rm T}=
		\begin{bmatrix}
			(\mathbf{p}_i\times \mathbf{n}_i)^{\rm T} & \mathbf{n}_i^{\rm T}
		\end{bmatrix}
	\end{equation}
	
	\subsubsection{Point-to-line distance}
	A LiDAR measured point, $\mathbf{p}_i$, and the corresponding map point, $\mathbf{q}_i$, which is defined as a point on the line. Its pointing direction is the unit vector $\mathbf{n}_i$, as shown in Fig. \ref{figure:LineResidual}. Before forming the distance, a new vector $\mathbf{d}_i$ should first be defined.
	\begin{equation}
		\mathbf{d}_i=(\mathbf{q}_i-\mathbf{Rp}_i-\mathbf{t})\times(\mathbf{q}_i+\mathbf{n}_i-\mathbf{Rp}_i-\mathbf{t})
	\end{equation}
	where $\mathbf{d}_i$ denotes a $3\times1$ vector. Its norm is the parallelogram area of vectors $\mathbf{p}_i$ to $\mathbf{q}_i$ and $\mathbf{p}_i$ to $\mathbf{q}_i+\mathbf{n}_i$. Its direction was orthogonal to the plane of the two vectors. Because the norm vector $\mathbf{n}_i$ is a unit vector, the number of areas is exactly equal to the distance. The error in the $i$-index residual term point-to-line distance $e^{li}_i$ is defined as
	\begin{equation}
		e^{li}_i=\mathbf{d}_i^{\rm T}\mathbf{d}_i
	\end{equation}
	This differed from the point-to-plane distance. First, $\mathbf{d}_i$ is derived as
	\begin{equation}
		\begin{aligned}
			\mathbf{d}_i & \approx(\mathbf{q}_i-\mathbf{p}_i-\mathbf{t}-\mathbf{r}\times \mathbf{p}_i)\times(\mathbf{q}_i-\mathbf{p}_i-\mathbf{t}-\mathbf{r}\times \mathbf{p}_i+\mathbf{n}_i) \\
			& \approx(\mathbf{q}_i-\mathbf{p}_i-\mathbf{t}-\mathbf{r}\times \mathbf{p}_i)\times \mathbf{n}_i
		\end{aligned}
	\end{equation}
	The Jacobian of the distance $\mathbf{d}_i$ is
	\begin{equation}
		\label{equation:sensitivity point to line}
		\mathbf{J}_{\mathbf{d}_i}=\left[\frac{\partial \mathbf{d}_i^{\rm T}}{\partial (\mathbf{r},\mathbf{t})}\right]^{\rm T}=
		\begin{bmatrix}
			(\mathbf{n}_i^{\rm T}\mathbf{p}_i){\rm \mathbf{I}_3}-\mathbf{p}_i \mathbf{n}_i^{\rm T} & \mathbf{n}^\wedge
		\end{bmatrix}
	\end{equation}
	where $\mathbf{J}_{\mathbf{d}_i}$ is a $3\times6$-matrix. Moreover,
	\begin{equation}
		\Delta e^{li}_i=\Delta
		\begin{bmatrix}
			\mathbf{r} & \mathbf{t}
		\end{bmatrix}
		\mathbf{J}_{\mathbf{d}_i}^{\rm T}\mathbf{J}_{\mathbf{d}_i}\Delta
		\begin{bmatrix}
			\mathbf{r} \\
			\mathbf{t} \\
		\end{bmatrix}
	\end{equation}
	Thus, Hessian matrix $\mathbf{H}_{\mathbf{d}_i}=\mathbf{J}_{\mathbf{d}_i}^{\rm T}\mathbf{J}_{\mathbf{d}_i}$ is defined. Therefore, point-to-line $e^{li}_i$ is a quadratic form of the optimization parameters, different from the linear form in the point-to-plane distance. Direct decoupling into six dimensions is impossible because the partial derivatives of the quadratic function approximating $\Delta \mathbf{r}={\bf 0}$ and $\Delta \mathbf{t}={\bf 0}$ are consistently zero. Thus, we have focused on the growing gradient in a small region. The Hessian matrix was projected onto the six axes. Every eigenvalue with vectors was projected onto the $j$-index axis. They were regrouped in linear form.
	
	\subsection{Uncertainty model}
	Before introducing the uncertainty model, the accuracy and variance should be defined. For the standard variable $\theta$, the accuracy is $\Delta\theta$. As shown in Table \ref{table:Distribution Variance}, when the error associated with the variable $\theta$ is defined with distinct distributions, the variance is different.
	\begin{table}[htbp]
		\begin{center}
			\caption{Distribution and variance notations}
			\renewcommand\arraystretch{1.4}
			\resizebox{0.45\textwidth}{!}
			{
				\begin{tabular}{|c|c|c|c|}
					\hline
					variable          &            accuracy             & distribution type &                   variance                    \\ \hline
					\multirow{3}{*}{$\theta$} & \multirow{3}{*}{$\Delta\theta$} &      Uniform      & $\sigma_\theta=\frac{\Delta\theta}{\sqrt{3}}$ \\ \cline{3-4}
					&                                 &     Gaussian      &         $\sigma_\theta=\Delta\theta$          \\ \cline{3-4}
					&                                 &   Not measured    &               $\sigma_\theta=0$               \\ \hline
				\end{tabular}
			}
			\label{table:Distribution Variance}
		\end{center}
	\end{table}
	
	\subsubsection{Laser scan beam}
\begin{figure}[htbp]
		\centering
		\subfigure[coordinates]{
			\includegraphics[width=0.25\textwidth ,height=0.15\textwidth]{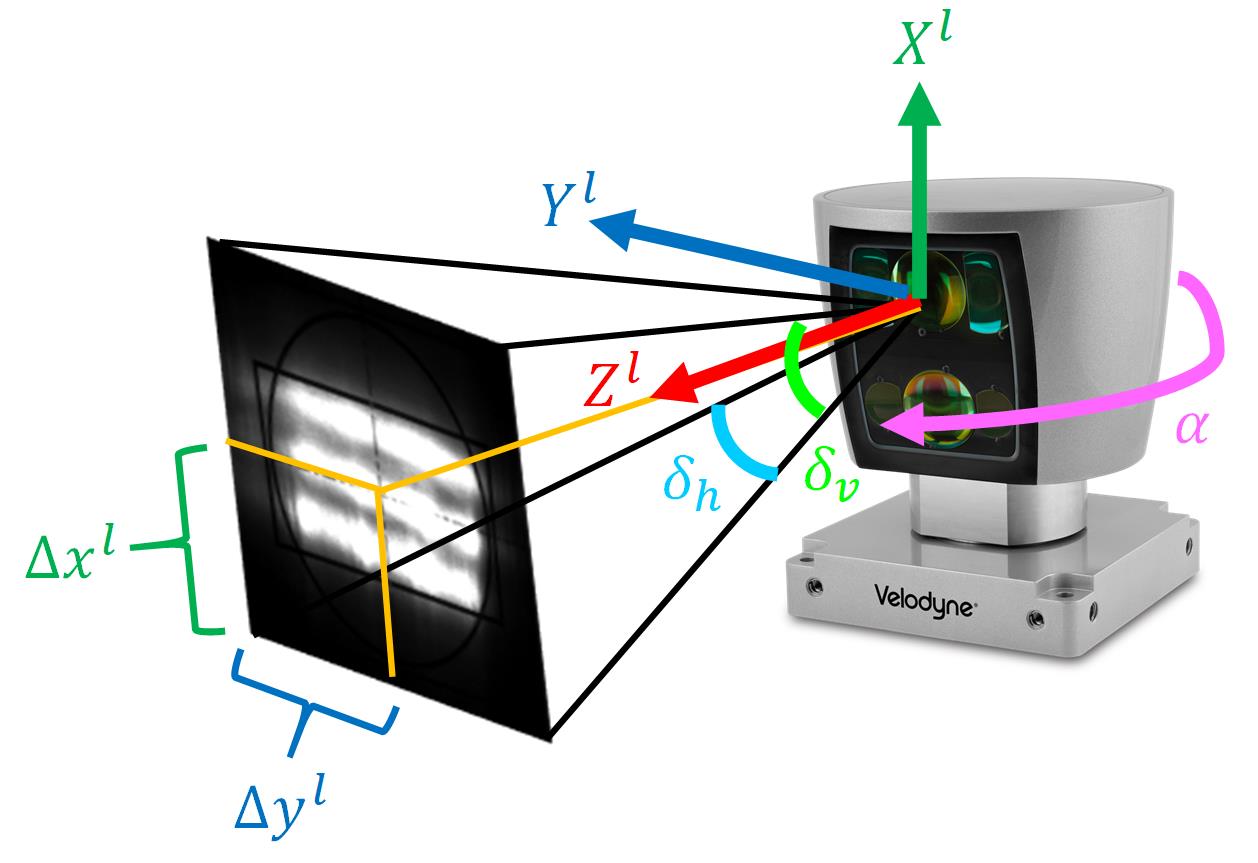}
			\label{figure:LidarScanModel4}}
		\hspace{0.15\textwidth}
		\subfigure[diode stack]{
			\includegraphics[width=0.13\textwidth ,height=0.1\textwidth]{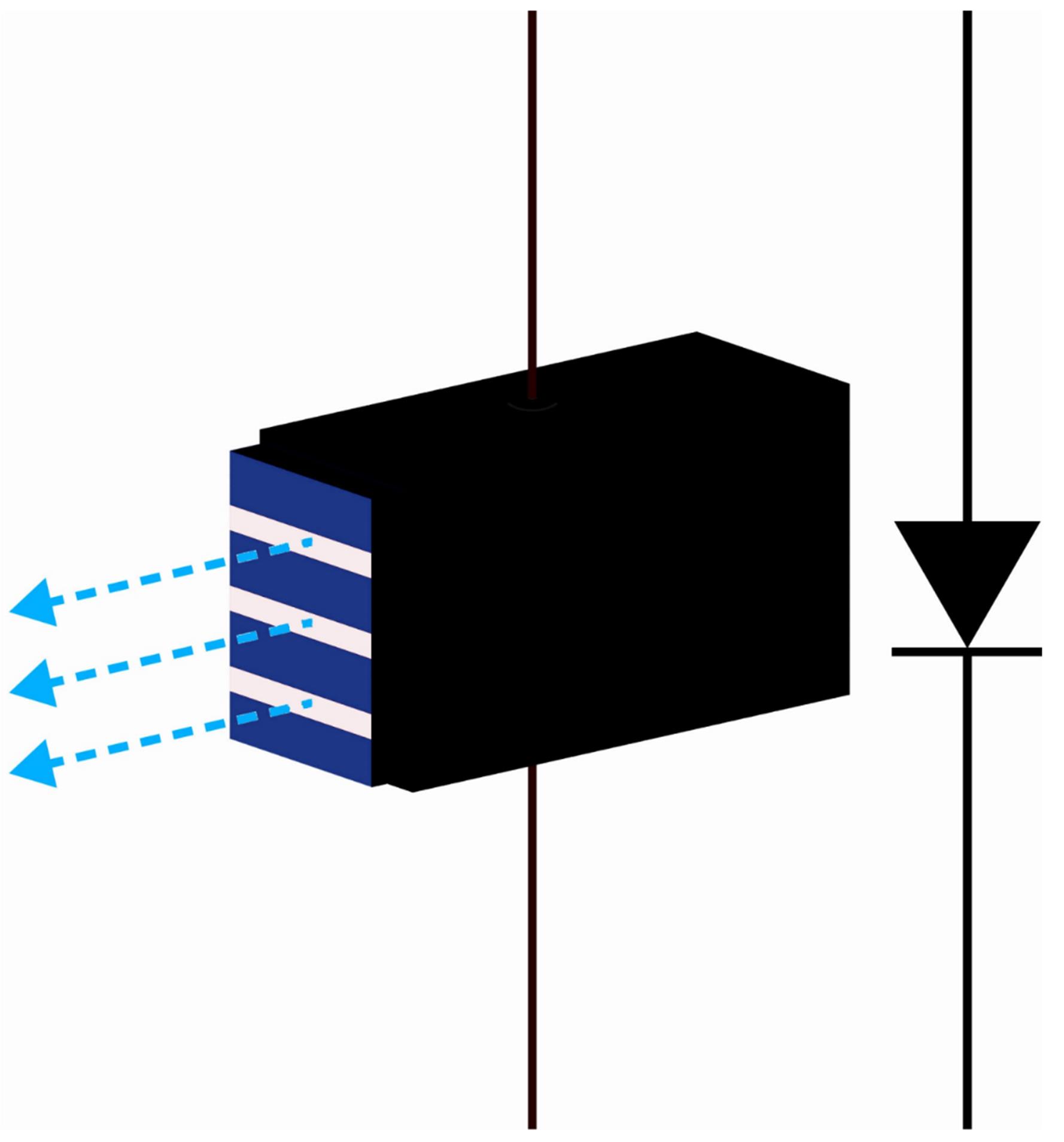}
			\label{figure:LidarScanModel1}}
		\subfigure[receiver]{
			\includegraphics[width=0.13\textwidth ,height=0.1\textwidth]{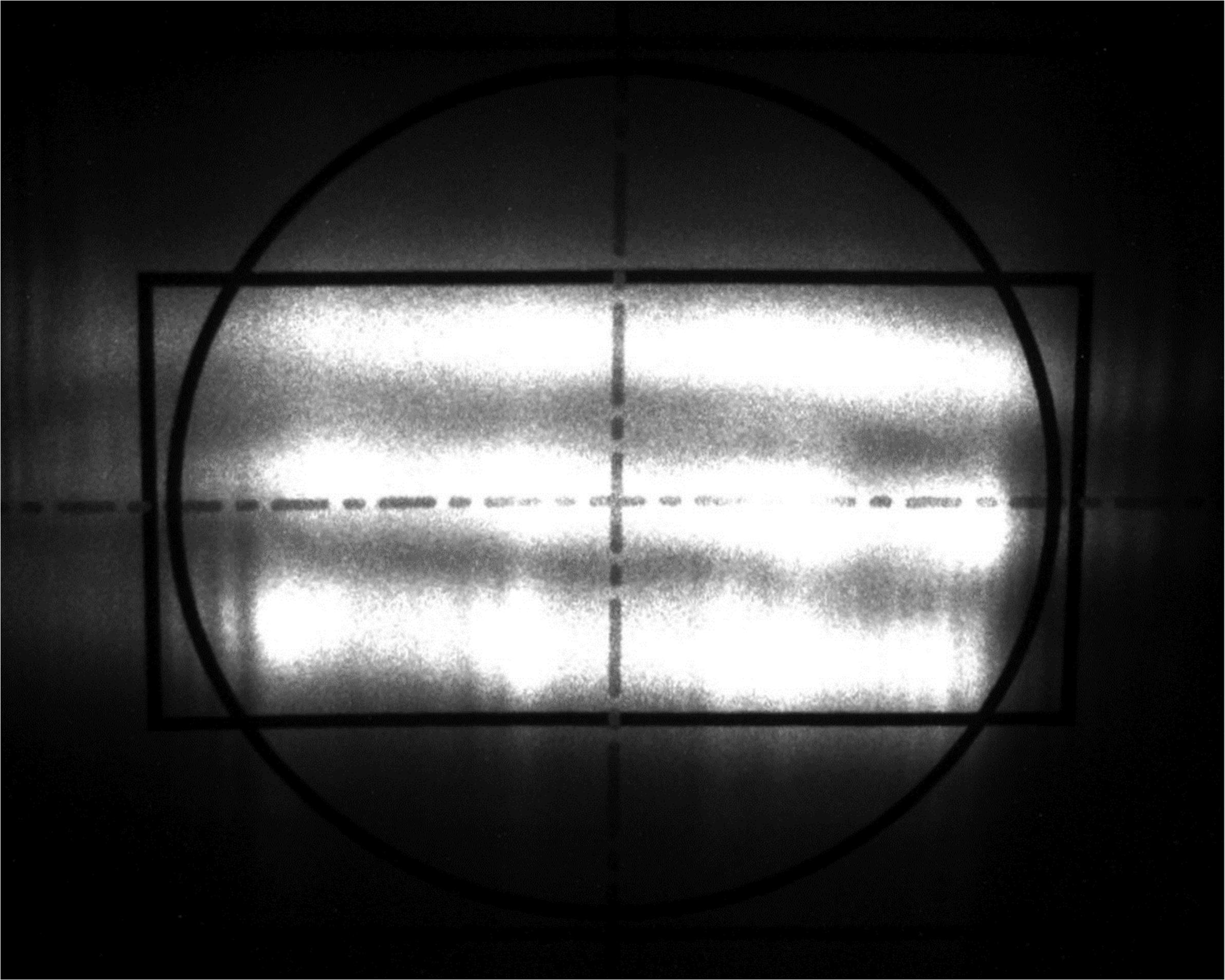}
			\label{figure:LidarScanModel2}}
		\subfigure[laser scan lines]{
			\includegraphics[width=0.13\textwidth ,height=0.1\textwidth]{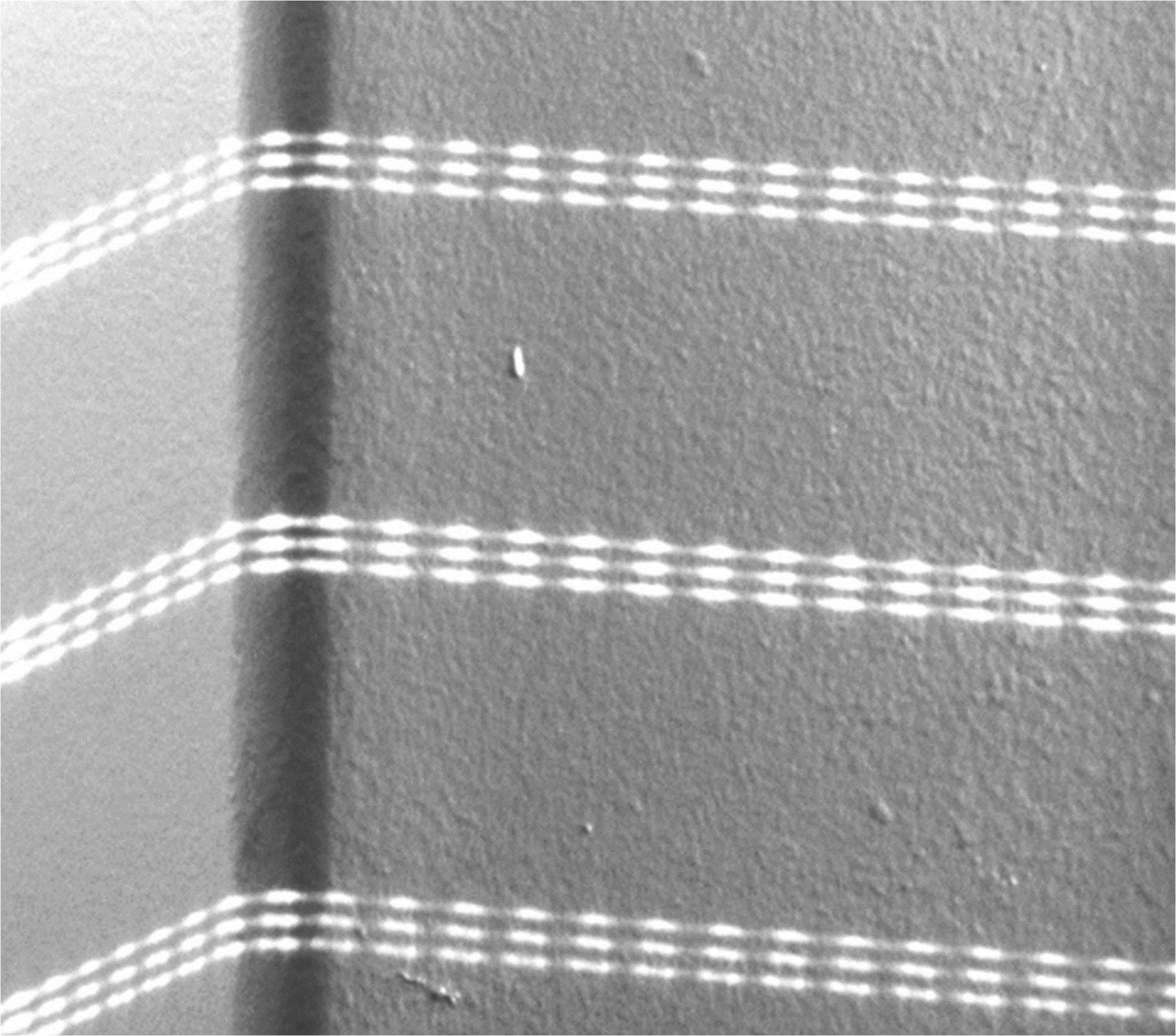}
			\label{figure:LidarScanModel3}}
		\caption{LiDAR laser scan beam model. (a) is the coordinate from laser to scanner. (b) is a laser diode stack emits three light beams. (c) is an infrared observation window. (d) indicates three laser lines and their footprints.}
	\end{figure}

	Based on a multibeam laser scanner system \cite{2007LidarError,2007AirLaser,2018VLP16Model}, the rotation $\mathbf{R}^{sl}$ is from the laser coordinate $l$ to the scanner coordinate $s$. Typically, a mechanical spinning device, which creates a fixed laser to a circular scanner, as shown in Fig. \ref{figure:LidarScanModel4}.
	\begin{equation}
		\mathbf{R}^{sl}(\alpha,\omega)=
		\begin{bmatrix}
			cos\omega  & 0 & sin\omega \\
			0          & 1 & 0         \\
			-sin\omega & 0 & cos\omega
		\end{bmatrix}
		\begin{bmatrix}
			1 & 0         & 0          \\
			0 & cos\alpha & -sin\alpha \\
			0 & sin\alpha & cos\alpha
		\end{bmatrix}
	\end{equation}
	where $\alpha$ is the azimuth angle and $\omega$ is the elevation angle of the laser beam channel.
	
	As illustrated in Fig. \ref{figure:LidarScanModel1}, a laser diode stack emits three light beams. They fall on the environment surface and are reflected in the LiDAR observation window, as shown in Fig. \ref{figure:LidarScanModel2}. LiDAR records the emission time and the most intense time to calculate depth. Owing to the observation window, the laser depth can be simulated as a divergent beam. The true location can lie anywhere within the beam footprint. According to the manual, the Velodyne Puck (VLP-16) claims $\Delta z^l=3\ cm$. The horizontal and vertical divergence angles of the rectangular window were $\delta _h=3 \times 10^{-3}\ rad$ and $\delta _v=1.5 \times 10^{-3}\ rad$. Therefore, assuming that point is uniform in this region,
	\begin{equation}
		\sigma _{x^l}=\frac{\Delta x^l}{\sqrt{3}}=\frac{z^l tan(\frac{\delta _v}{2})}{\sqrt{3}},
		\sigma _{y^l}=\frac{\Delta y^l}{\sqrt{3}}=\frac{z^l tan(\frac{\delta _h}{2})}{\sqrt{3}}
	\end{equation}
	
	Following the self-rotation, a laser scan line was formed, as shown in Fig. \ref{figure:LidarScanModel3}. Every elevation angle $\omega$ was carefully calibrated and rectified; thus, $\sigma _\omega=0$. For the azimuth $\alpha$, the manual states that the rotation angular resolution is $0.01\ \degree$. All studies in \cite{2007LidarError,2020LidarComparing,2007AirLaser,2018VLP16Model} assumed that
	\begin{equation}
		0<\sigma _\alpha<\frac{\pi\Delta\alpha}{180\sqrt{3}}
	\end{equation}
	where $\Delta\alpha= 0.005\ \degree$ is the half resolution.
	
	The LiDAR parameters and coordinates are shown in Fig. \ref{figure:LidarScanModel4}. Because self-rotation is nonlinear, $\mathbf{R}^{sl}(\alpha,\omega)$ must be linearized, and uncertainty propagation works. Finally, uncertainty is a matrix. In laser coordinate $l$, it is a $5\times 5$ matrix $\Sigma^l$. The scanner coordinate $s$ is a $3\times 3$ matrix $\Sigma^s$. They are connected by uncertainty propagation as follows:
	\begin{equation}
		\Sigma^s=\mathbf{J}_{\mathbf{R}^{sl}(\alpha,\omega)}^T\Sigma^l\mathbf{J}_{\mathbf{R}^{sl}(\alpha,\omega)}
	\end{equation}
	where $\mathbf{J}_{\mathbf{R}^{sl}(\alpha,\omega)}$ denotes a $3\times 5$ matrix. This is derived from the first-order Taylor expansion formula of $\mathbf{R}^{sl}(\alpha,\omega)$.
	\begin{equation}
		\mathbf{J}_{\mathbf{R}^{sl}(\alpha,\omega)}={\left[\frac{\partial \mathbf{R}^{sl}(\alpha,\omega)}{\partial (x,y,z,\alpha,\omega)}\right]}^T
	\end{equation}
	$\Sigma^l$ is generated as a diagonal matrix from the individual sources:
	\begin{equation}
		\Sigma^l=\rm diag
		\begin{bmatrix}
			\sigma^2_{x^l} & \sigma^2_{y^l} & \sigma^2_{z^l} & \sigma^2_\alpha & \sigma^2_\omega
		\end{bmatrix}
	\end{equation}
	These variable variances have been discussed previously, and some can be found in the LiDAR sensor manual.
	
	\subsubsection{Geometry pattern}
	\begin{figure}[htbp]
		\centering
		\subfigure[]{
			\includegraphics[width=0.2\textwidth]{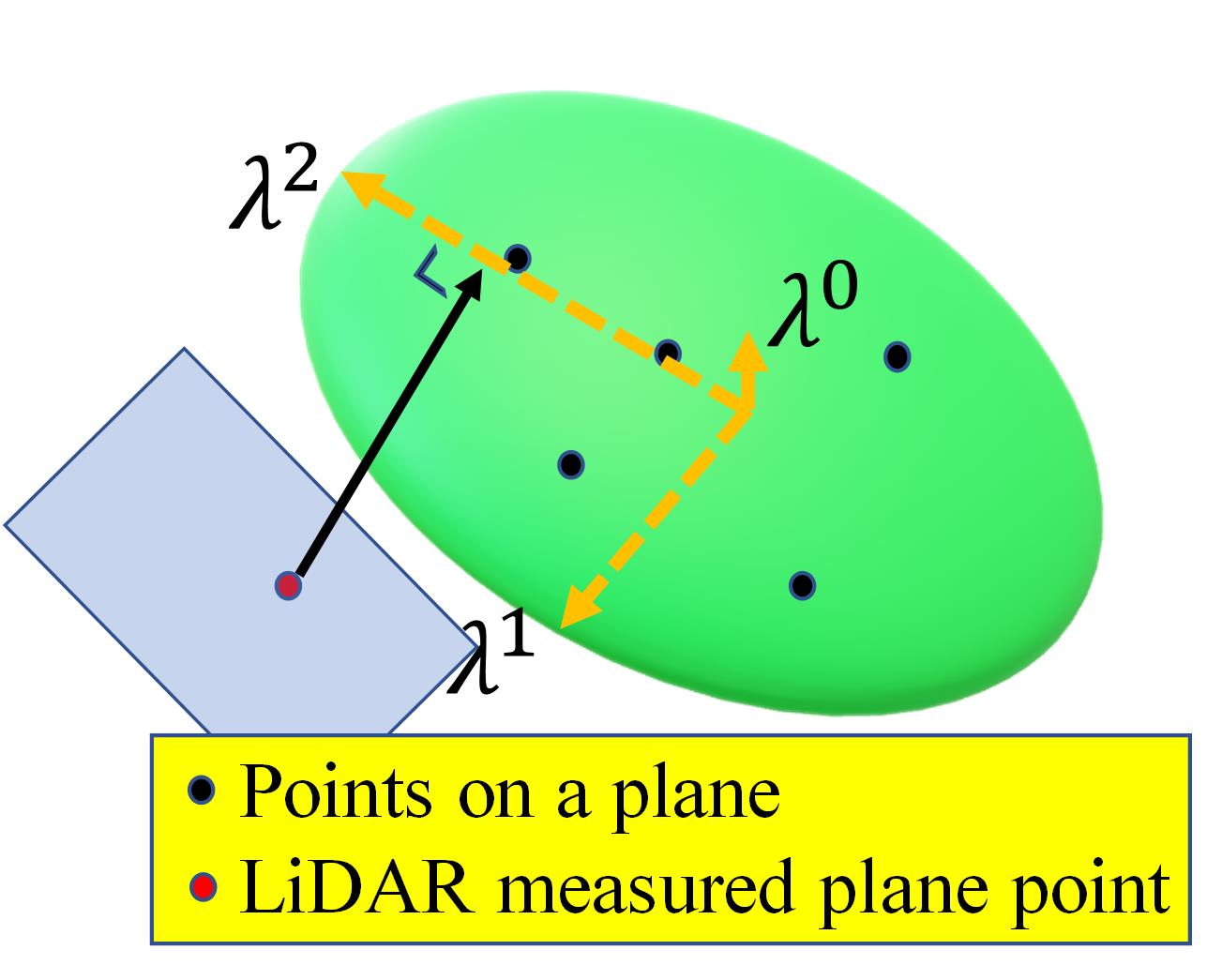}
			\label{figure:PointPlane}}
		\hspace{0.05\textwidth}
		\subfigure[]{
			\includegraphics[width=0.15\textwidth]{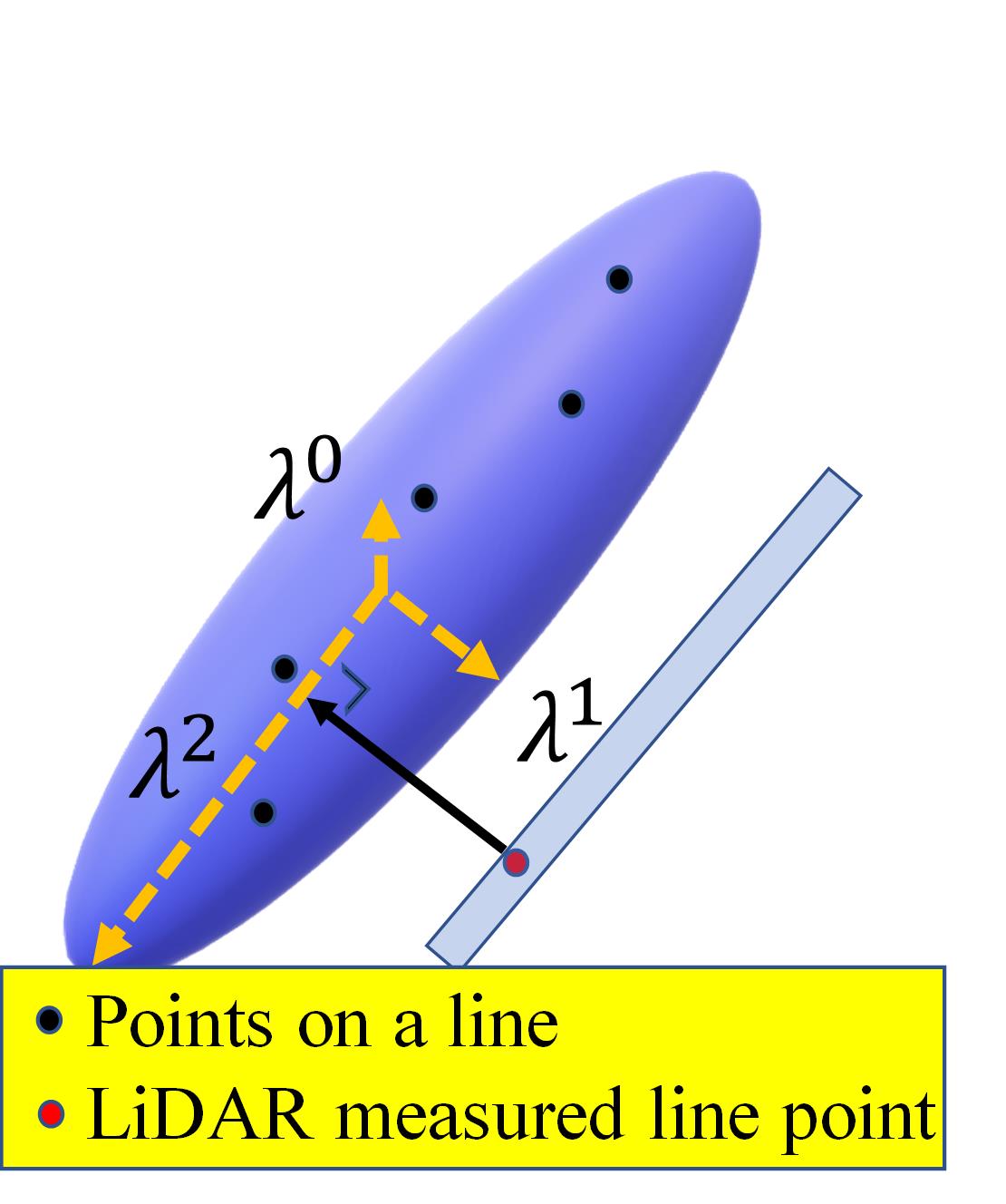}
			\label{figure:PointLine}}
		\caption{Two error types establish different residual terms. (a) Point-to-plane distance and (b) point-to-line distance. Registration attempts to adjust rotation and translation and then decrease these error distances.}
	\end{figure}
	Owing to inhomogeneous noise, sparse density, and missing data in LiDAR sampling \cite{2018SurfaceBasedGICP}, pose estimation typically employs plane and line patterns. Several studies \cite{2014LOAM,2018LeGOLOAM,2019LIOmapping,2020LIOSAM,2021MULLS} have minimized the alignment distance. The LO baseline LOAM \cite{2014LOAM} uses five neighboring points to model the plane or line shown in Fig. \ref{figure:PointPlane} and \ref{figure:PointLine}. Applying PCA technique, eigenvalues $\lambda^0<\lambda^1<\lambda^2$ and eigenvectors $\mathbf{\nu}^0$, $\mathbf{\nu}^1$ and $\mathbf{\nu}^2$ are calculated by applying the PCA technique. These correspond to $x$,$y$, and $z$ dimensions. This process essentially involves modeling the surface as a 3D Gaussian ellipsoid.
	
\begin{figure}[htbp]
		\centering
		\subfigure[low uncertainty fusion result]{
			\includegraphics[width=0.15\textwidth]{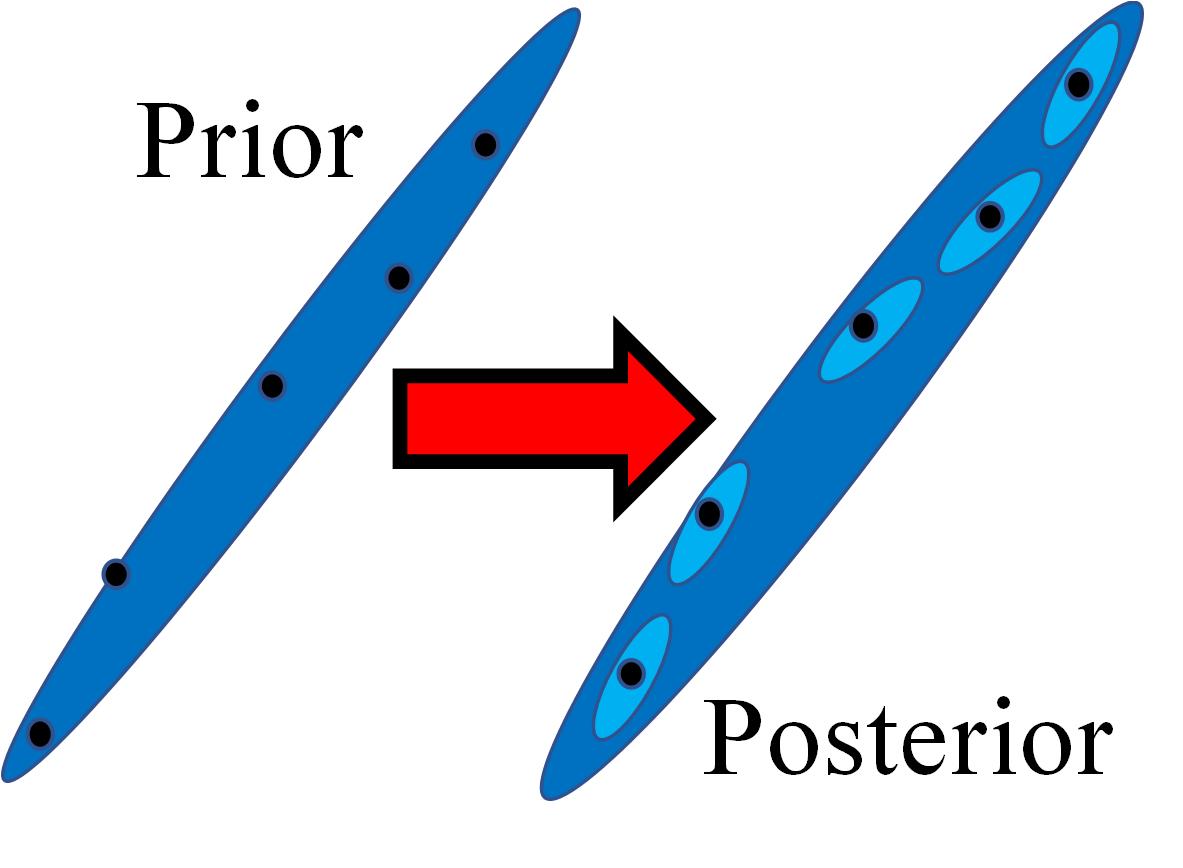}
			\label{figure:SmallUncertaintyFusion}}
		\hspace{0.05\textwidth}
		\subfigure[high uncertainty fusion result]{
			\includegraphics[width=0.17\textwidth]{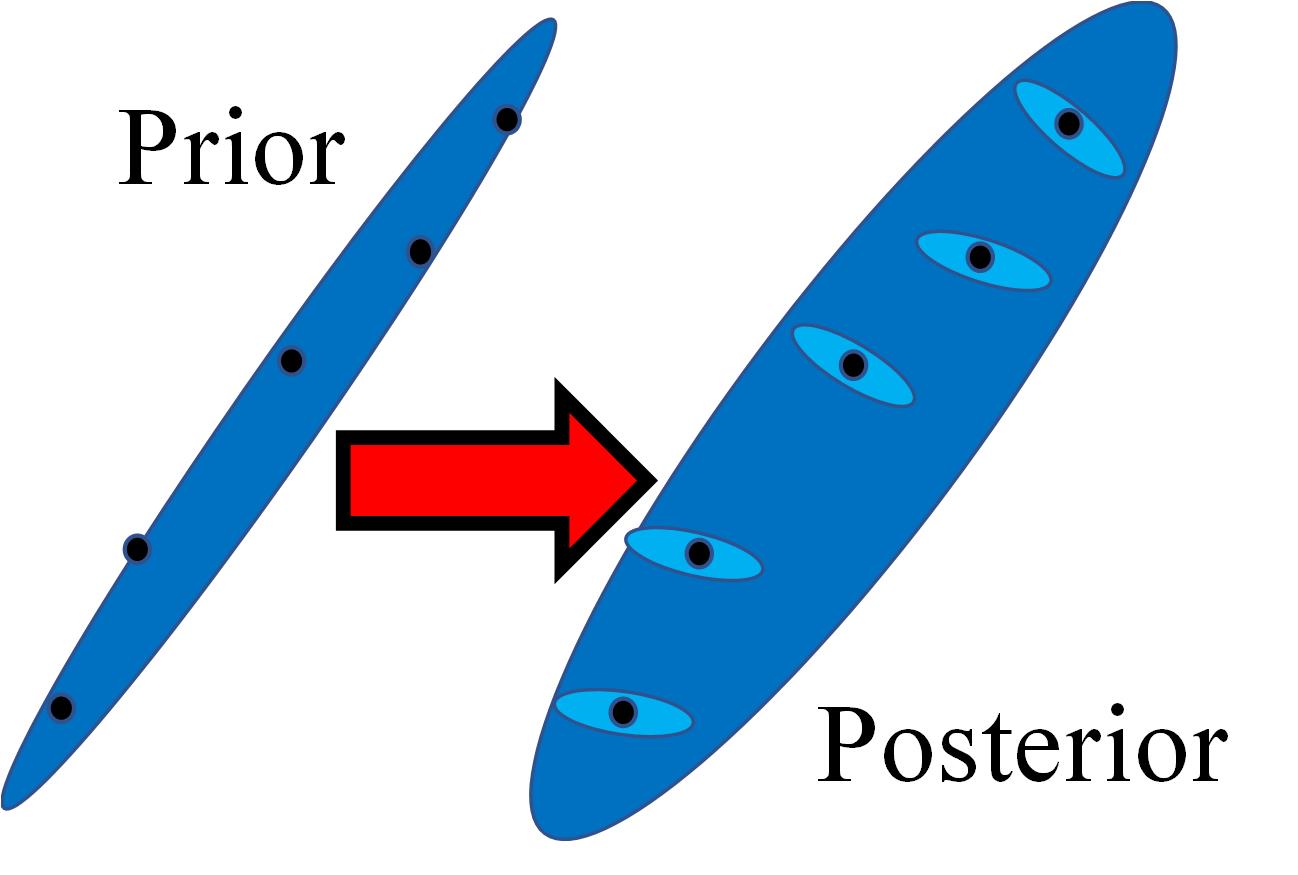}
			\label{figure:BigUncertaintyFusion}}
		\caption{Resampling points to measurement uncertainties. These points' distances to the mean are one sigma; low uncertainty fusion is satisfactory for estimation.}
	\end{figure}

	We should comprehensively consider the influences of both current LiDAR measured information uncertainties (laser scan beam) and history-map model uncertainties (geometry pattern). As shown in Fig. \ref{figure:SmallUncertaintyFusion}, the prior only considers modeling these map points as a plane. After adding the information of every point uncertainty, although the posterior becomes slightly fat, this fusion result indicates that this model is sufficiently good for pose estimation. In Fig. \ref{figure:BigUncertaintyFusion}, considering every point uncertainties, the posterior becomes thick in the main direction, and this fusion result is bad. Because our purpose is to model uncertainties in registration, the main error direction uncertainties are modeled using the sigma-point transform technique \cite{1996SigmaPointTransform}. The points were resampled around the ellipsoid to infer the posterior Gaussian distribution. The distances of these points to the mean are one sigma.
	
	Similar to minimizing the Kullback-Leibler divergence \cite{2007KLDivergenceGMM} between two Gaussians, we provide a simple fusion method in which the eigenvalue along the registration direction is employed to reflect the disparity:
	\begin{equation}
		\label{equation:uncertainty}
		\Phi_{e_i}=
		\begin{cases}
			(\lambda_{i_{sou}}^0+\lambda_{i_{tar}}^0)/2                                         & (plane) \\
			(\lambda_{i_{sou}}^0+\lambda_{i_{sou}}^1+\lambda_{i_{tar}}^0+\lambda_{i_{tar}}^1)/4 & (line)
		\end{cases}
	\end{equation}
	where $\Phi_{e_i}$ is a scalar that evaluates uncertainty. $\lambda_{i_{sou}}^0$ is the smallest eigenvector of the source point distribution, and $\lambda_{i_{tar}}^0$ is for the target.
	
	\subsection{Sort by score}
	Motivated by Eq. (\ref{equation:Point Contribution Model 5}), combining the sensitivity and uncertainty models from Lemmas \ref{lemma:Sensitivity} and \ref{lemma:Uncertainty} into a score, and judging the residual influence on the pose estimation accuracy. In practice, using Eqs. (\ref{equation:sensitivity point to plane}), (\ref{equation:sensitivity point to line}), and (\ref{equation:uncertainty}), the score for a residual can be derived as
	\begin{equation}
		\Psi_{e_i}=\frac{\mathbf{J}_{e_i}}{\Phi_{e_i}^2}
	\end{equation}
	where the score $\Psi_{e_i}$ is a $6\times1$ vector that corresponds to three rotations and three translations. Next, we performed score sorting and selected residuals from the top big score in every six dimensions, in parallel. The repeated terms were recorded only once. Because the dislocation match and geometry assumption (plane line) cause four point pairs to be unstable, we set a threshold parameter to judge the stop rule: (1) the selected residual amount reaches $200$ per dimension and (2) the residual score decreases to $10\%$ of the maximal.
	
	\section{Experiments}
	We used simulation, benchmark, and our captured real data to introduce the experiments, which were segmented into three parts. The first part is a two-frame point cloud registration simulation (Section \ref{subsection:Simulation}), which controls the noise amplitude in the measurements and models. This verifies the validity of the residual selection scheme in a controlled environment. The second part is the KITTI benchmark \cite{2012KITTI} comparison (section \ref{subsection:KITTI Benchmark}), which aims to prove the selection's general effectiveness in decreasing time cost and improving pose accuracy. The third part comprises our captured real indoor and outdoor data (Section \ref{subsection:Online Captured Scenario}). It contains two types of scan-mode LiDAR and inertial measurement unit (IMU) data. This part proves our method's validation in both the LO and LIO algorithms and is also applicable for different LiDARs. Finally, the IMU is used only in LIOmapping \cite{2019LIOmapping} for comparison purposes, which is unnecessary for our proposed algorithm.
	
	Therefore, the sensitivity- and uncertainty-theory-based residual term selection scheme achieved significant improvements in accuracy. It exhibits real-time performance with fewer residual terms and lower computational costs in nonlinear optimization. Our codes were implemented in C++. The program was executed on a desktop computer with hardware parameters of a six-core CPU AMD 2600x, 48-GB RAM, and an Nvidia RTX 2070 GPU.
	
	\subsection{Simulation}
	\label{subsection:Simulation}
	\begin{figure}[htbp]
		\centering
		\subfigure[]{
			\includegraphics[width=0.23\textwidth]{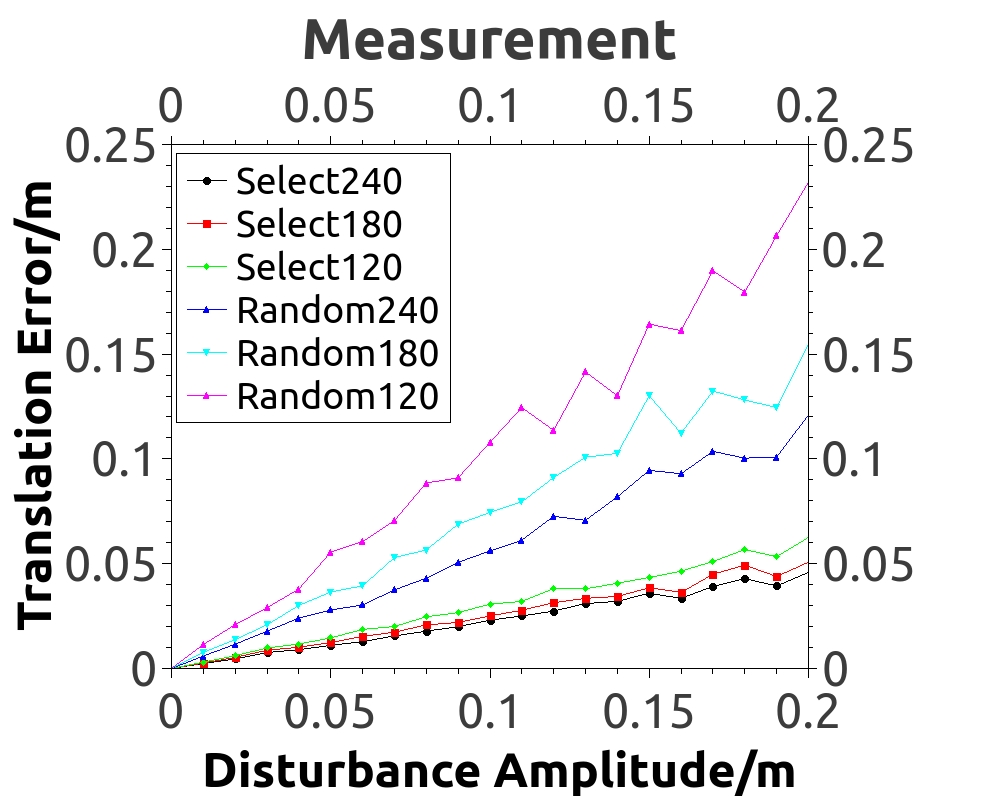}}
		\subfigure[]{
			\includegraphics[width=0.23\textwidth]{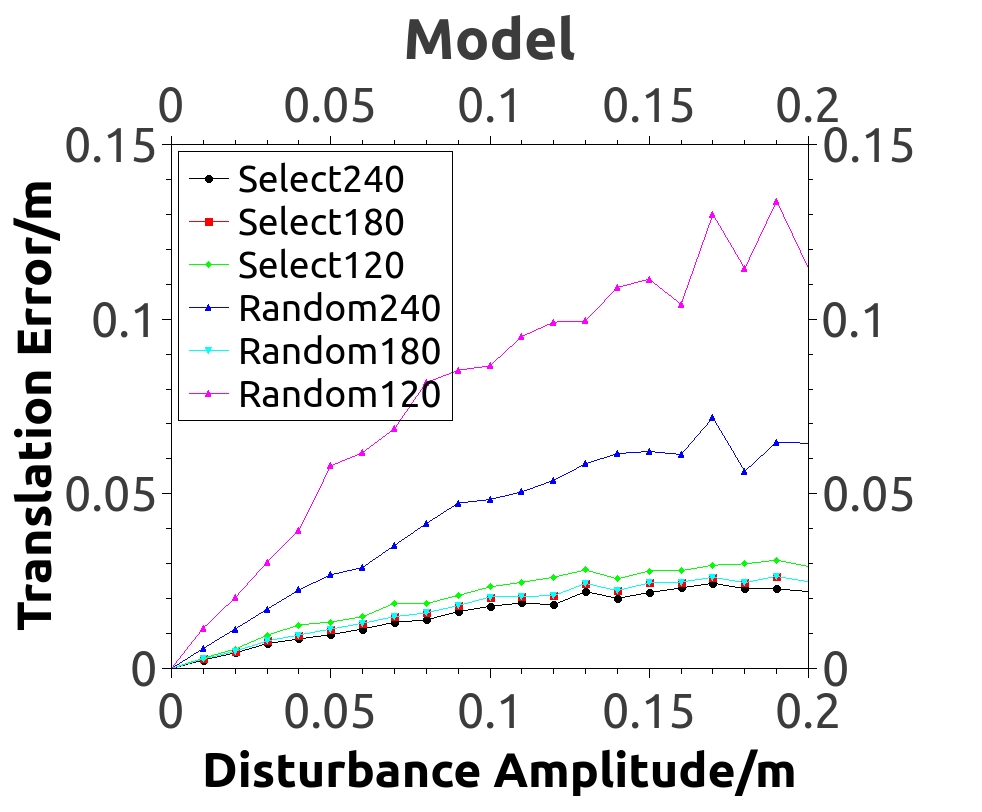}}
		\caption{LiDAR measurement points are randomly generated. The initial pose is set as an identity matrix, and 100 times experiments calculate mean translation errors for one specific parameter set. Disturbances are separately added. Under both conditions, the random method samples the same point amount as the selection method. When disturbance amplitudes are zero, both methods have no deviations. When amplitudes grow, translation errors of both methods increase; however, the growth of the proposed selection method is slower.}
		\label{figure:SimulationDisturbance}
	\end{figure}
	\begin{algorithm}[htbp]
		\label{algorithm:Simulation}
		\caption{Two-frame (source and target) registration simulation}
		\LinesNumbered
		\KwIn{input parameters: disturbance amplitude $Da$, residual number $Rn$}
		\KwOut{output results: translation error in selection $t^{sel}$ and random $t^{ran}$}
		Source LiDAR points were randomly generated from $1\ m$ to $100\ m$ in 64-circles with different depths. {/*generate points*/}\;
		The PCD density satisfies the Velodyne HDL64 angle resolution
		Every point is randomly allocated to a specific geometry pattern model, a plane (three points), or a line (two points).
		The measured points remain associated with their models, and these matches do not change during optimizations\;
		\For{$Da=0$;$Da<0.2$;$Da+=0.01${/*disturbances*/}}{
			\For{$Rn=120$;$Rn<=240$;$Rn+=60${/*select*/}}{
				\For{int $i=0$;$i<100$;$i++${/*multi-samples*/}}{
					The ground truth transformation $T_{gt}$ is randomly generated{/*generate a gt pose*/}\;
					Apply $T_{gt}$ to the source LiDAR points to generate target points{/*apply pose*/}\;
					Apply disturbance $Da$ to the measurements or models{/*apply disturbances*/}\;
					The proposed theory-based method selects $Rn$ terms and calculates the transformation result $T^{sel}_i${/*selection method*/}\;
					The random method selects $Rn$ terms and calculates the transformation result $T^{ran}_i${/*random method*/}\;
					Calculate the translation errors of the selection $t^{sel}_i=||T_{gt}-T^{sel}_i||$ and random $t^{ran}_i=||T_{gt}-T^{ran}_i||${/*one-test comparison*/}.
				}
				Calculate the mean translation errors $t^{sel}$ and $t^{ran}$ $100$ times{/* multi-sample mean*/}\;
			}
		}
	\end{algorithm}
	To evaluate the proposed theory, in Algorithm \ref{algorithm:Simulation}, a two-frame (source and target) registration simulation is implemented. The ground truth (gt) transformation and LiDAR measured points (source) were randomly generated. Every point was randomly allocated to a specific pattern model, plane (three points), or line (two points). Next, the gt pose was applied to the source points to generate the target points. Subsequently, the disturbances increase. Finally, theory-based and random selection methods were applied to solve the registration. This was repeated 100 times, and then an average translation error was derived. A comparison between the two methods reveals that the proposed method is superior, as shown in Fig. \ref{figure:SimulationDisturbance}.
	
	The resulting curves are presented in Fig. \ref{figure:SimulationDisturbance}. The disturbance amplitude increased along the horizontal axis. When the disturbance is zero, the data associations are accurate and do not change during the optimization. The proposed and random methods converge to zero. The error in the proposed method gradually increased as the disturbance amplitude increased. The random method probably selected large-error residual terms. The proposed method sorted all the terms; thus, the selected terms were optimal. This simulation demonstrated the influence of the proposed theory on improving pose estimation accuracy.
	
	\subsection{KITTI benchmark}
	\label{subsection:KITTI Benchmark}
	The KITTI benchmark is a well-known autonomous driving benchmark \cite{2012KITTI}. It includes a Velodyne LiDAR ($64$ scans), two gray cameras, and two color cameras. The GPS and IMU were used for the gt. It provides $11$ sequences with ground truths in urban, city, natural, and highway environments, and has been widely used for VO and LO evaluation.
	
	A few comparison algorithms are introduced in this section. ALOAM is a well-known advanced c++ realization of the LO baseline LOAM \cite{2014LOAM}. LOAM is now a closed source. Several other LO/LIO algorithms have been modified, such as LeGOLOAM \cite{2018LeGOLOAM} and LIO-SAM \cite{2020LIOSAM}. We compared the original ALOAM with ALOAM-select, which was added to our proposed selection scheme in front of the mapping thread optimization. The pose accuracy and time costs are compared in the following sections. MLOAM \cite{2021ICRAmloam} is another relevant work in this field that has a residual selection process. Although it was designed for a multi-LiDAR system, we modified it for one LiDAR. In particular, MLOAM (we modified) and ALOAM-select were only compared in a standard benchmark.
	
	\subsubsection{Pose accuracy}
	The first test directly utilized selection in the ALOAM mapping thread, indicating that our selected residual terms are a subset of the original code used. The results are summarized in TABLE \ref{table:KITTI Subset}. On average, the proposed method employs fewer planes and lines for optimization than the original method. Although the residuals ALOAM-select used were a subset of the original ALOAM, improvements were achieved in seven sequences. For the other four sequences, the proposed method was not superior. Nevertheless, these four sequences fell by approximately $0.05\%$, including $5\ cm$ drift over $100\ m$. We believe that waiting to be selected as a feature set restricts the improvement in accuracy. In particular, the detected feature point set of ALOAM was insufficiently large for our selection. We modify the feature detection parameter of the original ALOAM in the next test to illustrate our hypothesis.
	\begin{table*}[htbp]
		\begin{center}
			\caption{KITTI subset points}
			\resizebox{\textwidth}{!}
			{
				\begin{tabular}{l|ccccccccccc|c}
					\hline
					Sequence            &      00       &      01       &      02       &      03       &      04       &      05       &      06       &      07       &      08       &      09       &      10       &    Average    \\
					FrameNum            &     4541      &     1101      &     4661      &      801      &      271      &     2761      &     1101      &     1101      &     4071      &     1591      &     1201      &       -       \\ \hline
					\bf{ALOAM/m}        &   0.7556\%    &   1.9629\%    &   4.5316\%    &   0.9507\%    & \bf{0.7201}\% &   0.5421\%    &   0.6053\%    & \bf{0.4203}\% & \bf{1.0482}\% &   0.7235\%    & \bf{1.0075}\% &   1.7318\%    \\
					LineNum             &     1051      &     1477      &     1070      &     1230      &     1304      &     1166      &     1414      &     1050      &     1190      &     1214      &     1067      &     1203      \\
					PlaneNum            &     1707      &     2865      &     1968      &     2885      &     2628      &     2015      &     3521      &     1746      &     2265      &     2298      &     1718      &     2328      \\ \hline
					\bf{ALOAM-select/m} & \bf{0.7244}\% & \bf{1.9339}\% & \bf{4.4820}\% & \bf{0.8636}\% &   0.7295\%    & \bf{0.4828}\% & \bf{0.5798}\% &   0.4493\%    &   1.0662\%    & \bf{0.6609}\% &   1.0659\%    & \bf{1.7041}\% \\
					LineNum             &   \bf{676}    &   \bf{646}    &   \bf{662}    &   \bf{658}    &   \bf{719}    &   \bf{688}    &   \bf{700}    &   \bf{676}    &   \bf{674}    &   \bf{690}    &   \bf{670}    &   \bf{678}    \\
					PlaneNum            &   \bf{1294}   &   \bf{1451}   &   \bf{1325}   &   \bf{1438}   &   \bf{1449}   &   \bf{1352}   &   \bf{1523}   &   \bf{1287}   &   \bf{1341}   &   \bf{1394}   &   \bf{1207}   &   \bf{1369}   \\ \hline
				\end{tabular}
			}
			\label{table:KITTI Subset}
		\end{center}
	\end{table*}
	\begin{table*}[htbp]
		\begin{center}
			\caption{KITTI twice potential points}
			\resizebox{\textwidth}{!}
			{
				\begin{tabular}{l|ccccccccccc|c}
					\hline
					Sequence            &      00       &      01       &      02       &      03       &      04       &      05       &      06       &      07       &      08       &      09       &      10       &    Average    \\
					FrameNum            &     4541      &     1101      &     4661      &      801      &      271      &     2761      &     1101      &     1101      &     4071      &     1591      &     1201      &       -       \\ \hline
					\bf{ALOAM/m}        &   0.7556\%    &   1.9629\%    &   4.5316\%    &   0.9507\%    &   0.7201\%    &   0.5421\%    &   0.6053\%    &   0.4203\%    &   1.0482\%    &   0.7235\%    & \bf{1.0075}\% &   1.7318\%    \\
					LineNum             &     1051      &     1477      &     1070      &     1230      &     1304      &     1166      &     1414      &     1050      &     1190      &     1214      &     1067      &     1203      \\
					PlaneNum            &     1707      &     2865      &     1968      &     2885      &     2628      &     2015      &     3521      &     1746      &     2265      &     2298      &     1718      &     2328      \\ \hline
					\bf{ALOAM-select2/m} & \bf{0.7462}\% & \bf{1.8716}\% & \bf{4.0993}\% & \bf{0.7866}\% & \bf{0.6817}\% & \bf{0.3984}\% & \bf{0.5625}\% & \bf{0.3927}\% & \bf{0.9670}\% & \bf{0.5844}\% &   1.0778\%    & \bf{1.5781}\% \\
					LineNum             &   \bf{838}    &   \bf{777}    &   \bf{758}    &   \bf{748}    &   \bf{833}    &   \bf{784}    &   \bf{816}    &   \bf{761}    &   \bf{768}    &   \bf{792}    &   \bf{755}    &   \bf{784}    \\
					PlaneNum            &   \bf{1674}   &   \bf{1418}   &   \bf{1303}   &   \bf{1414}   &   \bf{1424}   &   \bf{1327}   &   \bf{1509}   &   \bf{1255}   &   \bf{1321}   &   \bf{1370}   &   \bf{1181}   &   \bf{1381}   \\ \hline
				\end{tabular}
			}
			\label{table:KITTI Twice}
		\end{center}
	\end{table*}
	\begin{figure}[htbp]
		\centering
		\includegraphics[width=0.45\textwidth]{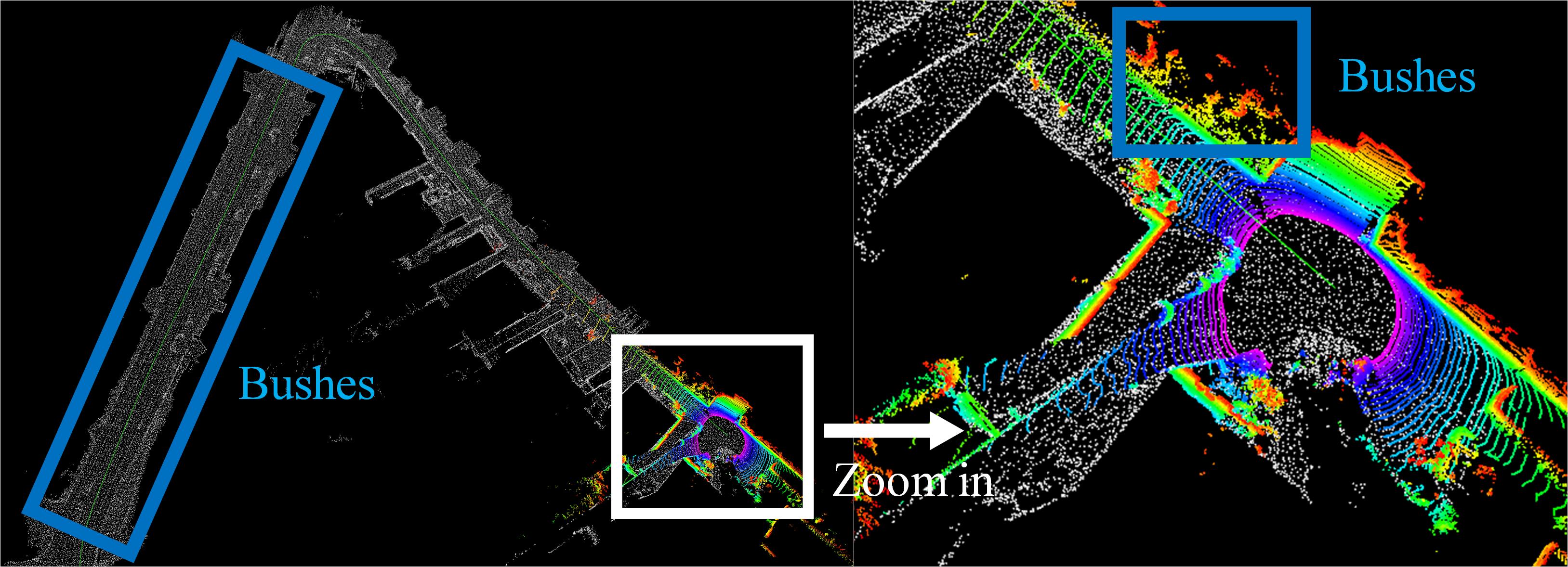}
		\caption{KITTI sequence 10: LiDAR points restricted to the local region by bushes on both sides. Scores are similar in this environment. Error matches influence the proposed selection; hence, falling behind by $7\ cm$ is possible under this extreme environment.}
		\label{figure:Sequence10}
	\end{figure}
	
	For the second test, we used twice the number of potential feature points for selection (ALOAM-select2). As summarized in TABLE \ref{table:KITTI Twice}, the accuracy improves by approximately $20\ cm$ per $100\ m$, with advancements in ten sequences. Moreover, only approximately half of the planes and lines were used to obtain this result. These two tests demonstrated the validity of our theory for improving accuracy.
	
	To identify shortages, in the second test sequence 10, the checking of the LiDAR frame points is shown in Fig. \ref{figure:Sequence10}. The car in this sequence traverses a wild-field road with bushes on both sides. In this environment, LiDAR observations were restricted to a local region. Our calculations yielded similar results. The proposed algorithm trades off the robustness to achieve accuracy. Error matches have a stronger influence; thus, falling behind $7\ cm$ is possible in this extreme environment.
	
	MLOAM \cite{2021ICRAmloam} was designed for multi-LiDAR systems. The results are compared on KITTI, as summarized in TABLE \ref{table:KITTI MLOAM}. Regarding the proposed method, only Sequence 02 falls behind; the other sequences have better results. MLOAM's selection scheme defines manual prior information, solves a metric Max-$log$Det, and then derives residuals that persist. Our proposed theory considers the sensitivity and uncertainty of sensor data, which are closer to the natural process of an LO. This enables more accurate measurements and map models to improve accuracy and avoids the calculation of matrix determinants. For sequence 02, the LiDAR goes through a crossroad with fast turning, and the proposed method drifts significantly from this location. From that point onward, it was considerably misled, resulting in a large error in the total path.
	
	MLOAM was originally intended for multi-LiDAR sensors, and we modified its code for one LiDAR running. Considering fairness and limitations on the article length, the next section's comparison focuses on the single LiDAR algorithm.
	\begin{table*}[htbp]
		\begin{center}
			\caption{KITTI MLOAM}
			\resizebox{\textwidth}{!}
			{
				\begin{tabular}{l|ccccccccccc|c}
					\hline
					Sequence     &      00       &      01       &      02       &      03       &      04       &      05       &      06       &      07       &      08       &      09       &      10       &    Average    \\
					FrameNum     &     4541      &     1101      &     4661      &      801      &      271      &     2761      &     1101      &     1101      &     4071      &     1591      &     1201      &       -       \\ \hline
					\bf{MLOAM/m} &   1.7015\%    &   2.3043\%    & \bf{2.3271}\% &   1.0544\%    &   1.1347\%    &   0.8285\%    &   1.4445\%    &   1.4053\%    &   1.0679\%    &   1.5106\%    &   1.9189\%    &   1.6152\%    \\ \hline
					\bf{ALOAM-select2/m} & \bf{0.7462}\% & \bf{1.8716}\% &   4.0993\%    & \bf{0.7866}\% & \bf{0.6817}\% & \bf{0.3984}\% & \bf{0.5625}\% & \bf{0.3927}\% & \bf{0.9670}\% & \bf{0.5844}\% & \bf{1.0778}\% & \bf{1.5781}\% \\ \hline
				\end{tabular}
			}
			\label{table:KITTI MLOAM}
		\end{center}
	\end{table*}
	
	\subsubsection{Time Cost}
	\begin{figure*}[htbp]
		\centering
		\subfigure[]{
			\includegraphics[width=0.3\textwidth]{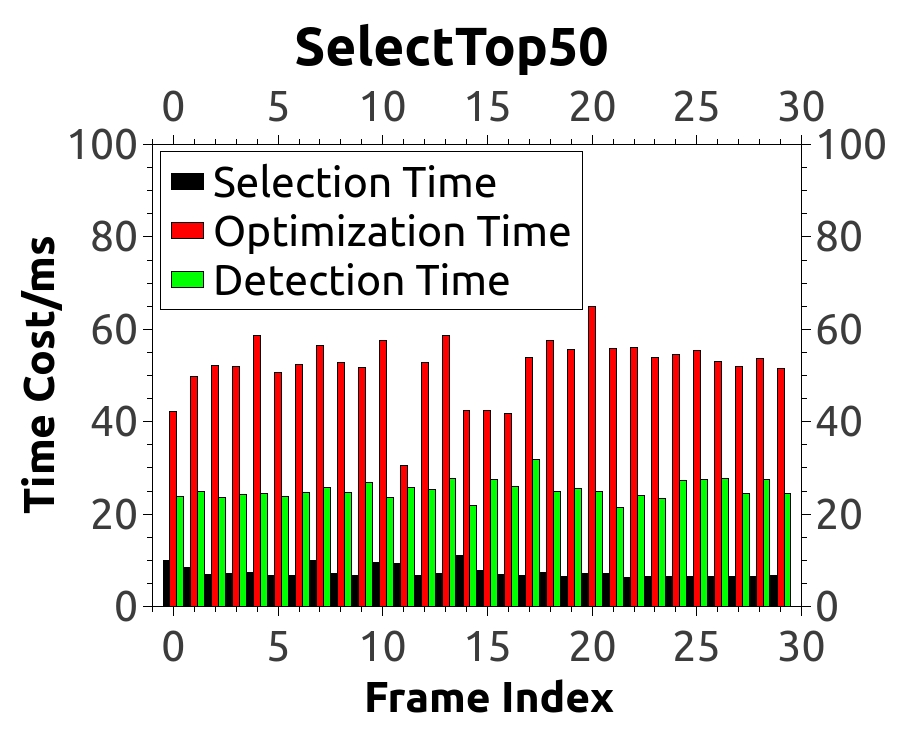}}
		\subfigure[]{
			\includegraphics[width=0.3\textwidth]{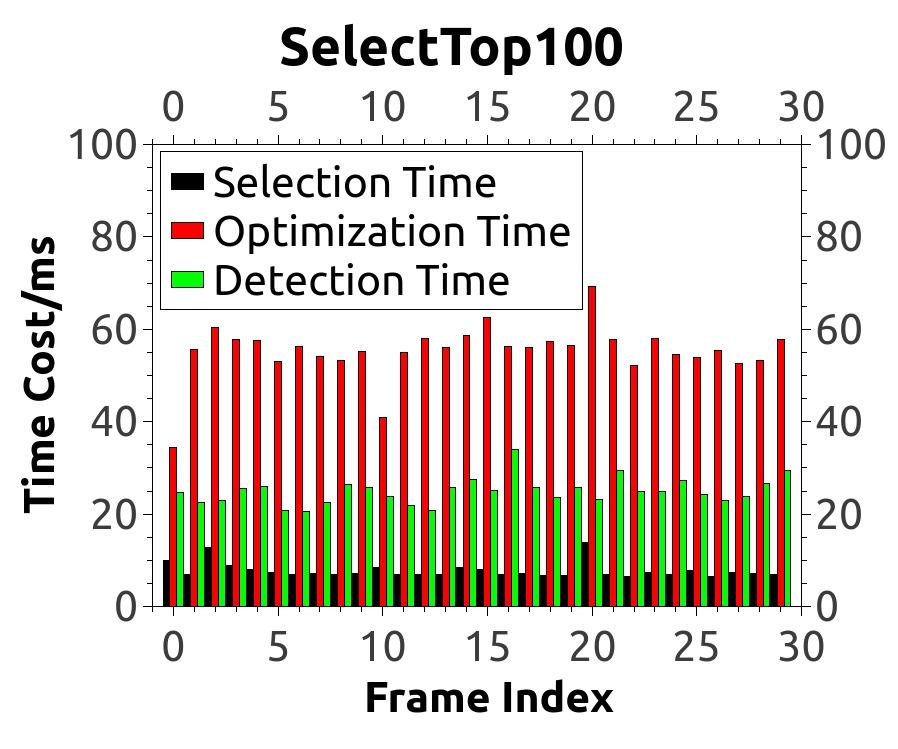}}
		\subfigure[]{
			\includegraphics[width=0.3\textwidth]{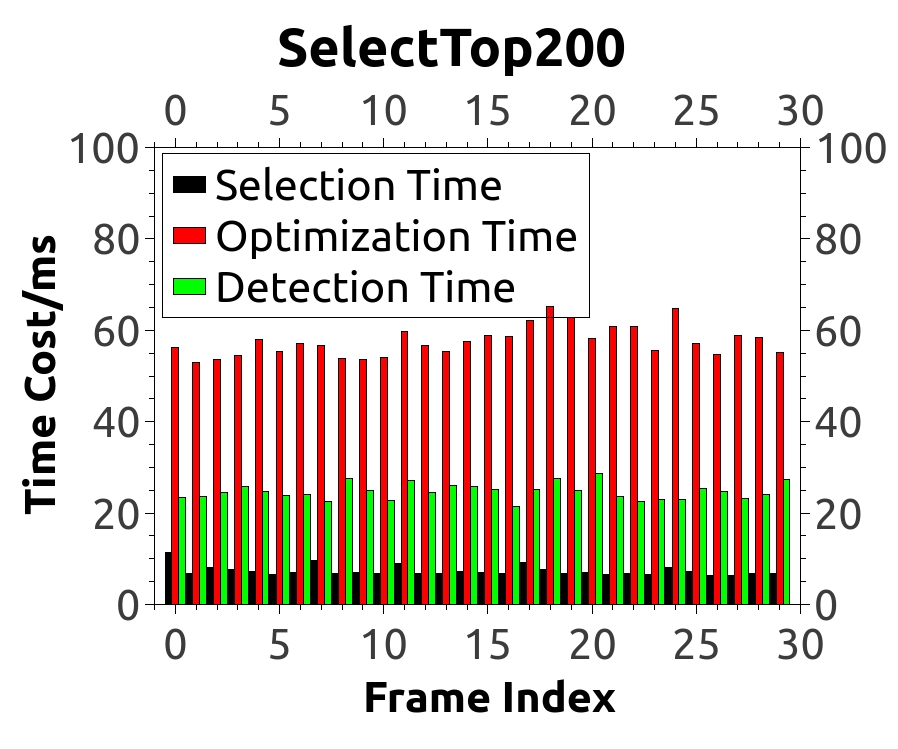}}
		\subfigure[]{
			\includegraphics[width=0.3\textwidth]{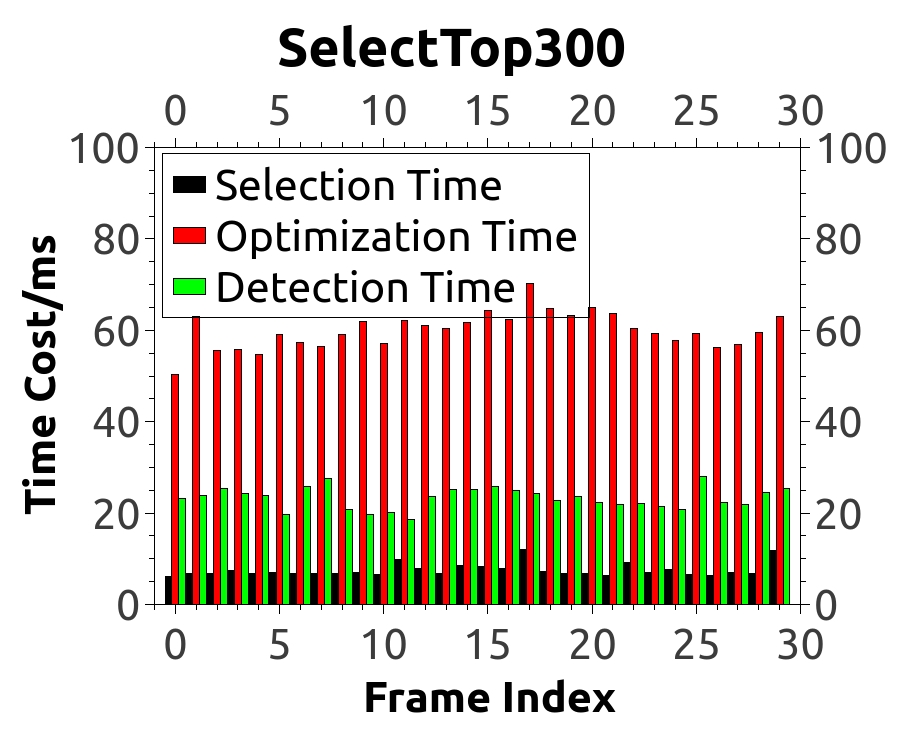}}
		\subfigure[]{
			\includegraphics[width=0.3\textwidth]{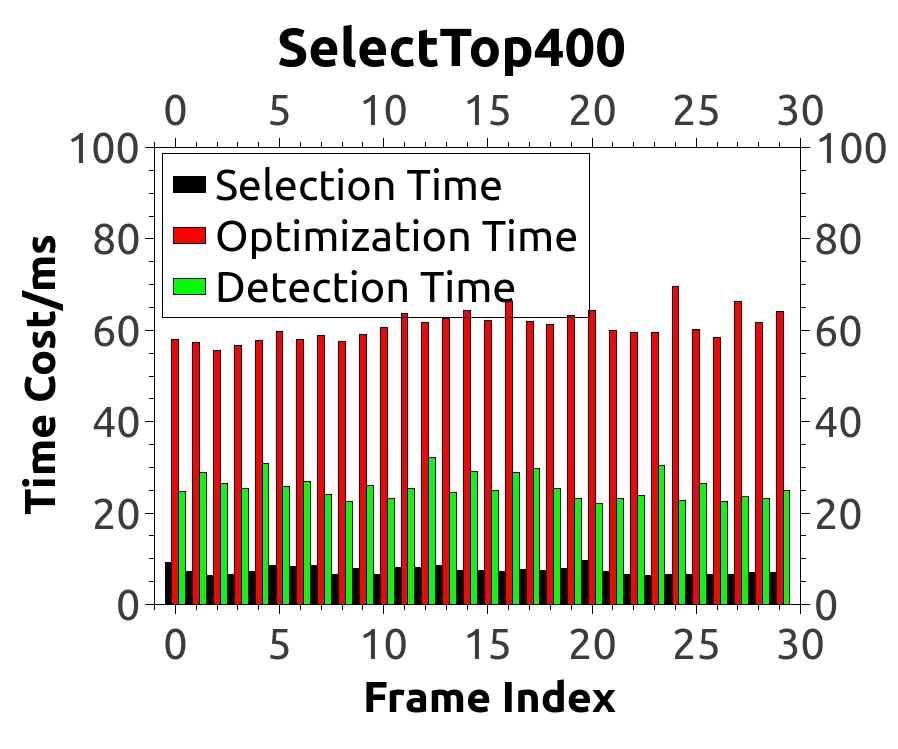}}
		\subfigure[]{
			\includegraphics[width=0.3\textwidth]{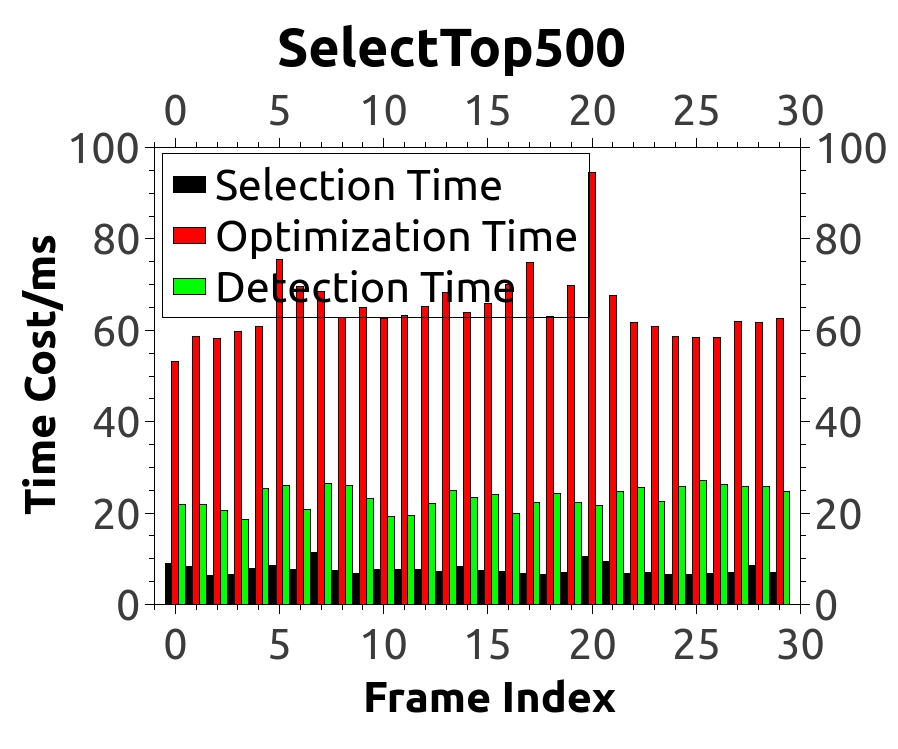}}
		\caption{Black is selection time, which is within almost $15\ ms$. The green is detection time, which is constant in comparisons. The red is optimization time. Provided that increasing of the residual selection amount per dimension, is from $50$ to $500$; more residuals are selected and entered into optimization, which makes it more time-consuming.}
		\label{figure:TimeCost}
	\end{figure*}
	To verify the effect of the selection part on the overall LO time performance, the time costs of the main part are shown in Fig. \ref{figure:TimeCost}. Compared with feature point detection (green) and residual optimization (red), our selection process (black) costs less than $15\ ms$. The optimization part accounts for a large proportion of the time cost. Provided that the selected residual amount per dimension decreases from $500$ to $50$, the optimization becomes faster.
	
	Some coding tricks have been introduced here to illustrate why the selection parts require less time. If an LO/LIO algorithm adopts our selection, its accuracy can be improved, and more computation time can be achieved. First, not every residual must participate in the sorting. When calculating the score, the maximal scores in $6$ dimensions were recorded. Residuals scores are higher than the ratio threshold ($60\%$ we adopt) of maximal retention sorting. Second, point-to-plane and point-to-line are both independent in $6$ dimensions; thus, multithread parallel operations accelerate sorting.
	
	Therefore, the optimization part requires time, which is related to the residual amount. If we utilize our selected residuals, compared with using all obtainable residuals, the residual term amount decreases significantly, and the selection process is still lightweight. The pose estimation accuracy is simultaneously improved in the next section, and we use residual amounts to illustrate the computation cost.
	
	\subsection{Online captured scenario}
	\label{subsection:Online Captured Scenario}
	\begin{figure}[htbp]
		\centering
		\includegraphics[width=0.45\textwidth]{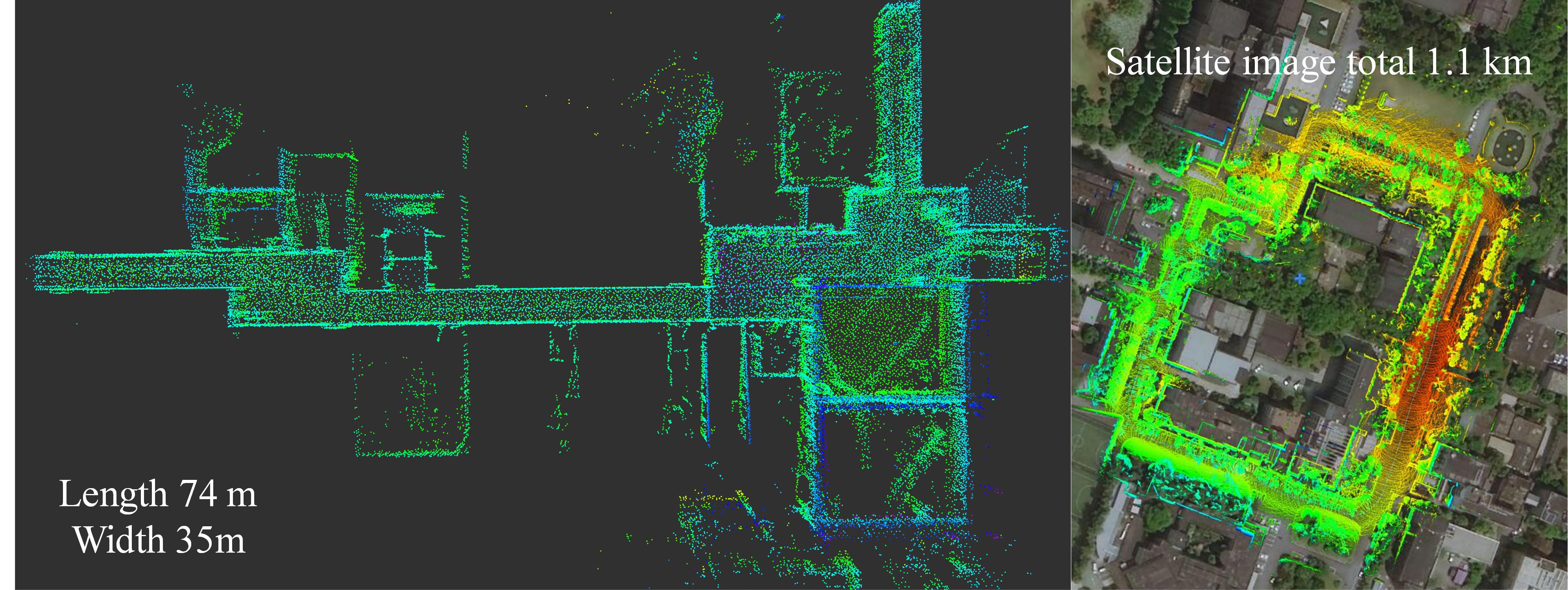}
		\caption{Our online captured scenarios are shown in Fig. \ref{figure:Scenario}. The indoor environment is a building with a long corridor. Its length is $74m$, and width is $35m$.The outdoor path is approximately $1.1 km$ long. Both capturing tours start and end at exactly the same location, and execute five times for fairness.}
		\label{figure:Scenario}
	\end{figure}
\begin{figure}[htbp]
		\centering
		\subfigure[Collection device: Velodyne LiDAR (Puck VLP16) and IMU (Xsens MTI-100), Robosense LiDAR (Blind Spot 32)]
		{
			\label{figure:VLP16Xsense BS LiDAR}
			\includegraphics[width=0.45\textwidth]{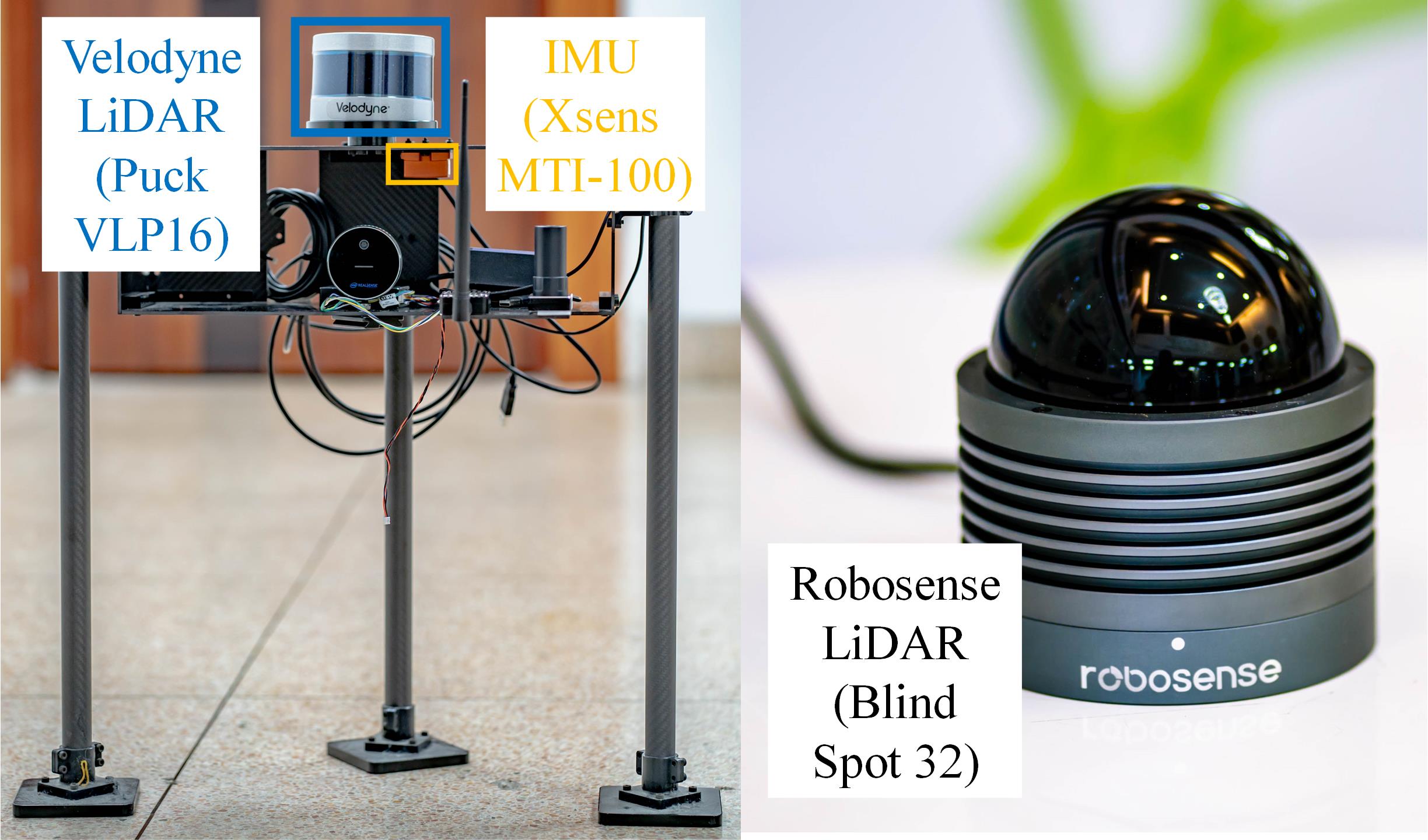}
		}
		\subfigure[These two type of LiDAR sensors have different scan modes.]
		{
			\label{figure:VLP16Xsense BS LiDAR Points}
			\includegraphics[width=0.45\textwidth]{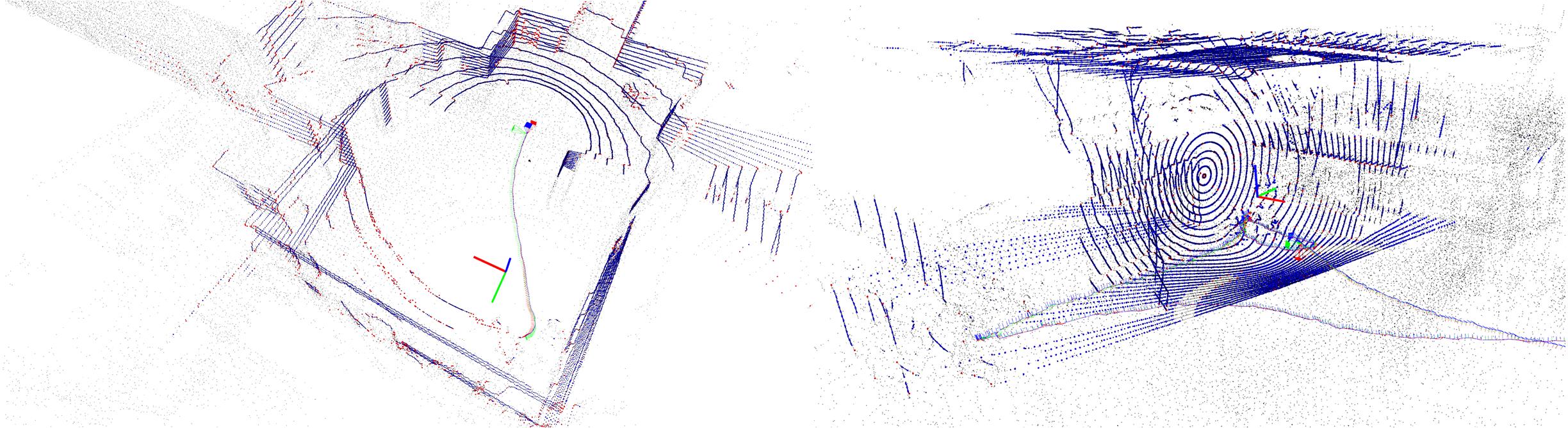}
		}
		\caption{Two collection devices and their different scan modes.}
	\end{figure}

	We used our sensors to operate in real environments to illustrate the details. The captured online scenarios are shown in Fig. \ref{figure:Scenario}. The indoor environment includes walking inside a building with a long corridor. The outdoor path was approximately $1.1 km$ long. By pasting a landmark on the ground, both tours start and end at exactly the same location. The same path is captured five times to ensure fairness and credibility. Two collection devices were used during capture, as shown in Fig. \ref{figure:VLP16Xsense BS LiDAR}. The first was a Velodyne LiDAR (Puck VLP16) with an IMU (Xsens MTI-100). The second is Robosense LiDAR (Blind Spot 32). As shown in Fig. \ref{figure:VLP16Xsense BS LiDAR Points}, these two LiDARs have completely different scan modes. The horizon of VLP16 was $360 \degree$, and the BS LiDAR was a half-sphere window with 32 scans.
	
	We apply our method to ALOAM and LIOmapping as ALOAM-select and LIOmapping-select, respectively. We both set the same strategy and parameters: (1) stop when the residual amount reaches a maximum of 200 per dimension; (2) stop when the residual score decreases to $10\%$ of the maximum. For convenience, we focused on the pose accuracy using the loop-closure error.
	
	\subsubsection{VLP16 LO indoor}
	\begin{table}[htbp]
		\begin{center}
			\caption{VLP16 LO indoor}
			\resizebox{0.5\textwidth}{!}
			{
				\begin{tabular}{l|ccccc}
					\hline
					Environment       & \multicolumn{5}{|c}{indoor (loop closure error/m)}                  \\ \hline
					Sequence          &     00      &     01      &     02      &     03      &     04      \\ \hline
					\bf{ALOAM}        &   0.0334    & \bf{0.0163} &   0.1131    &   0.0338    &   6.7570    \\ \hline
					\bf{ALOAM-select} & \bf{0.0274} &   0.0184    & \bf{0.0654} & \bf{0.0162} & \bf{0.0400} \\ \hline
					&             &             &  \\ \hline
					\bf{BALM}         &   0.0278    & \bf{0.0091} &   0.0597    &   0.0303    &   0.1471    \\ \hline
					\bf{BALM-select}  & \bf{0.0261} &   0.0146    & \bf{0.0534} & \bf{0.0244} & \bf{0.0846} \\ \hline
				\end{tabular}
			}
			\label{table:VLP16 LO Indoor}
		\end{center}
	\end{table}
	
\begin{figure}[htbp]
		\centering
		\includegraphics[width=0.45\textwidth]{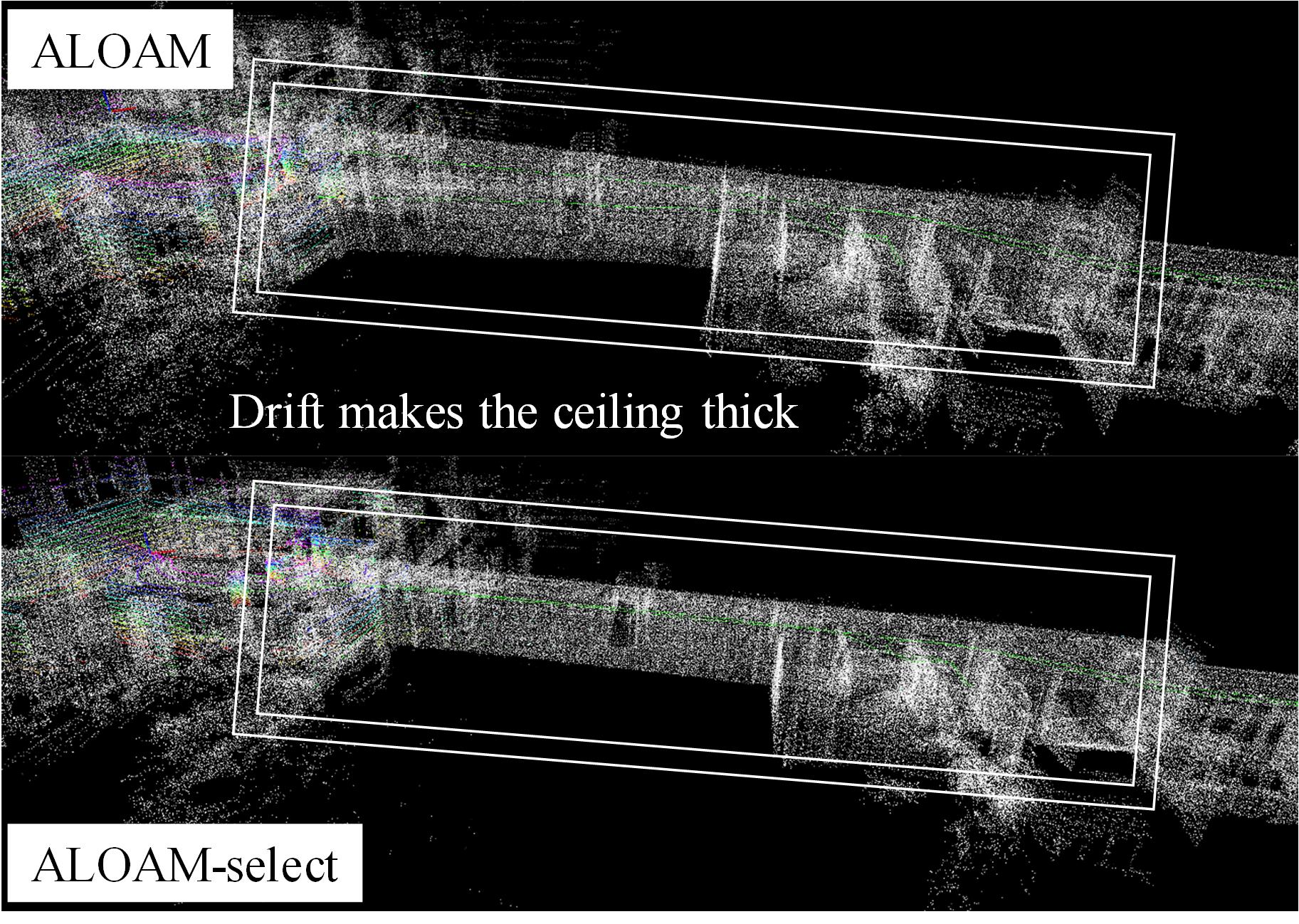}
		\caption{Sequence 04 involves walking into a restroom. More surrounding points are used by ALOAM, but the ALOAM-select uses more points outside the door and window, which are more sensitive.}
		\label{figure:LOsq04}
	\end{figure}

	The indoor loop closure errors are summarized in TABLE \ref{table:VLP16 LO Indoor}. Because the LiDAR range was sufficiently long to measure the farthest wall, the drift was within the centimeter level. Compared with ALOAM, ALOAM-select achieved better results in the four sequences. In sequence 01, because ALOAM's translation is virtually $1\ cm$, we believe that the accuracy of the start and end locations of this sequence is insufficient for evaluation. Sequence 04 involved walking in the restroom. More surrounding points are used by ALOAM, but ALOAM-select uses more points outside the door and window, which are more sensitive. The ALOAM drift was distinct, whereas that of the ALOAM-select was low. This result verifies the validity of the electioproposed sn scheme. The paths, maps, and drifts are shown in Figs. \ref{figure:LOsq04}.
	\begin{figure}[htbp]
		\centering
		\includegraphics[width=0.5\textwidth]{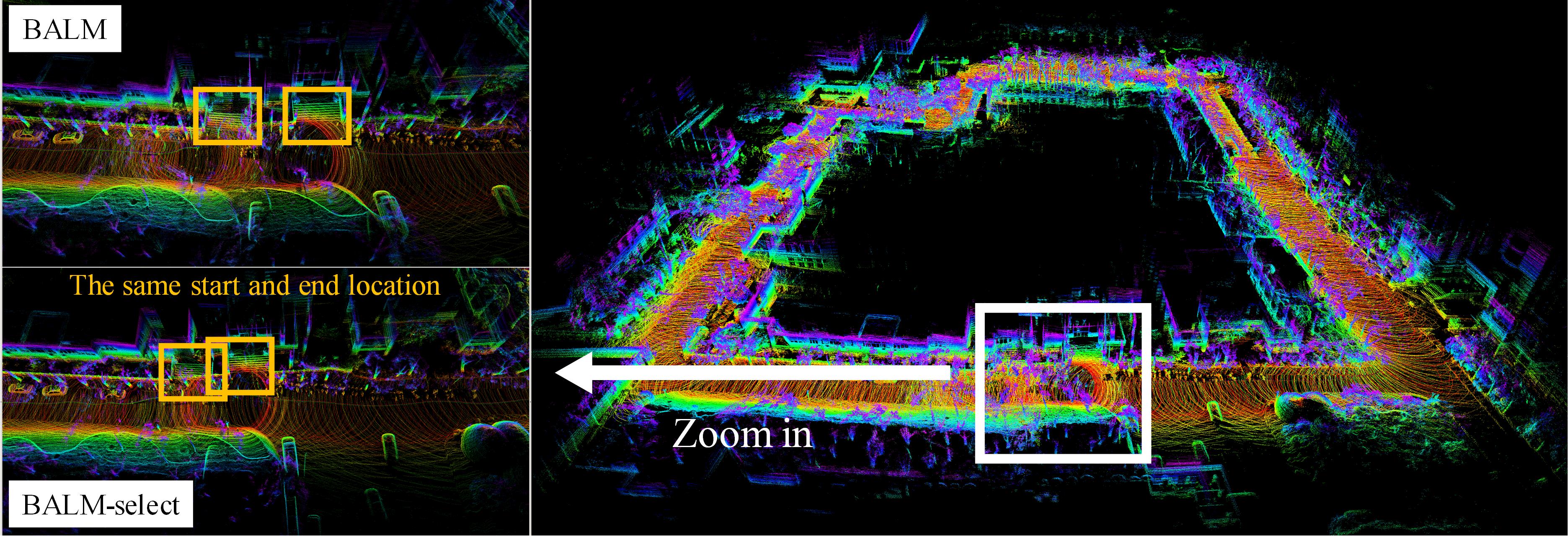}
		\caption{Drift of BALM is shown in the start and end locations, which should be exactly the same. The drift of BALM-select is smaller, as seen from the building steps.}
		\label{figure:LOsq04BA}
	\end{figure}
	
	Furthermore, we compared it with the recent famous bundle-adjustment LiDAR mapping algorithm, BALM \cite{2021BALM} in TABLE \ref{table:VLP16 LO Indoor}. The BALM adopts the bundle adjustment concept for visual SLAM. It has a sliding window and adjusts inside frame poses, which aims to make voxels more compact, such as planes flatter and lines more slender. This is also the same idea in eigenfactor \cite{2019EigenFactors}, plane-adjustment \cite{2021PlaneAdjustmentLiDAR}, and $\pi$-LSAM \cite{2021pi-LSAM}. The BALM code was originally designed for the VLP16 LiDAR. We applied our selection scheme in front of the decision regarding, which voxels enter the optimization; thus, we called it BALM-select. Because the inside of the building has many smooth wall constraints, the loop-closure errors are significantly small. Because of the sliding windows in the bundle adjustment, the indoor environment is insufficiently large for LiDAR sensors. Current frame observations may be connected to early information, which strongly constrains the sensor pose. Therefore, after comparing the indoor data, the BALM accuracy disparity was not as obvious as that of ALOAM.
	
	\subsubsection{VLP16 LO outdoor}
	\begin{table}[htbp]
		\begin{center}
			\caption{VLP16 LO outdoor}
			\resizebox{0.5\textwidth}{!}
			{
				\begin{tabular}{l|ccccc}
					\hline
					Environment       & \multicolumn{5}{|c}{outdoor (loop closure error/m)}                 \\ \hline
					Sequence          &     00      &     01      &     02      &     03      &     04      \\ \hline
					\bf{ALOAM}        &   7.8131    &   10.0851   &   8.0538    &   5.0375    &   6.9043    \\ \hline
					\bf{ALOAM-select} & \bf{5.0001} & \bf{5.8372} & \bf{4.4030} & \bf{3.3837} & \bf{5.4447} \\ \hline
					& \\ \hline
					\bf{BALM}         &   6.3223    &   11.7821   &   6.2510    &   4.8214    &   5.9736    \\ \hline
					\bf{BALM-select}  & \bf{4.6401} & \bf{6.3965} & \bf{4.1028} & \bf{2.0205} & \bf{4.2007} \\ \hline
				\end{tabular}
			}
			\label{table:VLP16 LO Outdoor}
		\end{center}
	\end{table}
	
\begin{figure}[htbp]
		\centering
		\includegraphics[width=0.45\textwidth]{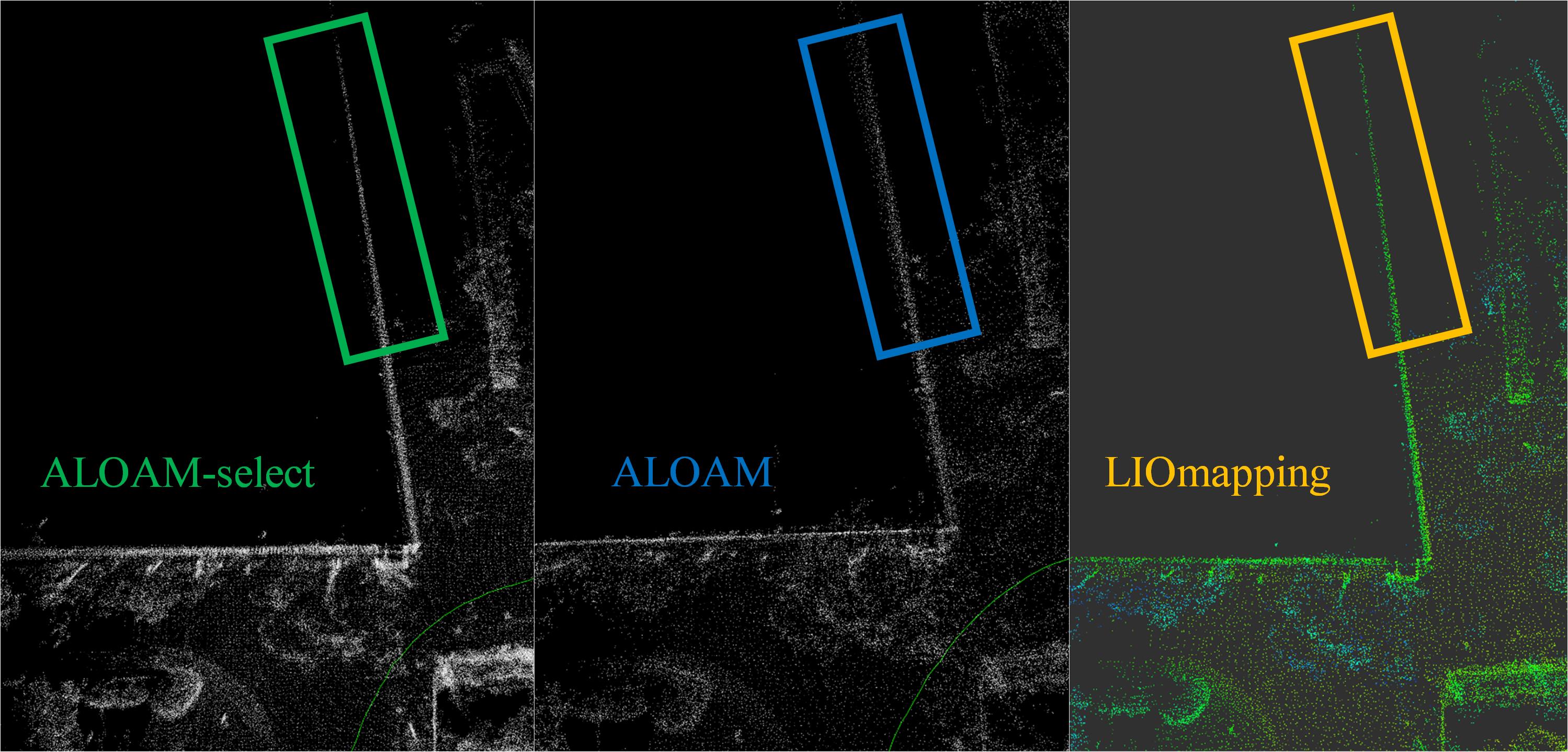}
		\caption{Building map quality. Points in the blue rectangle represent the ALOAM building wall; points in the green rectangle are generated by ALOAM-select, and points in the orange rectangle are the LIOmapping results. The ALOAM building wall is thick, indicating that the estimated LiDAR poses have drifted. Wall thickness generated by the proposed method is as thin as that of LIOmapping, indicating higher accuracy; only LiDAR is used to reach this LiDAR with the IMU level.}
		\label{figure:Wall}
	\end{figure}

\begin{figure}[htbp]
		\centering
		\includegraphics[width=0.45\textwidth]{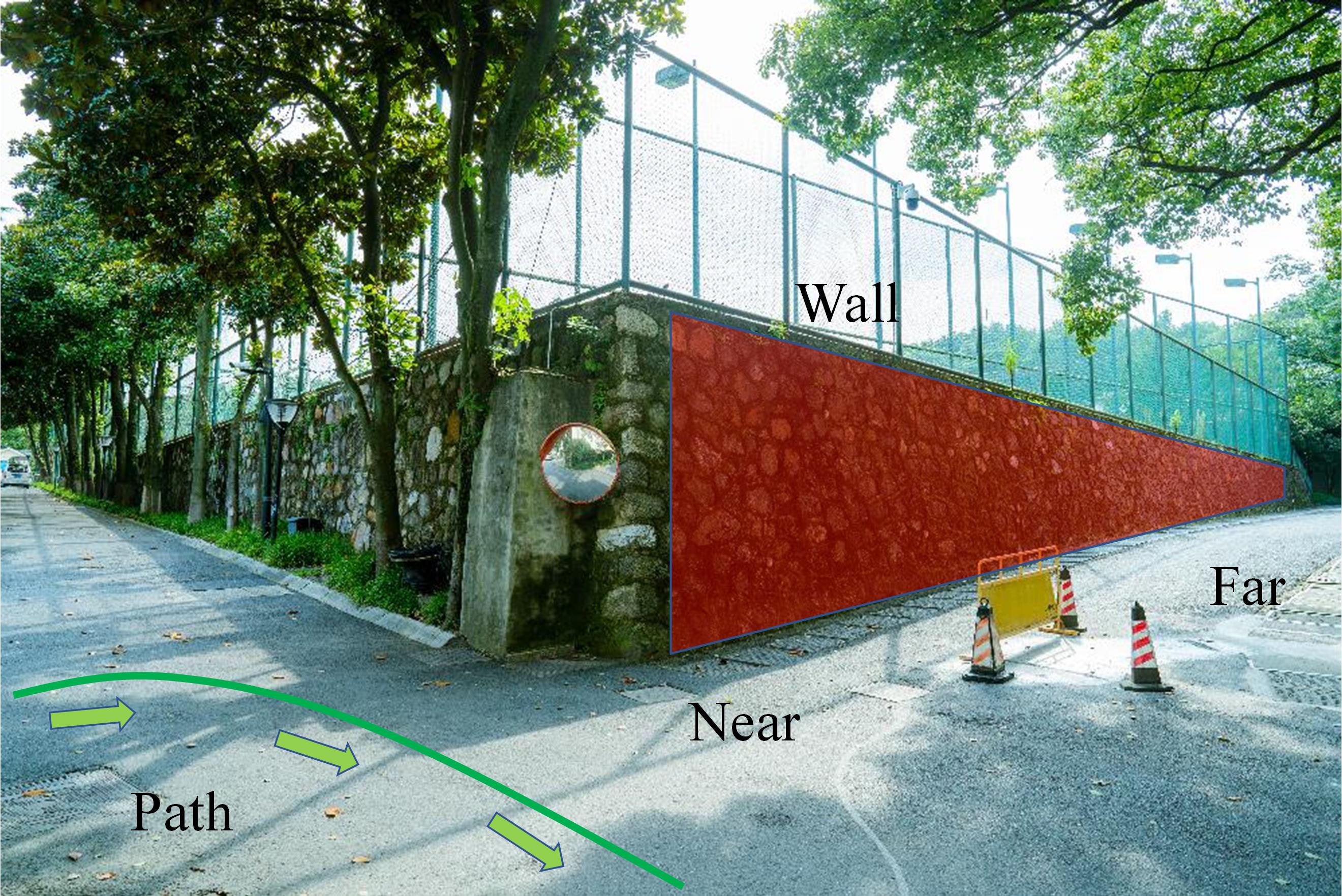}
		\caption{Real wall scene: LiDAR moving path, the scanned wall, far and near the scanning surface.}
		\label{figure:WallScene}
	\end{figure}

\begin{figure}[htbp]
		\centering
		\includegraphics[width=0.5\textwidth]{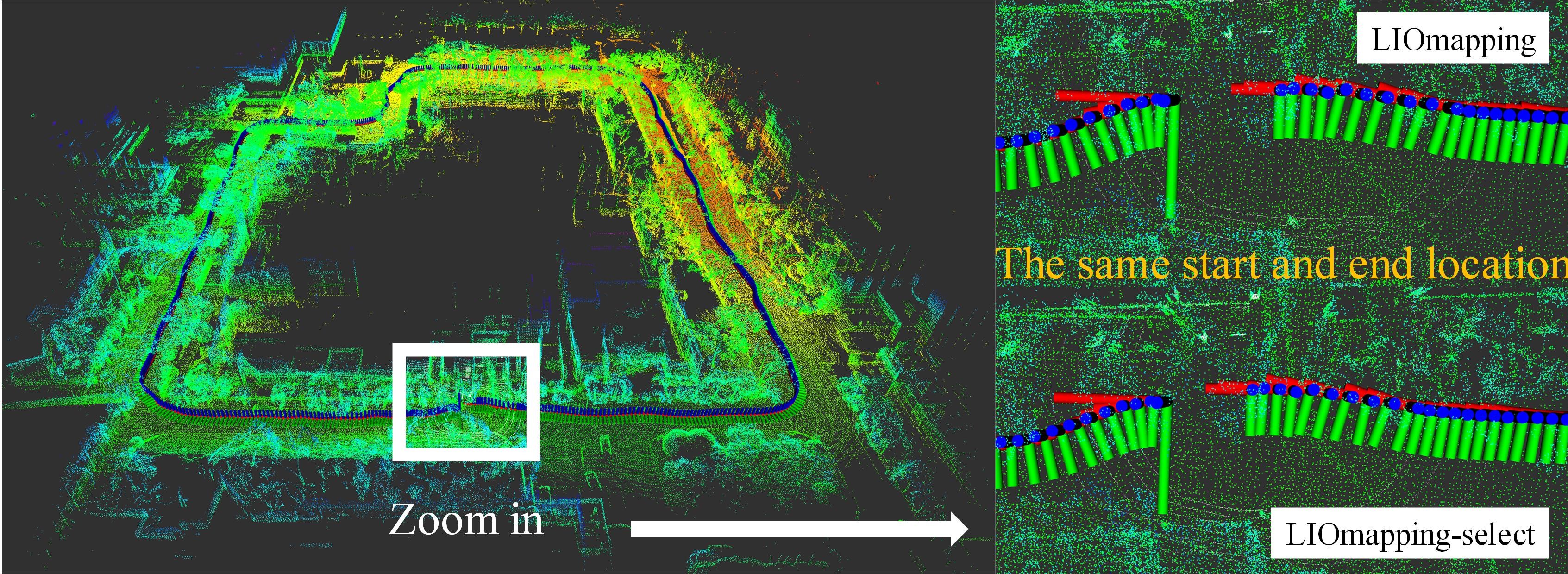}
		\caption{After fusing IMU data, the LIO algorithm's loop closure error is significantly smaller than LO. However, from the start and end location data, before and after adding selection's drifts are still visible.}
		\label{figure:LIOLoopClosureError}
	\end{figure}

	The outdoor loop closure errors are summarized in TABLE \ref{table:VLP16 LO Outdoor}. The long outdoor path demonstrates the superiority of the ALOAM-select and BALM-select. This path is approximately $1.1\ km$-long circle in Fig. \ref{figure:LIOLoopClosureError}. The proposed method achieved almost twice the accuracy of all five sequences. These results further support the analysis of the proposed method using the KITTI benchmark dataset. Under large-scale conditions, the proposed method achieves more significant results. The wall mapping quality is illustrated in Fig. \ref{figure:Wall} and \ref{figure:WallScene}. In Fig. \ref{figure:Wall}, points in the blue rectangle represent the ALOAM building wall; points in the green rectangle are generated by ALOAM-select, and points in the orange rectangle are LIOmapping result. Fig. \ref{figure:WallScene} shows the LiDAR's moving path, and the scanned wall. The ALOAM building wall is thick, indicating that the estimated LiDAR poses a drift. The wall generated by ALOAM-select is as thin as that of LIOmapping, indicating higher accuracy; here, only LiDAR is used to reach this LiDAR with an IMU level.
	
	\subsubsection{VLP16 LIO Indoor}
	\begin{table}[htbp]
		\begin{center}
			\caption{VLP16 LIO indoor}
			\resizebox{0.5\textwidth}{!}
			{
				\begin{tabular}{l|ccccc}
					\hline
					Environment     & \multicolumn{5}{|c}{indoor (loop closure error/m)}                  \\ \hline
					Sequence        &     00      &     01      &     02      &     03      &     04      \\ \hline
					\bf{LIOmapping} & \bf{0.0332} &   0.0399    &   0.0297    & \bf{0.0363} & \bf{0.0163} \\ \hline
					\bf{LIOmapping-select}       &   0.0335    & \bf{0.0158} & \bf{0.0198} &   0.0380    &   0.0359    \\ \hline
				\end{tabular}
			}
			\label{table:VLP16 LIO Indoor}
		\end{center}
	\end{table}
	The loop closure error is small for LIO because the indoor path is less than that of the outdoor path; the results are summarized in TABLE \ref{table:VLP16 LIO Indoor}. The path starts from a hall, and the front side of the corridor can be observed. The LiDAR data are repeated on the surrounding walls. The results of LIOmapping and LIOmapping-select were similar. We presumed that the small region restricted the improvement in accuracy.
	
	\subsubsection{VLP16 LIO outdoor}
	\begin{table}[htbp]
		\begin{center}
			\caption{VLP16 LIO outdoor}
			\resizebox{0.5\textwidth}{!}
			{
				\begin{tabular}{l|ccccc}
					\hline
					Environment     & \multicolumn{5}{|c}{outdoor (loop closure error/m)}                 \\ \hline
					Sequence        &     00      &     01      &     02      &     03      &     04      \\ \hline
					\bf{LIOmapping} &   3.6719    &   4.5632    &   5.0961    &   4.2081    &   3.6909    \\ \hline
					\bf{LIOmapping-select}       & \bf{3.2095} & \bf{3.6629} & \bf{4.2257} & \bf{4.0576} & \bf{3.1955} \\ \hline
				\end{tabular}
			}
			\label{table:VLP16 LIO Outdoor}
		\end{center}
	\end{table}
	The outdoor results are summarized in TABLE \ref{table:VLP16 LIO Outdoor}. The average improvement in accuracy of LIOmapping-select was approximately $0.5\ m$. Therefore, the application of our theory to LIO systems is also valuable. After fusing the IMU data, the loop closure error of the LIO algorithm was significantly smaller than that of the LO. However, from the start and end location data, before and after the addition of drifts in the selection are still visible. Because LIO mapping does not have a loop-closing function, this error cannot be eliminated. After the addition of selection, LIOmapping-select tends to use far-away observations, which strongly constrain the sensor pose. Thus, the estimation accuracy was higher on average.
	
	\subsubsection{Blind spot LO indoor}
	\begin{figure}[htbp]
		\centering
		\includegraphics[width=0.45\textwidth]{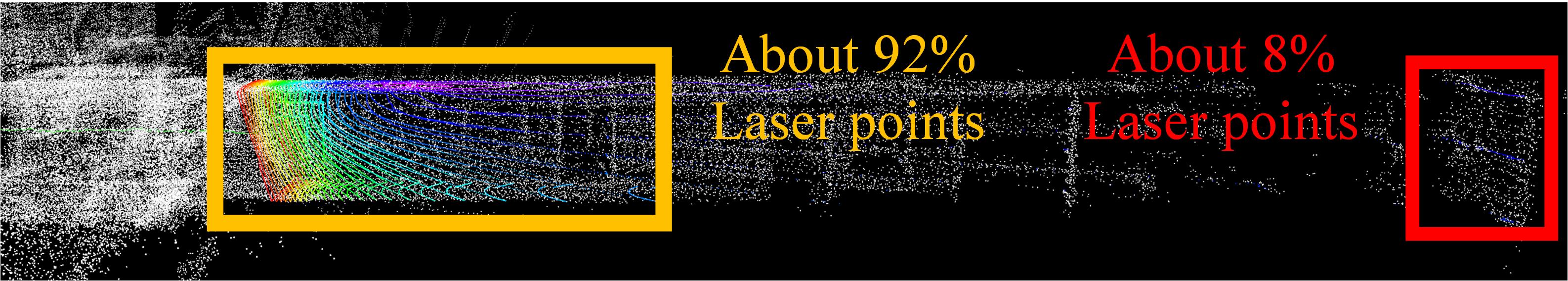}
		\caption{BS LiDAR point distribution considerably differs from that of Velodyne VLP16. Many points lie in a small region within $10\ m$ in front of LiDAR. However, distant points on the wall are more suitable for pose estimation in the forwarding moving direction.}
		\label{figure:BS Points Distribution}
	\end{figure}
	\begin{figure}[htbp]
		\centering
		\includegraphics[width=0.45\textwidth]{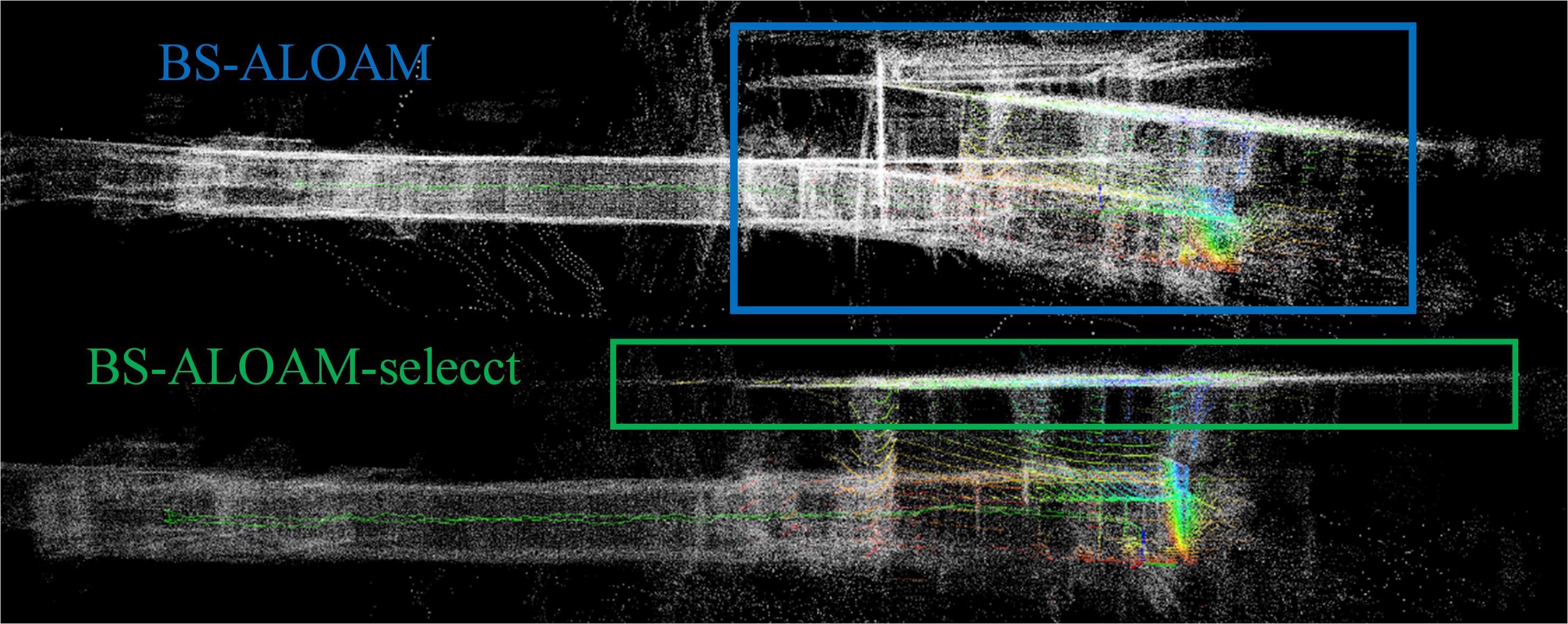}
		\caption{BS LiDAR indoor built map. Ground floor and ceiling generated by BS-ALOAM are distorted, because poses drifts are at meter level. BS-ALOAM-select maintains with considerably lower drift.}
		\label{figure:Indoor Blind Spot}
	\end{figure}
	Another scan mode, the Robosense blind spot (BS) LiDAR, is shown in Fig. \ref{figure:VLP16Xsense BS LiDAR}. Its view was a half-sphere with 32 laser scans from $0 \degree$ (horizontal) to $89 \degree$ (vertical). Its direction is set toward the ceiling, which is less likely to be scanned by VPL16. To fit the ALOAM code, we modified the point allocation and feature detection functions and renamed them to BS-ALOAM.
	
	Upon entering the corridor, the BS LiDAR was placed toward the front to measure more points. In Fig. \ref{figure:BS Points Distribution}, the BS LiDAR point distribution shows extremely different from that of VLP16. Virtually, $92\%$ of the laser points lie in a small region within $10\ m$ of the surroundings. However, points that are distant from the wall are more useful. In BS-ALOAM selection, more suitable points are selected. The results are summarized in TABLE \ref{table:Blind Spot LO Indoor}; the results of the proposed method are distinct. The accuracy of the BS-ALOAM-select improved from the meter to decimeter level. A building map is shown in Fig. \ref{figure:Indoor Blind Spot}. When we return to the hall, the floor and ceiling map of the BS-ALOAM is distorted because the pose drifts are at the meter level. The BS-ALOAM-select maintained a considerably lower drift.
	\begin{table}[htbp]
		\begin{center}
			\caption{BS LO indoor}
			\resizebox{0.5\textwidth}{!}
			{
				\begin{tabular}{l|ccccc}
					\hline
					Environment&\multicolumn{5}{|c}{indoor (loop closure error/m)}\\
					\hline
					Sequence&00&01&02&03&04\\
					\hline
					\bf{BS-ALOAM} &      2.9735 &      3.2668 &      3.2795 &      2.1737 &      5.0697 \\
					\hline
					\bf{BS-ALOAM-select}  & \bf{0.1301} & \bf{0.1012} & \bf{0.0886} & \bf{0.7348} & \bf{0.1299} \\
					\hline
				\end{tabular}
			}
			\label{table:Blind Spot LO Indoor}
		\end{center}
	\end{table}
	
	\subsubsection{BlindSpot LO Outdoor}
	\begin{table}[htbp]
		\begin{center}
			\caption{BS LO outdoor}
			\resizebox{0.5\textwidth}{!}
			{
				\begin{tabular}{l|ccccc}
					\hline
					Environment&\multicolumn{5}{|c}{outdoor (loop closure error/m)}\\
					\hline
					Sequence&00&01&02&03&04\\
					\hline
					\bf{BS-ALOAM} &      4.4345 &      19.3979 &      17.8245 &      20.5416 & \bf{43.1282} \\
					\hline
					\bf{BS-ALOAM-select}  & \bf{3.1535} & \bf{13.5462} & \bf{15.1132} & \bf{12.7554} &       52.5632 \\
					\hline
				\end{tabular}
			}
			\label{table:Blind Spot LO Outdoor}
		\end{center}
	\end{table}
	Because of the BS LiDAR scan characteristics, distant points are extremely sparse, and near points are considered dense. According to our theory, the estimation accuracy is significantly lower than that of VLP16. The results summarized in TABLE \ref{table:Blind Spot LO Outdoor} explain this phenomenon. The translation error exceeded that of VLP16. Although certain improvements were achieved, the resulting errors can be ignored. Thus, BS LiDAR is unsuitable for outdoor SLAM applications. SLAM requires a LiDAR sensor capable of capturing distant points, which is more favorable for estimation.
	
	\section{Conclusion}
	\label{section:Conclusion}
	In this paper, we proposed a theory of LiDAR point sensitivity and uncertainty to enhance LiDAR odometry accuracy. We demonstrated that our selection method is a global statistical optimal. To explain this realization, LiDAR measurement uncertainties and fusing mechanisms were calculated, and residual sensitivities were analyzed. The scores were decoupled into six dimensions. Thereafter, the algorithm sorted and selected the residuals for optimization. The experiment results revealed that superior pose estimation accuracy was achieved. This selection makes it possible to simultaneously achieve high optimization accuracy and guarantee real-time performance.
	
	Owing to laser time-of-flight sensing and careful rotary mechanism calibration, the LiDAR uncertainty region does not grow as large as that of the binocular cameras. Therefore, LiDAR had a more distinct effect on our theory. The problem of data association has not yet been addressed. This work adopted traditional data association methods in the LO, relying on a uniform motion model or IMU, which is the neighborhood principle in ICP. Because this study concentrates on improving the pose estimation accuracy, a uniform motion model for walking or low-speed driving is sufficient. The proposed theory attempts to select residual terms with small uncertainties and high sensitivities. This fundamentally decreases the robustness of the pose estimation and simultaneously increases its accuracy; this is the reason for the tradeoff between robustness and accuracy. To improve the accuracy of pose estimation from another perspective, our next objective is to investigate data association.
	
	\appendix
	\section*{A.ICP-SVD}
	\begin{equation}
		\begin{aligned}
			\mathbf{R}^* & =\mathop{\arg\min}\limits_{\mathbf{R}^*\in\rm{SO}(3)}\sum_{i=1}^N ||\mathbf{R}^*\mathbf{p}^*_i-\mathbf{q}^*_i||^2                                                                                                  \\
			& =\mathop{\arg\min}\limits_{\mathbf{R}^*\in\rm{SO}(3)}\sum_{i=1}^N ({\mathbf{q}^*_i}^T \mathbf{q}^*_i+{\mathbf{p}^*_i}^T {\mathbf{R}^*}^T \mathbf{R}^*\mathbf{p}^*_i-2{\mathbf{q}^*_i}^T\mathbf{R}^*\mathbf{p}^*_i) \\
			& =\mathop{\arg\max}\limits_{\mathbf{R}^*\in\rm{SO}(3)}\sum_{i=1}^N ({\mathbf{q}^*_i}^T\mathbf{R}^*\mathbf{p}^*_i)                                                                                                   \\
			& =\mathop{\arg\max}\limits_{\mathbf{R}^*\in\rm{SO}(3)} tr(\mathbf{R}^*\sum_{i=1}^N \mathbf{p}^*_i {\mathbf{q}^*_i}^T)
		\end{aligned}
	\end{equation}
	\begin{center}
		\bf{Q.E.D.}
	\end{center}
	
	\section*{B.Theorem \ref{theorem:Riemannian Distance} Riemannian Distance}
	Utilizing angle axis Eq. (\ref{equation:Angle Axis}) to Eq. (\ref{equation:Rotation Under Disturbance}) yields
	\begin{equation}
		\label{equation:Riemannian Distance 1}
		\begin{aligned}
			{\mathbf{R}^*}^T\mathbf{R}^\prime & =({\mathbf{R}^*}^T\mathbf{K}^\prime \mathbf{R}^*) \mathbf{L}^{\prime T}                                      \\
			& =({\mathbf{R}^*}^T\exp(\phi_{\mathbf{K}^\prime}^\wedge) \mathbf{R}^*) \exp(-\phi_{\mathbf{L}^\prime}^\wedge)
		\end{aligned}
	\end{equation}
	When $\mathbf{R}\in\rm{SO}(3)$, $\mathbf{u}\in\mathbb{R}^3$, there exists Lie group adjoint properties \cite{2017StateEstimation} as
	\begin{equation}
		\label{equation:Lie group adjoint properties}
		\begin{aligned}
			(\mathbf{Ru})^\wedge       & =\mathbf{Ru}^\wedge\mathbf{R}^T       \\
			\exp((\mathbf{Ru})^\wedge) & =\mathbf{R}\exp(\mathbf{u}^\wedge)\mathbf{R}^T
		\end{aligned}
	\end{equation}
	Substitute Eq. (\ref{equation:Lie group adjoint properties}) into Eq. (\ref{equation:Riemannian Distance 1}), receive
	\begin{equation}
		\label{equation:equation:Riemannian Distance 2}
		{\mathbf{R}^*}^T\mathbf{R}^\prime=\exp(-(\mathbf{R}^*\phi_{\mathbf{K}^\prime})^\wedge)\exp(-\phi_{\mathbf{L}^\prime}^\wedge)
	\end{equation}
	Because the $\mathbf{L}^\prime$ and $\mathbf{K}^\prime$ are small, left or right Jacobian is not suitable. Directly apply the BCH formula \cite{1974BCH} to two matrices $\mathbf{X}=\phi_\mathbf{X}^\wedge$ and $\mathbf{Y}=\phi_\mathbf{Y}^\wedge$. Keep the first-order terms as
	\begin{equation}
		\label{equation:BCH}
		\log[\exp(\mathbf{X})\exp(\mathbf{Y})]=\mathbf{X}+\mathbf{Y}+\frac{1}{2}\{\mathbf{X},\mathbf{Y}\}+\ldots
	\end{equation}
	where $\{\mathbf{X},\mathbf{Y}\}$ is the Lie bracket satisfying
	\begin{equation}
		\label{equation:Lie bracket}
		\begin{aligned}
			\{\mathbf{X},\mathbf{Y}\} & =\mathbf{X}\mathbf{Y}-\mathbf{Y}\mathbf{X}                                                     \\
			& =\phi_\mathbf{X}^\wedge \phi_\mathbf{Y}^\wedge - \phi_\mathbf{Y}^\wedge \phi_\mathbf{X}^\wedge \\
			& =(\phi_\mathbf{X}^\wedge \phi_\mathbf{Y})^\wedge
		\end{aligned}
	\end{equation}
	Substituting Eqs. (\ref{equation:BCH}) and (\ref{equation:Lie bracket}) into Eq. (\ref{equation:equation:Riemannian Distance 2}) yields
	\begin{equation}
		\label{equation:Riemannian Distance 3}
		{\mathbf{R}^*}^T\mathbf{R}^\prime=-(\mathbf{R}^*\phi_{\mathbf{K}^\prime})^\wedge-\phi_{\mathbf{L}^\prime}^\wedge+\frac{1}{2}((\mathbf{R}^*\phi_{\mathbf{K}^\prime})^\wedge\phi_{\mathbf{L}^\prime})^\wedge
	\end{equation}
	The Forobenius norm property equals to matrix trace
	\begin{equation}
		\label{equation:Forobenius norm}
		\begin{aligned}
			||\mathbf{X}+\mathbf{Y}||_F^2 & =tr((\mathbf{X}+\mathbf{Y})^T(\mathbf{X}+\mathbf{Y}))                              \\
			& =tr(\mathbf{X}^T\mathbf{X})+tr(\mathbf{Y}^T\mathbf{Y})+2tr(\mathbf{X}^T\mathbf{Y})
		\end{aligned}
	\end{equation}
	The matrix trace property exists
	\begin{equation}
		\label{equation:matrix trace}
		\begin{aligned}
			tr(\mathbf{a}^{\wedge T} \mathbf{b}^\wedge)&=tr\{
			\begin{bmatrix}
				0    & a_3  & -a_2 \\
				-a_3 & 0    & a_1  \\
				a_2  & -a_1 & 0
			\end{bmatrix}
			\begin{bmatrix}
				0    & -b_3 & b_2  \\
				b_3  & 0    & -b_1 \\
				-b_2 & b_1  & 0
			\end{bmatrix}\}\\
			&=tr\begin{bmatrix}
				a_3 b_3+a_2 b_2 &                 &                 \\
				& a_3 b_3+a_1 b_1 &                 \\
				&                 & a_1 b_1+a_2 b_2
			\end{bmatrix}\\
			&=2\mathbf{a}^T\mathbf{b}
		\end{aligned}
	\end{equation}
	then apply Eq. (\ref{equation:matrix trace}) to Eq. (\ref{equation:Forobenius norm}), acquires
	\begin{equation}
		\label{equation:Forobenius norm matrix trace}
		\begin{aligned}
			||\mathbf{X}+\mathbf{Y}||_F^2 & =tr({\phi_\mathbf{X}^\wedge}^T\phi_\mathbf{X}^\wedge)+tr({\phi_\mathbf{Y}^\wedge}^T\phi_\mathbf{Y}^\wedge)+2tr({\phi_\mathbf{X}^\wedge}^T\phi_\mathbf{Y}^\wedge) \\
			& =2\phi_\mathbf{X}^T\phi_\mathbf{X}+2\phi_\mathbf{Y}^T\phi_\mathbf{Y}+4\phi_\mathbf{X}^T\phi_\mathbf{Y}
		\end{aligned}
	\end{equation}
	Substituting Eqs. (\ref{equation:Forobenius norm matrix trace}) and (\ref{equation:Riemannian Distance 3}) into Eq. (\ref{equation:Riemanniandistance}) results in
	\begin{equation}
		\label{equation:Riemannian Distance 4}
		\begin{aligned}
			& Riem(\mathbf{R}^*,\mathbf{R}^\prime)=\mathbf{A}+\mathbf{B}+\mathbf{C}+\mathbf{D}                                                                                                                                    \\
			& -2(\mathbf{R}^* \phi_{\mathbf{K}^\prime})^T (\mathbf{R}^*\phi_{\mathbf{K}^\prime})^\wedge\phi_{\mathbf{L}^\prime}-2\phi_{\mathbf{L}^\prime}^T(\mathbf{R}^*\phi_{\mathbf{K}^\prime})^\wedge \phi_{\mathbf{L}^\prime}
		\end{aligned}
	\end{equation}
	satisfying
	\begin{equation}
		\label{equation:Riemannian Distance 4 ABCD}
		\begin{aligned}
			\mathbf{A}&=2(\mathbf{R}^* \phi_{\mathbf{K}^\prime})^T(\mathbf{R}^* \phi_{\mathbf{K}^\prime})\\
			\mathbf{B}&=2\phi_{\mathbf{L}^\prime}^T\phi_{\mathbf{L}^\prime}\\
			\mathbf{C}&=\frac{1}{2}[(\mathbf{R}^* \phi_{\mathbf{K}^\prime})^\wedge \phi_{\mathbf{L}^\prime}]^T (\mathbf{R}^* \phi_{\mathbf{K}^\prime})^\wedge \phi_{\mathbf{L}^\prime}\\
			\mathbf{D}&=4(\mathbf{R}^*\phi_{\mathbf{K}^\prime})^T\phi_{\mathbf{L}^\prime}
		\end{aligned}
	\end{equation}
	Eq. (\ref{equation:Riemannian Distance 4}) is composed of six terms. But the last two terms are consistently zero. To demonstrate it, a new rotation vector $\mathbf{G}^\prime$ is defined as Eq. (\ref{equation: new vector G})
	\begin{equation}
		\label{equation: new vector G}
		\phi_{\mathbf{G}^\prime}=\mathbf{R}^* \phi_{\mathbf{K}^\prime}
	\end{equation}
	\begin{figure}[htbp]
		\centering
		\includegraphics[width=0.35\textwidth]{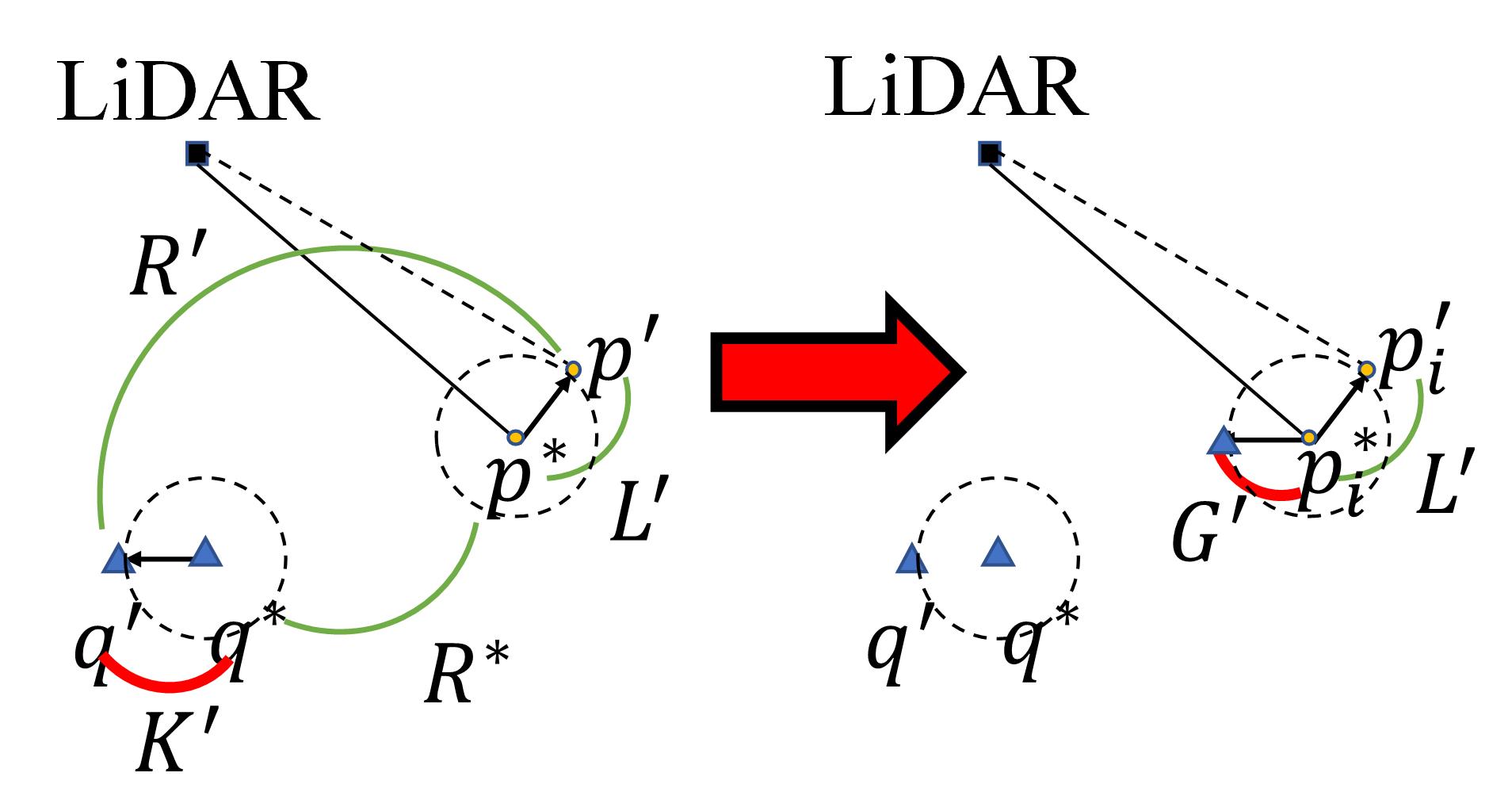}
		\caption{New rotation vector, $\phi_{\mathbf{G}^\prime}$ can be regarded as rotating disturbance from $\mathbf{q}^*$ coordinate to $\mathbf{p}^*$ coordinate.}
		\label{figure:NewRotation}
	\end{figure}
	The structure of this new rotation vector is shown in Fig. \ref{figure:NewRotation} for comprehension, and then
	\begin{equation}
		\label{equation:Riemannian Distance Zero}
		\begin{aligned}
			(\mathbf{R}^* \phi_{\mathbf{K}^\prime})^T (\mathbf{R}^* \phi_{\mathbf{K}^\prime})^\wedge \phi_{\mathbf{L}^\prime} & = \phi_{\mathbf{G}^\prime}^T \phi_{\mathbf{G}^\prime}^\wedge \phi_{\mathbf{L}^\prime}   \\
			& =-(\phi_{\mathbf{G}^\prime}^\wedge \phi_{\mathbf{G}^\prime})^T \phi_{\mathbf{L}^\prime} \\
			& =-\mathbf{0}^T \phi_{\mathbf{L}^\prime}                                                 \\
			& =0
		\end{aligned}
	\end{equation}
	\begin{equation}
		\label{equation:Riemannian Distance Under small disturbance 9}
		\begin{aligned}
			\phi_{\mathbf{L}^\prime}^T (\mathbf{R}^* \phi_{\mathbf{K}^\prime})^\wedge \phi_{\mathbf{L}^\prime} & =\phi_{\mathbf{L}^\prime}^T (\phi_{\mathbf{G}^\prime} \times \phi_{\mathbf{L}^\prime})  \\
			& =-\phi_{\mathbf{G}^\prime}^T (\phi_{\mathbf{L}^\prime} \times \phi_{\mathbf{L}^\prime}) \\
			& =0
		\end{aligned}
	\end{equation}
	Thus, Theorem \ref{theorem:Riemannian Distance} is confirmed;
	\begin{center}
		\bf{Q.E.D.}
	\end{center}
	
	\section*{C.Theorem \ref{theorem:Expectation of Riemannian Distance} Expectation of Riemannian Distance}
\begin{figure}[htbp]
		\centering
		\subfigure[]
		{
			\label{figure:DoubleIntegration}
			\includegraphics[width=0.15\textwidth]{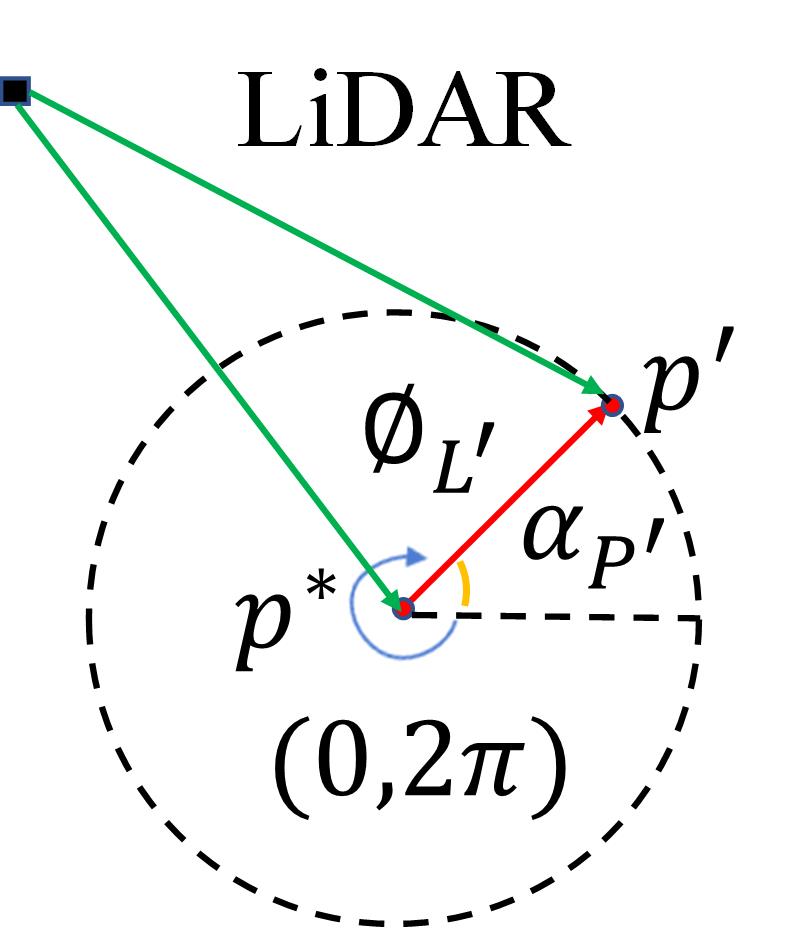}
		}
		\hspace{0.05\textwidth}
		\subfigure[]
		{
			\label{figure:NewNewRotation}
			\includegraphics[width=0.2\textwidth]{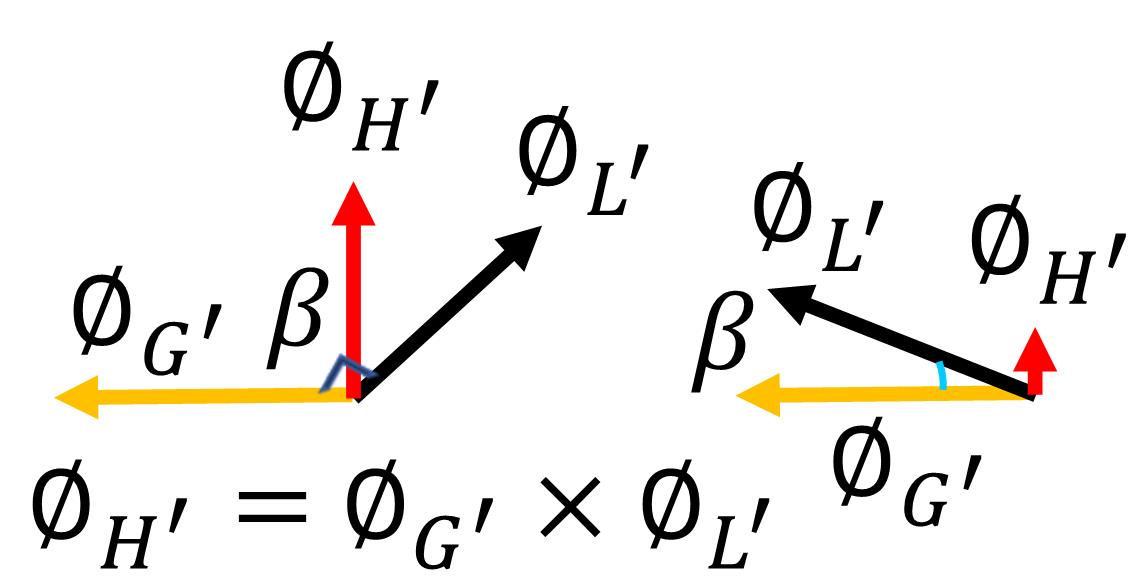}
		}
		\caption{(a) Angle $\alpha_{\mathbf{p}^\prime}$ represents all possible locations of $\mathbf{p}^\prime$ on ring. (b) New rotation vector, $\phi_{\mathbf{H}^\prime}=\phi_{\mathbf{G}^\prime} \times \phi_{\mathbf{L}^\prime}$, where $\phi_{\mathbf{G}^\prime}=\mathbf{R}^* \phi_{\mathbf{K}^\prime}$; $\beta$ is included angle of cross product between $\phi_{\mathbf{G}^\prime}$ and $\phi_{\mathbf{L}^\prime}$}
	\end{figure}

	The expectation of Riemannian distance is calculated by double integration as
	\begin{equation}
		\label{equation:Double Integration}
		\begin{aligned}
			& E(Riem(\mathbf{R}^*,\mathbf{R}^\prime))                                                                                                                                         \\
			& =\frac{1}{2\pi}\int_{0}^{2\pi}\frac{1}{2\pi}\int_{0}^{2\pi}||\log(\mathbf{R}^{*T} \mathbf{R}^\prime)||_F^2 d\alpha_{\mathbf{p}^\prime} d\alpha_{\mathbf{q}^\prime}
		\end{aligned}
	\end{equation}
	where $\alpha_{\mathbf{p}^\prime}$ and $\alpha_{\mathbf{q}^\prime}$ are the possible angles of $\mathbf{p}^\prime$ and $\mathbf{q}^\prime$, respectively. $\alpha_{\mathbf{p}^\prime}$ is shown in Fig. \ref{figure:DoubleIntegration}.
	
	On $\mathfrak{so}(3)$ space, these two vector move on the circle, and double integration can cover all possible arrangements. $\mathbf{A}$ and $\mathbf{B}$ are discussed in Remark \ref{remark:Disturbance Terms} Eq. (\ref{equation:Disturbance Terms A B}). Considering $\mathbf{D}$, the inner integration can hold $\mathbf{R}^* \phi_{\mathbf{K}^\prime}$ as a fixed value with inner integration on $\alpha_{\mathbf{p}^\prime}$. $\mathbf{D}$ becomes a zero vector
	\begin{equation}
		\label{equation:Double Integration D}
		\begin{aligned}
			& \int_{0}^{2\pi} \int_{0}^{2\pi} (\mathbf{R}^* \phi_{\mathbf{K}^\prime})^T \phi_{\mathbf{L}^\prime} d\alpha_{\mathbf{p}^\prime} d\alpha_{\mathbf{q}^\prime}    \\
			& =\int_{0}^{2\pi} (\mathbf{R}^* \phi_{\mathbf{K}^\prime})^T [\int_{0}^{2\pi} \phi_{\mathbf{L}^\prime} d\alpha_{\mathbf{p}^\prime}] d\alpha_{\mathbf{q}^\prime} \\
			& =\int_{0}^{2\pi} (\mathbf{R}^* \phi_{\mathbf{K}^\prime})^T \mathbf{0} d\alpha_{\mathbf{q}^\prime}                                                             \\
			& =0
		\end{aligned}
	\end{equation}
	Considering $\mathbf{C}$, a new vector is also formulated as cross product, as shown in Fig. \ref{figure:NewNewRotation}:
	\begin{equation}
		\label{equation:new vector H}
		\begin{aligned}
			\phi_{\mathbf{H}^\prime} & =\phi_{\mathbf{G}^\prime}^\wedge \phi_{\mathbf{L}^\prime}                \\
			& =\phi_{\mathbf{G}^\prime} \times \phi_{\mathbf{L}^\prime}                \\
			& =(\mathbf{R}^* \phi_{\mathbf{K}^\prime}) \times \phi_{\mathbf{L}^\prime}
		\end{aligned}
	\end{equation}
	
	The integration of $\mathbf{C}$ becomes $\phi_{\mathbf{H}^\prime}^T \phi_{\mathbf{H}^\prime}$. Moreover, it contains one angle, $\alpha_{\mathbf{p}^\prime}$, when LiDAR is implemented. The inner integration and exchange of integration variables from $\alpha_{\mathbf{q}^\prime}$ to $\beta$ shown in Fig. \ref{figure:NewNewRotation} indicate that $\phi_{\mathbf{G}^\prime} \times \phi_{\mathbf{L}^\prime}$ includes the angle:
	\begin{equation}
		\label{equation:Double Integration C}
		\begin{aligned}
			& \int_{0}^{2\pi} \int_{0}^{2\pi} [(\mathbf{R}^* \phi_{\mathbf{K}^\prime})^\wedge \phi_{\mathbf{L}^\prime}]^T (\mathbf{R}^* \phi_{\mathbf{K}^\prime})^\wedge \phi_{\mathbf{L}^\prime} d\alpha_{\mathbf{p}^\prime} d\alpha_{\mathbf{q}^\prime} \\
			& = \int_{0}^{2\pi} \int_{0}^{2\pi} \phi_{\mathbf{H}^\prime}^T \phi_{\mathbf{H}^\prime} d\alpha_{\mathbf{p}^\prime} d\alpha_{\mathbf{q}^\prime}                                                                                               \\
			& =\theta^2_{\mathbf{K}^\prime}\theta^2_{\mathbf{L}^\prime}\int_{0}^{2\pi} \int_{0}^{2\pi} sin^2(\beta) d\beta d\alpha_{\mathbf{p}^\prime}                                                                                                                                                            \\
			& = 2\pi^2\theta^2_{\mathbf{K}^\prime}\theta^2_{\mathbf{L}^\prime}
		\end{aligned}
	\end{equation}
	Substituting Eqs. (\ref{equation:Disturbance Terms A B}), (\ref{equation:Double Integration D}), and (\ref{equation:Double Integration C}) into Eq. (\ref{equation:Double Integration}) yields
	\begin{equation}
		\label{equation:Expectation of Riemannian Distance Finally}
		\begin{aligned}
			& E(Riem(\mathbf{B}^\prime))                                                                                                                                                           \\
			& =\frac{1}{2\pi}\int_{0}^{2\pi}\frac{1}{2\pi}\int_{0}^{2\pi}(2\theta_{\mathbf{K}^\prime}^2+2\theta_{\mathbf{L}^\prime}^2)d\alpha_{\mathbf{p}^\prime}d\alpha_{\mathbf{q}^\prime}+\pi^2 \\
			& =2\theta_{\mathbf{K}^\prime}^2+2\theta_{\mathbf{L}^\prime}^2+\frac{\theta^2_{\mathbf{K}^\prime}\theta^2_{\mathbf{L}^\prime}}{4}
		\end{aligned}
	\end{equation}
	\begin{center}
		\bf{Q.E.D.}
	\end{center}
	
	\section*{Acknowledgments}
	This work was supported by the National Key Research and Development Program of China under Grant 2021ZD0201403, the National Natural Science Foundation of China (Grant No.62088101), the Project of State Key Laboratory of Industrial Control Technology, Zhejiang University, China (No.ICT2021A10), the Open Research Project of the State Key Laboratory of Industrial Control Technology, Zhejiang University ,China (No.ICT2022B04). Zeyu Wan and Yu Zhang contributed equally to this work. The author is very grateful to Chang Deng for supporting experiments and encouragement during my hard time.
	
	\bibliographystyle{ieeetr}
	\bibliography{references}
	
	\begin{IEEEbiography}
		[{\includegraphics[width=1in,height=1.25in,clip,keepaspectratio]{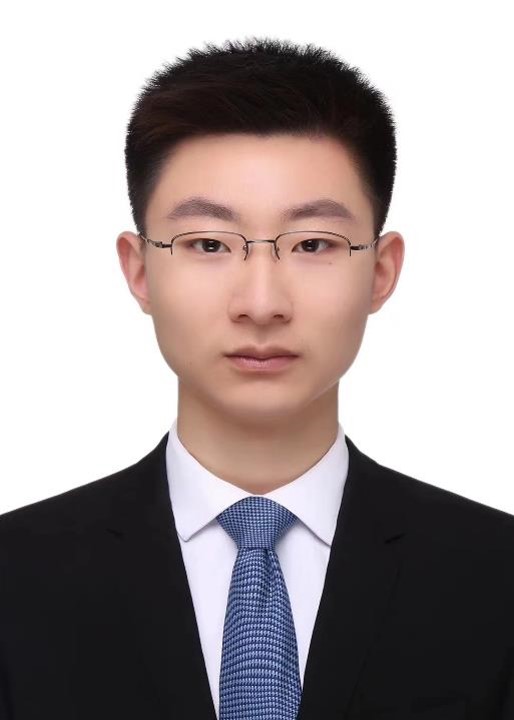}}]{Zeyu Wan}
		received his B.S degree in automation from NanKai University, Tianjin, China, in 2018. He is now a Ph.D student in the College of Control Science and Engineering at Zhejiang University, Hangzhou, China. His research interests include SLAM, 3D reconstruction, machine learning and robotics.
	\end{IEEEbiography}

    \begin{IEEEbiography}
    	[{\includegraphics[width=1in,height=1.25in,clip,keepaspectratio]{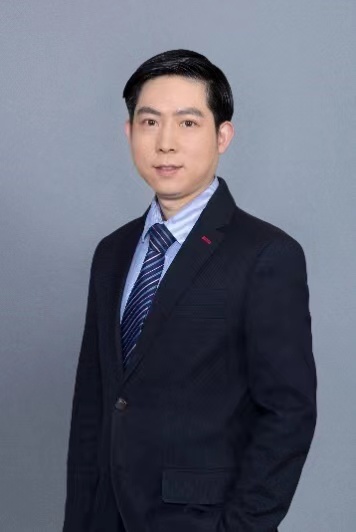}}]{Yu Zhang}
    	received his B.S. degree in information engineering from Xi’an Jiaotong University, Xi’an, China, in 2003, and M.S. and Ph.D. degrees in computer science from Tsinghua University, Beijing, China, in 2009. He is now an Associate Professor at the College of Control Science and Engineering at Zhejiang University, China. His research interests include visual navigation, intelligent control, computer vision, and intelligent autonomous systems.
    \end{IEEEbiography}

    \begin{IEEEbiography}
    	[{\includegraphics[width=1in,height=1.25in,clip,keepaspectratio]{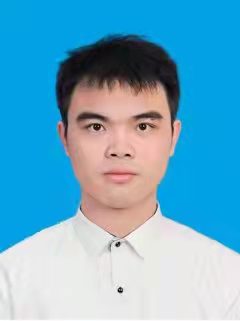}}]{Bin He}
    	received his B.S. degree in the College of Control Science and Engineering at Zhejiang University, Hangzhou, China, in 2021. He is now a M.S student in the College of Control Science and Engineering at Zhejiang University, China. His research interests include multi-sensors fusion, nonlinear optimization, and robotics.
    \end{IEEEbiography}
    
    \begin{IEEEbiography}
    	[{\includegraphics[width=1in,height=1.25in,clip,keepaspectratio]{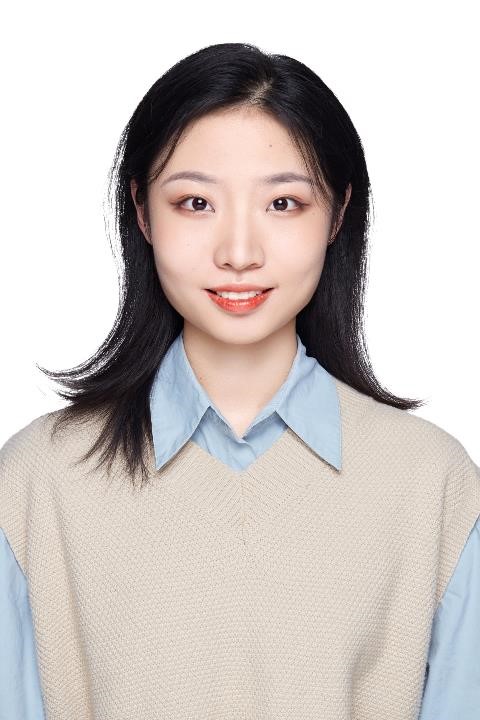}}]{Zhuofan Cui}
    	received her B.S. degree in the College of Control Science and Engineering at Zhejiang University, Hangzhou, China, in 2021. She is now a M.S student in the College of Control Science and Engineering at Zhejiang University, China. Her research interests include 3D reconstruction, autonomous navigation, and robotics.
    \end{IEEEbiography}

    \begin{IEEEbiography}
    	[{\includegraphics[width=1in,height=1.25in,clip,keepaspectratio]{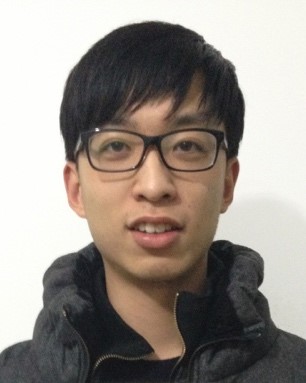}}]{Weichen Dai}
    	received his B.S. degree in information engineering from Zhejiang University of Technology, Hangzhou, China, in 2015, and his Ph.D. degree in the College of Control Science and Engineering at Zhejiang University, Hangzhou, China, in 2021. He is now working in the school of computer science at Hangzhou Dianzi University, China. His research interests include visual navigation, perception, and intelligent autonomous systems.
    \end{IEEEbiography}

    \begin{IEEEbiography}
    	[{\includegraphics[width=1in,height=1.25in,clip,keepaspectratio]{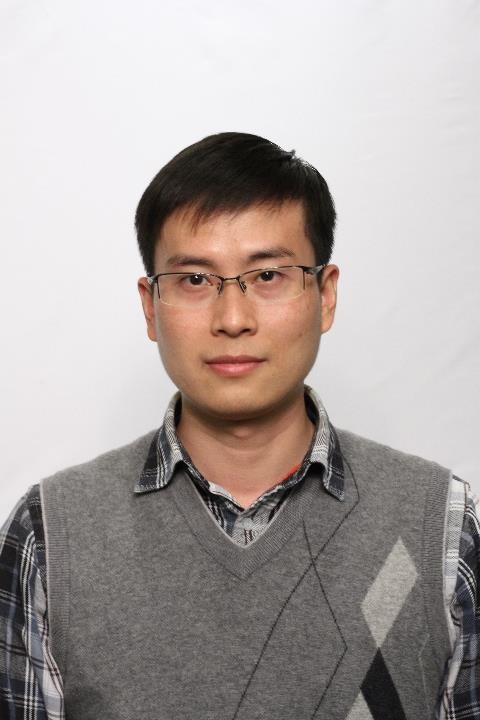}}]{Lipu Zhou}
    	received his B.S. degree from Tianjin University in 2004, M.S. degree from Peking University in 2009, and Ph.D. degree in computer science from Tsinghua University in 2015. He was a Postdoctoral Fellow at Carnegie Mellon University from 2017 to 2019. He was a senior computer vision researcher at Magic Leap, Sunnyvale, US from 2019 to 2021. He is now a senior computer vision researcher at Meituan, Beijing, China. His research interests include SLAM, 3D geometry, computer vision, and machine learning.
    \end{IEEEbiography}

    \begin{IEEEbiography}
    	[{\includegraphics[width=1in,height=1.25in,clip,keepaspectratio]{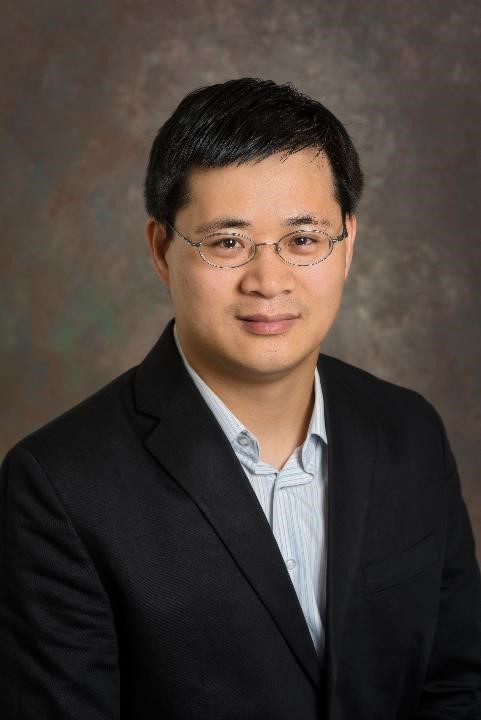}}]{Guoquan Huang}
    	received the B.Eng. degree in automation (electrical engineering) from the University of Science and Technology Beijing, Beijing, China, in 2002, and the M.Sc. and Ph.D. degrees in computer science from the University of Minnesota--Twin Cities, Minneapolis, MN, USA, in 2009 and 2012, respectively. He currently is an Associate Professor of Mechanical Engineering (ME), Computer and Information Sciences (CIS), and Electrical and Computer Engineering (ECE), at the University of Delaware (UD), where he is leading the Robot Perception and Navigation Group (RPNG).  From 2012 to 2014, he was a Postdoctoral Associate with the MIT Computer Science and Artificial Intelligence Laboratory (CSAIL), Cambridge, MA, USA. His research interests focus on state estimation and spatial AI for robotics, including probabilistic sensing, localization, mapping, perception and planning of autonomous ground, aerial and underwater vehicles.
    \end{IEEEbiography}
\end{document}